\documentclass{article}

\PassOptionsToPackage{numbers, compress}{natbib}

\usepackage[main, final]{neurips_2026}

\usepackage[utf8]{inputenc}
\usepackage[T1]{fontenc}
\usepackage{microtype}
\usepackage{graphicx}
\usepackage{subfigure}
\usepackage{booktabs}
\usepackage{longtable}
\usepackage{hyperref}
\usepackage{url}
\usepackage{amsmath}
\usepackage{amssymb}
\usepackage{amsfonts}
\usepackage{mathtools}
\usepackage{amsthm}
\usepackage{nicefrac}

\usepackage{algorithm}
\usepackage{algorithmic}

\usepackage{multicol}
\usepackage{multirow}
\usepackage{colortbl}
\usepackage{xcolor}
\usepackage{color}
\usepackage{xspace}
\usepackage{pifont}
\usepackage{bbding}
\usepackage{enumitem}
\usepackage{tabularx}
\usepackage[bold]{hhtensor}
\usepackage{wrapfig}
\usepackage{etoc}
\usepackage{caption}

\usepackage[capitalize,noabbrev]{cleveref}
\crefname{section}{Sec.}{Secs.}
\Crefname{section}{Section}{Sections}
\Crefname{table}{Table}{Tables}
\crefname{table}{Tab.}{Tabs.}

\theoremstyle{plain}

\theoremstyle{definition}

\theoremstyle{remark}

\usepackage[textsize=tiny]{todonotes}
\usepackage{fontawesome5}

\newcommand{\eg}{\emph{e.g.}}
\newcommand{\ie}{\emph{i.e.}}

\title{UniDG: Unified Defect Generation via\\ Defect-Context Editing}

\author{
  \mdseries
  Yuanting Fan\textsuperscript{1}\thanks{Equal contribution.} ,
  Jun Liu\textsuperscript{1}\footnotemark[1] ,
  Bin-Bin Gao\textsuperscript{1} ,
  Xiaochen Chen\textsuperscript{1} ,\\
  Yuhuan Lin\textsuperscript{1} ,
  Zhewei Dai\textsuperscript{1} ,
  Jiawei Zhan\textsuperscript{1} ,
  Chengjie Wang\textsuperscript{1}\thanks{Corresponding author: \texttt{jasoncjwang@tencent.com}} \\[4pt]
  \mdseries\textsuperscript{1}Tencent Youtu Lab, China
}

\begin{document}
\maketitle

\begin{abstract}
Generating realistic visual defects is crucial for alleviating the scarcity of abnormal samples in anomaly detection, yet existing methods often rely on category-specific few-shot adaptation and struggle to generalize across objects, defect types, and domains.
This limitation stems in part from the lack of paired defect editing data and the fine-grained nature of defect appearance, where subtle variations in scale, texture, and morphology must be preserved while maintaining consistency with the target scene.
To address this challenge, we introduce \textbf{\textit{UDG}}, a dataset of 300K normal-abnormal-mask-caption quadruplets curated from diverse defect-related scenarios, and present \textbf{UniDG}, a unified defect generation model.
UniDG formulates defect synthesis as Defect-Context Editing: it extracts reference defect context with adaptive cropping, organizes reference and target inputs in a structured diptych format, and fuses multimodal conditions through MM-DiT attention.
We further develop a two-stage training strategy: \emph{Diversity-SFT} learns diverse transferable defect priors from paired editing data, while optional \emph{Consistency-RFT} can further improve reference adherence, local realism, and defect-category consistency.
Without per-category fine-tuning or using MVTec-AD/VisA for training, UniDG outperforms prior few-shot anomaly generation and image insertion/editing baselines in both synthesis quality and downstream single- and multi-class anomaly detection/localization.
\end{abstract}

\section{Introduction}
\label{sec:intro}

Anomaly detection is crucial in industrial inspection, medical diagnosis, and other safety-critical applications, but abnormal samples are often scarce in real deployments.
Defect generation therefore offers a practical way to synthesize abnormal samples for training downstream detectors and localizers.
In this paper, we use \emph{defect generation} to refer to synthesizing localized abnormal visual patterns, including industrial defects, natural product anomalies, and medical abnormal regions.
Unlike generic image editing, this task requires a generator to preserve the normal background, obey the target mask, and transfer fine-grained defect appearance from limited references.

\begin{figure}[!t]
    \centering
    \begin{minipage}[t]{0.56\linewidth}
        \centering
        \includegraphics[width=\linewidth]{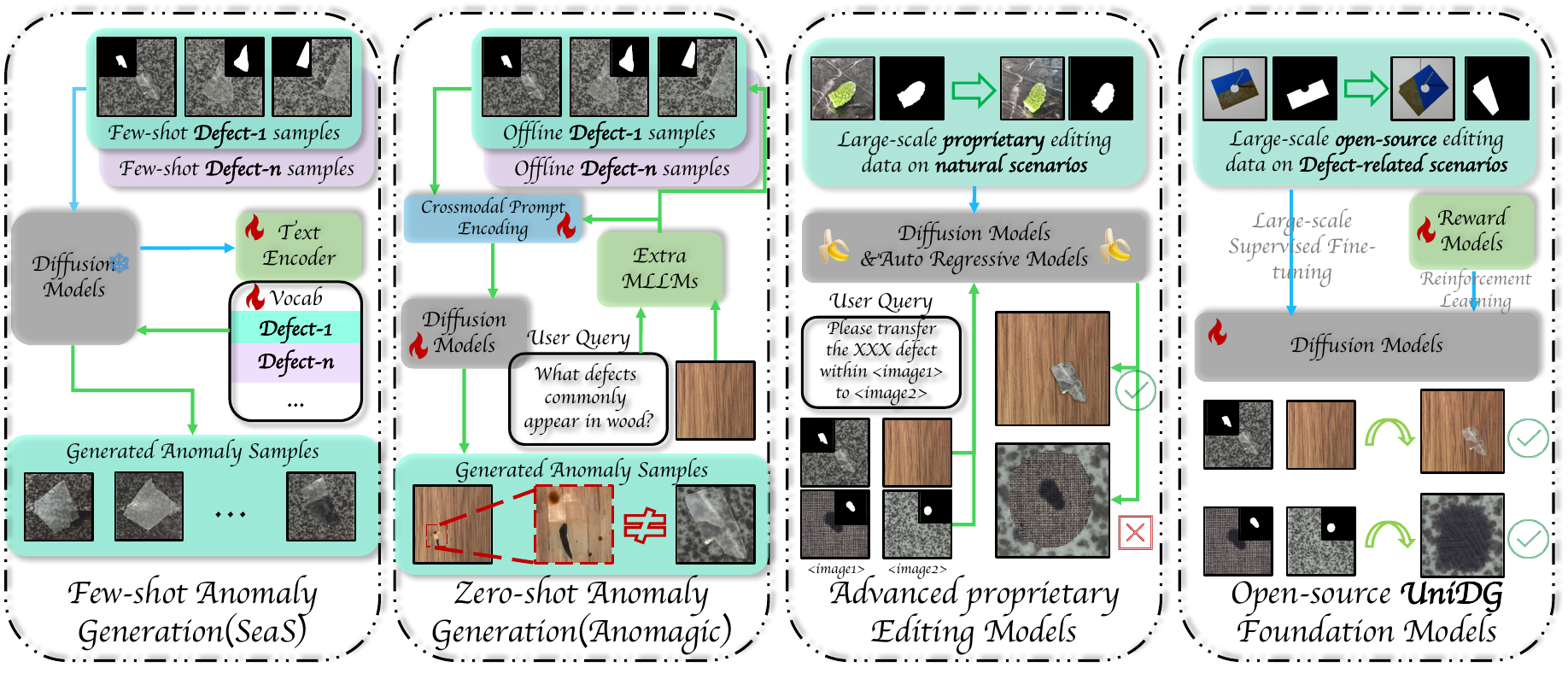}
        \\[2pt]{\small (a) Limitations of existing defect-generation paradigms.}
    \end{minipage}
    \hfill
    \begin{minipage}[t]{0.44\linewidth}
        \centering
        \includegraphics[width=\linewidth]{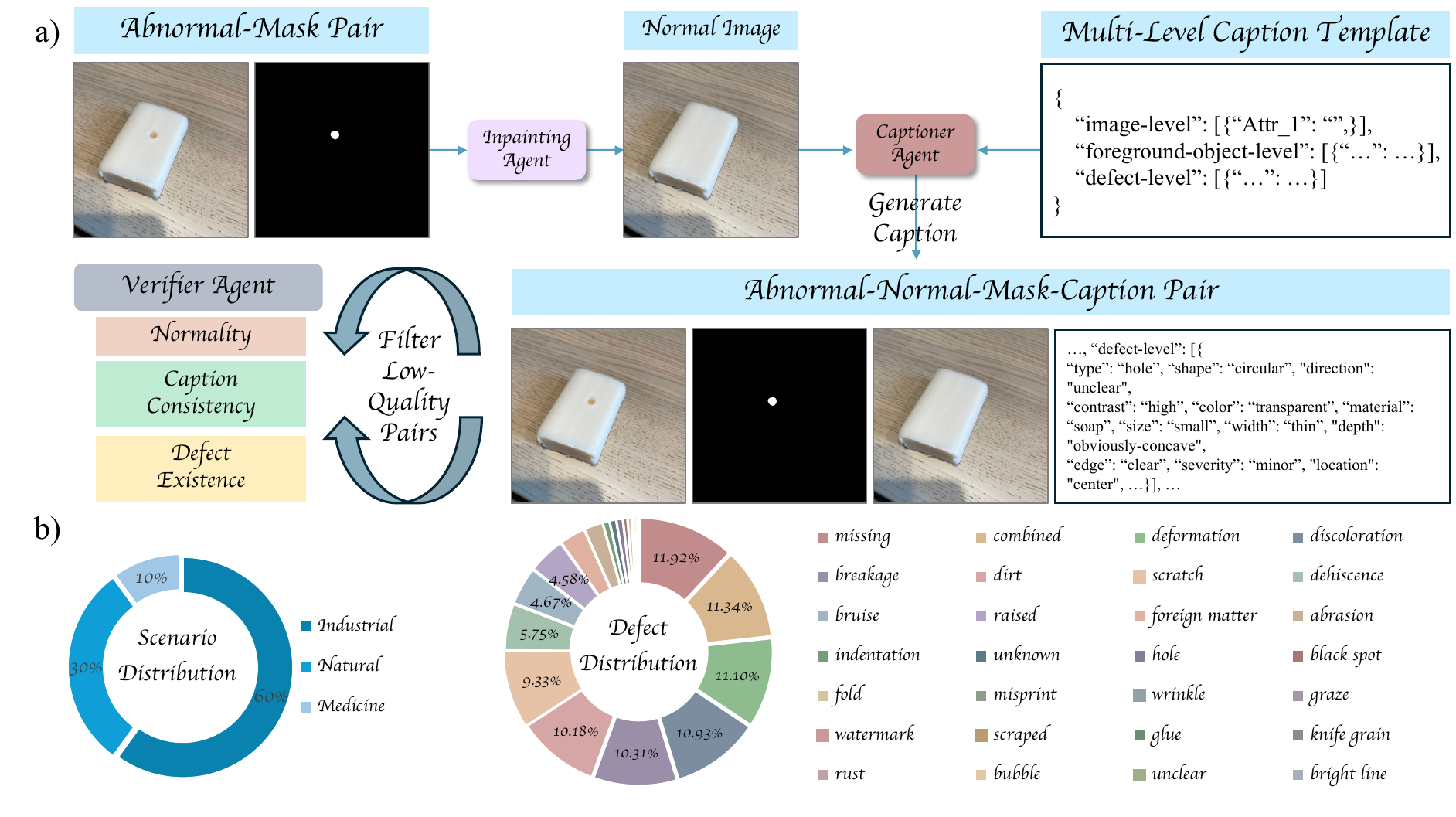}
        \\[2pt]{\small (b) Overview of the \textit{\textbf{UDG}} dataset.}
    \end{minipage}
    \caption{Motivation and the proposed \textit{\textbf{UDG}} dataset. (a) Existing defect-generation paradigms either rely on category-specific few-shot adaptation or on general-purpose image generators that lack defect-specific priors and fine-grained spatial control. (b) The \textit{\textbf{UDG}} dataset is built by a multi-agent curation pipeline that produces normal-abnormal-mask-caption quadruplets across diverse scenarios and mapped defect categories.}
    \label{fig:teaser_and_dataset}
\end{figure}

Existing defect generation methods can be broadly grouped into two paradigms: (1) \emph{zero-shot} approaches that edit normal images from textual descriptions with pre-trained generative models, and (2) \emph{few-shot} approaches that condition on a small set of real abnormal samples to synthesize anomalies with similar appearance statistics.
Despite encouraging progress, both paradigms still struggle to produce defects that are realistic, spatially precise, and visually consistent with the target scene.
As illustrated in Fig.~\ref{fig:teaser_and_dataset}(a), prior methods may generate visually salient anomalous regions, yet often fail to produce realistic and category-consistent defects.

We identify three bottlenecks behind these limitations.
First, the field lacks paired defect editing data, making it difficult to train generators that understand how a normal region becomes a realistic defect while preserving surrounding context.
Second, few-shot methods tend to overfit category-specific appearances and fail to learn transferable defect priors across objects, defect types, and domains.
Third, general-purpose image editors are optimized for semantic edits rather than rare, subtle, and spatially constrained defect patterns, which limits their ability to preserve texture-level details and mask-level precision.

Recent retrieval-based work such as Anomagic~\cite{jiang2025anomagic} partially expands category coverage by retrieving abnormal exemplars with an external MLLM.
However, it still depends on additional models and a large offline anomaly repository, which makes the pipeline costly and hard to scale.
A more scalable alternative is to learn defect priors once from diverse paired editing data, and then apply them to new objects and domains without per-category fine-tuning.

To this end, we formulate unified defect generation as reference-conditioned local editing: given a reference defect, a target image, and a target mask, the model synthesizes a realistic defect in the specified region while preserving the rest of the target scene.
We first construct \textbf{\textit{UDG}}, a dataset of 300K normal-abnormal-mask-caption quadruplets spanning industrial, natural, and medical scenarios.
UDG is curated by a multi-agent pipeline that recovers normal counterparts, generates structured captions, and verifies normality, caption consistency, and defect existence.
On top of UDG, we present \textbf{UniDG}, a unified defect generation model.
UniDG adopts \emph{Defect-Context Editing}, which combines adaptive defect cropping with a structured diptych input and fuses reference and target conditions through MM-DiT multimodal attention.
We further design a two-stage training recipe, \emph{Diversity-SFT} followed by \emph{Consistency-RFT}, to improve synthesis diversity, realism, and reference consistency.

The key contributions are as follows:

\begin{itemize}
    \item We introduce \textbf{\textit{UDG}}, a 300K-sample dataset of normal-abnormal-mask-caption quadruplets curated from diverse defect-related scenarios by a multi-agent pipeline.
    \vspace{-0.1cm}
    \item We present \textbf{UniDG}, a unified model for reference-based defect generation. UniDG uses Defect-Context Editing to transfer fine-grained defect appearance from a reference image to a specified target region.
    \vspace{-0.1cm}
    \item We develop a two-stage training strategy in which \emph{Diversity-SFT} learns diverse cross-object defect transfer from paired editing data, and \emph{Consistency-RFT} uses defect-aware rewards to improve realism, reference adherence, and defect-category consistency.
    \vspace{-0.1cm}
    \item Without using MVTec-AD or VisA for training, UniDG outperforms prior few-shot anomaly generation and image insertion/editing baselines in synthesis quality and downstream single- and multi-class anomaly detection/localization.
\end{itemize}

\section{Related Work}
\label{sec:related}

\noindent\textbf{Few-shot Anomaly Generation}
Few-shot anomaly generation methods use a small number of abnormal samples to synthesize additional anomalies with similar visual characteristics, thereby alleviating data scarcity for downstream anomaly detection.
Early works such as DFMGAN~\cite{duan2023dfmgan} use GANs to produce high-fidelity defects, while recent methods mainly adopt diffusion-based frameworks.
AnomalyDiffusion~\cite{hu2024anomalydiffusion} models anomaly semantics and mask distributions with text inversion; AnoGen~\cite{gui2024anogen} and DefectFill~\cite{song2025defectfill} cast anomaly synthesis as image inpainting; and DualAnoDiff~\cite{jin2025dualanodiff} decouples defect foreground and background modeling.
SeaS~\cite{dai2025seas} binds anomaly attributes to learnable tokens, allowing one model to cover multiple anomaly types.
However, these methods still rely on category-specific adaptation or limited defect coverage, which restricts transfer to new objects and defect types.
Anomagic~\cite{jiang2025anomagic} expands category coverage by retrieving abnormal exemplars with an online MLLM, but it requires an additional model and a large anomaly repository, making the pipeline difficult to scale.

\noindent\textbf{Reference-Based Image Generation}
Reference-based image generation, or subject-driven generation, takes a reference image as input and generates a target image that preserves subject-specific identity and appearance.
Existing methods fall into two groups: test-time fine-tuning methods~(\eg, DreamBooth~\cite{ruiz2023dreambooth} and Textual Inversion~\cite{gal2022textualinversion}, commonly used in FSAG) and training-based methods~(\eg, ControlNet~\cite{zhang2023controlnet} and IP-Adapter~\cite{ye2023ipadapter}).
The former can adapt from few images but generalize poorly across categories.
The latter use large-scale paired data to improve fidelity and semantic consistency; examples include AnyDoor~\cite{chen2024anydoor}, FreeEdit~\cite{he2024freeedit}, InsertAnything~\cite{song2025insertanything}, and recent in-context instruction-editing models such as ICEdit~\cite{zhang2025icedit}.
These works show that diptych or in-context layouts are effective for subject-driven editing.
However, they are mainly designed for natural images and encounter domain gaps in defect generation, where defects differ in scale, texture, and identity.
Motivated by this gap, we adapt reference-based editing to defect generation by coupling a defect-specific paired dataset with a unified model that learns transferable, mask-aligned defect priors.

\section{UDG Dataset}
\label{sec:dataset}

\subsection{Construction Pipeline}
We build a three-agent curation pipeline (Fig.~\ref{fig:teaser_and_dataset}(b)) that converts existing anomaly-mask data into normal-abnormal-mask-caption quadruplets.
Existing anomaly detection datasets~\cite{bergmann2019mvtec,zou2022visa,wang2024realiad} provide abundant abnormal images and pixel-level masks, but they rarely include pixel-aligned normal counterparts or structured descriptions.
These missing annotations are essential for paired defect editing and multimodal defect understanding.

Our pipeline contains three agents.
First, an Inpainting Agent recovers a normal counterpart from each abnormal image and its defect mask.
We train this agent from FLUX.1-Fill-dev on approximately 600K defect-related scenario images, so that it learns to remove localized defects while preserving the surrounding context.
Second, a Captioner Agent uses a structured prompt to describe each normal-abnormal-mask triplet from global image, foreground object, and local defect levels.
Third, a Verifier Agent filters the resulting quadruplets by checking the normality of the recovered image, the consistency of the caption, and the existence of the defect.
Details of agent training and system prompts are provided in Sec.~\ref{sec:app_udg}.

\subsection{Statistics}
The \textbf{\textit{UDG}} dataset aggregates data from 50 publicly accessible defect-related datasets, including Real-IAD~\cite{wang2024realiad}, MANTA~\cite{fan2025manta}, and 3CAD~\cite{yang20253cad}, spanning industrial, natural, and medical scenarios.
Source citations are listed in Sec.~\ref{sec:app_asset_documentation}.
Code is available at \url{https://github.com/RetoFan233/UniDG} under the Apache License 2.0.
The UDG dataset and UDG-Reward-Bench will be released.
It contains 300K normal-abnormal-mask-caption quadruplets, including a curated real-anomaly subset and a verified synthetic extension generated from the curated subset.
To avoid evaluation leakage, images from MVTec-AD and VisA are excluded from UDG training data and are used only for evaluation.

UDG covers 269 original defect types, which we manually map into 28 standardized categories.
Compared with AnomVerse~\cite{jiang2025anomagic}, a recent large-scale anomaly generation dataset, UDG contains approximately 23$\times$ more samples and provides richer supervision: each sample includes a normal counterpart, an abnormal image, a pixel-level mask, and a structured caption.
These annotations support both defect editing supervision and multimodal anomaly detection/generation research.
More details and visual examples are provided in Sec.~\ref{sec:app_udg}.

\section{Preliminaries}

\noindent\textbf{MM-DiT Architecture.}
Recent diffusion models, such as SD3~\cite{esser2024sd3} and the FLUX series~\cite{batifol2025fluxkontext}, adopt MM-DiT, a multimodal extension of Diffusion Transformers~\cite{peebles2023mmdit} with multimodal attention~(MMA), Rotary Position Embedding (RoPE), and RMSNorm.
MM-DiT jointly processes noisy image tokens $X_t \in \mathbb{R}^{n \times d}$ and text tokens $C_{\mathrm{T}} \in \mathbb{R}^{l \times d}$, as summarized in Eq.~\ref{eq:mma}.

\begin{equation}
\text {MMA}\left(\left[X_t ; C_{\mathrm{T}}\right]\right)=\operatorname{softmax}\left(\frac{\mathcal{R}(Q) \cdot \mathcal{R}(K)^{\top}}{\sqrt{d}}\right) V .
\label{eq:mma}
\end{equation}

Here, $Q$, $K$, and $V$ denote the projections of the concatenated input $[X_t; C_{\mathrm{T}}] \in \mathbb{R}^{(n+l)\times d}$, and $\mathcal{R}(\cdot)$ applies RoPE to inject positional information into $Q$ and $K$.

\noindent\textbf{Flow Matching.}
We train UniDG with Rectified Flow (RF)~\cite{liu2022rectifiedflow}.
Under the continuous normalizing flow~(CNF) formulation, the dynamics follow:
\begin{equation}
\frac{d}{d t} X_t=v\left(X_t, t\right)=X_1-X_0, \quad \forall t \in[0,1].
\end{equation}

Given a clean latent variable $X_0 \sim p_{\text{data}}$ and a Gaussian noise $X_1 \sim \mathcal{N}(0,1)$, we obtain $X_t$ by linear interpolation:
\begin{equation}
X_t=t X_1+(1-t) X_0, \quad \forall t \in[0,1] .
\end{equation}

We then optimize the Conditional Flow Matching~(CFM) loss~\cite{lipman2023cfmloss} to learn a velocity field predictor $v_\Theta$:
\begin{equation}
\begin{split}
\mathcal{L}_{\mathrm{CFM}}=\mathbb{E}_{t \sim p(t), X_1 \sim \mathcal{N}(0,1),\left(X_0, C_{\mathrm{T}}\right) \sim p_{\text {data }}} \\
\left[\left\|v_{\Theta}\left(X_t, C_{\mathrm{T}}, t\right)-\left(X_1-X_0\right)\right\|_2^2\right] .
\end{split}
\end{equation}

We sample $t$ from a Logit-Normal distribution~\cite{esser2024sd3} with density
$p(t)=\frac{\exp\left(-0.5\cdot(\operatorname{logit}(t)-\mu)^2/\sigma^2\right)}{\sigma\sqrt{2\pi}\cdot(1-t)\cdot t}$,
where $\operatorname{logit}(t)=\log\frac{t}{1-t}$. Under RF, we use $\mu=0$ and $\sigma=1$.

\begin{figure*}[!t]
    \centering
    \includegraphics[width=\textwidth]{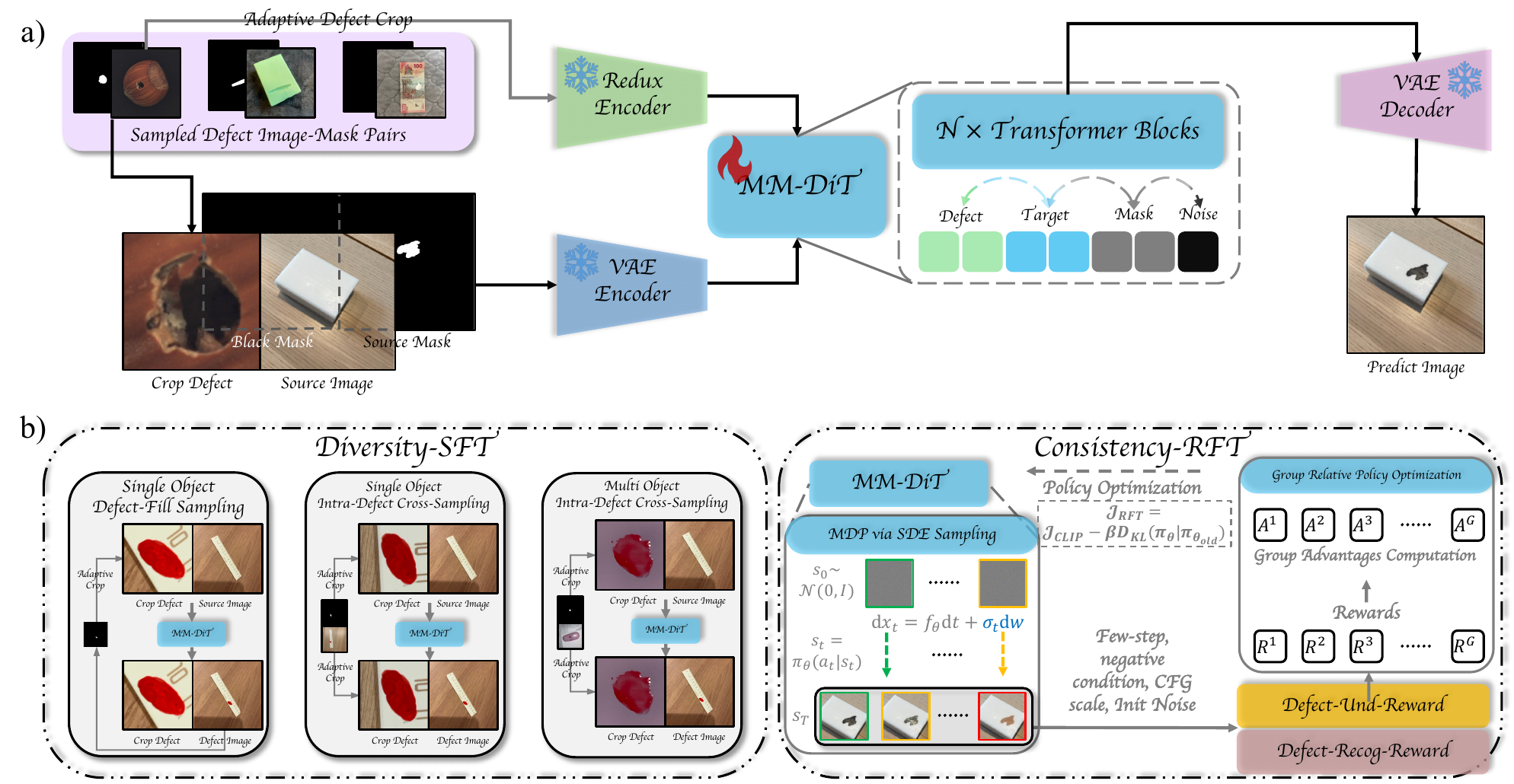}
    \caption{Overall framework of UniDG. UniDG leverages MM-DiT with a Defect-Context Editing strategy to integrate reference defects into target regions. \emph{Diversity-SFT} and \emph{Consistency-RFT} further improve synthesis quality.}
    \label{fig:overview}
\end{figure*}

\section{UniDG Models}
\label{sec:method}

\subsection{Motivation and Overview}

Existing defect generation methods often learn category-specific appearances from a few examples, which limits transfer to new objects, defect types, and scenes.
Retrieval-based systems such as Anomagic~\cite{jiang2025anomagic} reduce this limitation by retrieving abnormal exemplars with an external MLLM, but they require a large offline repository and incur additional inference cost.
UniDG instead learns transferable defect priors from UDG and synthesizes defects from a reference image and a target mask, without external MLLMs or offline template libraries.
We also avoid global/background generation: real-world backgrounds are highly diverse, so reconstructing them from a few samples often introduces artifacts (see Fig.~\ref{fig:qualitative_main}(a)).
Therefore, UniDG adopts a local editing paradigm~(\ie, image inpainting) that concentrates generation on the masked region while preserving the original target background.
As illustrated in Fig.~\ref{fig:overview}, UniDG has three key components~\S\ref{subsec:defect-context}: (1) a diptych input that encodes defect context, (2) an MM-DiT architecture that fuses multi-source conditions via multimodal attention, and (3) defect enhancement mechanisms that improve the fidelity and discriminability of synthesized defects.

\subsection{Defect-Context Editing}
\label{subsec:defect-context}

Defect-Context Editing transfers fine-grained appearance from a reference defect to a specified target region while preserving the surrounding target context.
We first apply Adaptive Defect Crop to isolate the defect subject from the reference image (see Sec.~\ref{subsec:app_adaptive_defect_crop}).
Given the cropped defect subject, we build a diptych input by concatenating it with the target image whose editing region is masked:
\begin{equation}
    I_\text{diptych} = \left[I_\text{cropped\_ref};\, I_\text{masked\_src}\right],
\end{equation}
where $I_\text{cropped\_ref}$ denotes the cropped defect subject and $I_\text{masked\_src}$ denotes the source image (the target image to be edited) with its editing region masked.
We construct the corresponding binary inpainting mask as $M_\text{diptych}=[\mathbf{0}_{h\times w};\, M]$, where the reference half is unmasked and the target half uses the editing mask $M$:
\begin{equation}
    M_\text{diptych} = \left[\mathbf{0}_{h\times w};\, M\right].
\end{equation}
This diptych representation places reference context and target editing guidance in one visual canvas, enabling spatially controlled defect transfer.

\noindent\textbf{Multimodal fusion in MM-DiT.}
UniDG conditions MM-DiT on two streams: (i) the latent inpainting input built from the diptych and the binary mask, and (ii) condition tokens that contain reference-defect semantics and optional text features.
Concretely, the image diptych $I_\text{diptych}$ is encoded by the VAE into latent features, while the binary mask diptych $M_\text{diptych}$ is directly resized to the latent spatial resolution rather than VAE-encoded.
The VAE-encoded image diptych and the resized mask diptych are concatenated along the \emph{channel} dimension and used with the noisy latent for flow matching.
In parallel, a SigLIP image encoder extracts reference-defect semantics, which are concatenated with text-branch features to form the condition tokens.
For reference-based defect synthesis, we zero-pad the text branch, so the condition branch primarily carries defect semantics; for instruction editing, the same branch can also carry text features.
\begin{equation}
\begin{aligned}
Q=[Q_i;\,Q_c],\ \ K=[K_i;\,K_c],\ \ V=[V_i;\,V_c], \\
\operatorname{MMA}(Q,K,V)=\operatorname{softmax}\!\left(\frac{QK^{\top}}{\sqrt{d}}\right)V .
\end{aligned}
\end{equation}
where $[\,;\,]$ denotes sequence-wise concatenation.
Here, $Q_i,K_i,V_i$ come from the noisy diptych latent (image/mask branch), while $Q_c,K_c,V_c$ come from the condition tokens (text + reference-defect semantics).
The VAE decoder then decodes the MM-DiT output, and we take the right half of the diptych as the synthesized defect image.
During training, we add two defect enhancement mechanisms to improve fidelity and discriminability.
\emph{Normal regularization} encourages the synthesized target region to differ from its normal counterpart, while \emph{defect attention} increases $\operatorname{MMA}$ weights on the reference defect and target region.
Ablations and details are provided in experiments and Sec.~\ref{subsec:app_defect_enhancement}.
UniDG also supports instruction-based defect editing; additional details and visualizations are included in Sec.~\ref{subsec:app_instruction_editing}.

\subsection{Training Strategy}
We train UniDG in two stages.
\emph{Diversity-SFT} teaches the model to synthesize diverse defects from paired editing data.
\emph{Consistency-RFT} is an optional alignment stage that can further improve reference adherence, local realism, and task-aware defect-category/region consistency.

\noindent\textbf{Diversity Supervised Fine-Tuning.}
For \emph{Diversity-SFT}, we construct three complementary sampling distributions from \textit{\textbf{UDG}}, as shown in Fig.~\ref{fig:overview}(b).
(1).~\textbf{Single Object Defect-Fill Sampling}: the same defect image serves as both reference and target, enforcing exact reference following and mask-localized editing.
(2).~\textbf{Single Object Intra-Defect Cross-Sampling}: reference and target come from the same object and defect type but different images, encouraging intra-class diversity.
(3).~\textbf{Multi-Object Intra-Defect Cross-Sampling}: reference and target share the defect type but come from different objects, enabling cross-object defect transfer.
Together, these distributions balance reference adherence, diversity, and transferability.
Implementation details are provided in Sec.~\ref{subsec:app_diversity_sft}.

\noindent\textbf{Consistency Reinforcement Fine-Tuning.}
After \emph{Diversity-SFT}, UniDG already produces diverse defects, but reference consistency can still drift in color, texture, or illumination; representative cases are shown in Sec.~\ref{subsec:app_failure_cases_sft}.
We optionally apply \emph{Consistency-RFT} as an alignment stage to improve reference-consistent synthesis while preserving diversity.
A policy-entropy constraint does not reliably preserve that diversity in flow-based RLHF; perceptual entropy is a more direct regularizer for this failure mode~\cite{tan2026policyentropyconstraintfails}.
We construct two complementary reward models and optimize UniDG online with Flow-GRPO~\cite{liu2025flowgrpo}.
\emph{Defect-Und-Reward} learns from high-quality editing pairs and MLLM-generated analyses/scores on reference adherence, region-background consistency, defect reasonability, and visual quality.
\emph{Defect-Recog-Reward} uses defect classification and segmentation models trained on the 28 mapped \textit{\textbf{UDG}} categories, providing task-aware rewards for identity/category and region following.
During online optimization, we sample varied-quality candidates by changing inference steps, negative reference conditions, CFG scale, and initial noise, compute relative advantages from the two rewards, and update the SFT-initialized policy.
The pipeline is illustrated in Fig.~\ref{fig:overview}(b), with details in Sec.~\ref{subsec:app_consistency_rft_framework}.

\vspace{-0.3cm}
\section{Experiments}
\label{sec:exp}

\subsection{Experimental Setup}
\noindent\textbf{Implementation details.}
UniDG builds on FLUX.1-Fill-dev and FLUX.1-Redux-dev: the former initializes the MM-DiT backbone, and the latter initializes the SigLIP image encoder for reference-defect features.
We fine-tune MM-DiT with LoRA (rank 256) on both single- and double-stream blocks, using batch size $64$ for \emph{Diversity-SFT}, batch size $8$ for \emph{Consistency-RFT}, and input resolution $768\times768$.
We use the Prodigy optimizer~\cite{mishchenko2023prodigy} with bias correction and weight decay $0.01$.
\emph{Diversity-SFT} and \emph{Consistency-RFT} run for about $14$K and $37$K iterations, respectively.
At inference, we use 28 denoising steps. \emph{Diversity-SFT} optimizes flow-matching velocity prediction with normal regularization on the target mask region, while \emph{Consistency-RFT} follows the Flow-GRPO objective.

\noindent\textbf{Dataset.}
For evaluation, we use MVTec-AD as the main benchmark and report VisA results in Sec.~\ref{sec:app_quantitative_experiments}.
All MVTec-AD and VisA images are excluded from UniDG training; the remaining datasets in \textit{\textbf{UDG}} are used for training.
To further rule out concept-level leakage, Sec.~\ref{sec:app_contamination} removes all \textit{\textbf{UDG}} samples whose defect categories overlap with MVTec-AD/VisA and retrains UniDG from scratch.
The resulting model shows only minor degradation, supporting that UniDG learns transferable defect priors rather than memorizing overlapping defect concepts.

\noindent\textbf{Metrics.}
For anomaly generation, we report Intra-Cluster LPIPS distance~(IL), IL-a (computed within defect regions), and independent MLLM-based scores for Reference Adherence, Region--Background Consistency, Reasonability, and Quality, using the prompt in Fig.~\ref{fig:defect_und_reward_prompt}. The reported MLLM scores are the equal-weight average of \texttt{gemini-3-pro-preview} and \texttt{gpt-5.1-2025-11-13}, and are not produced by the Defect-Und-Reward model used in RFT.
Distribution-level KID and IS, which the per-category baselines lead, are reported in Sec.~\ref{sec:app_review_followup}.
For anomaly detection and localization, we use image-/pixel-level AUROC (AUROC-I/P), image-/pixel-level average precision (AP-I/P), PRO (PRO-P), Accuracy, and mean intersection over union (mIoU).
Because many prior methods train separate binary segmenters for each object or defect category, we also train a unified multi-class semantic segmentation model on all generated data to evaluate category recognition and localization.
mIoU measures performance in this multi-class setting.
UniDG is not trained on MVTec-AD or VisA, whereas FSAG baselines use these datasets for category-specific training.

\begin{table*}[!t]\small
\centering
\caption{Quantitative results for single-class anomaly classification, detection, and synthesis quality on MVTec-AD. Bold and underline denote the best and second-best average results, respectively. For UniDG and the insertion baselines, no MVTec-AD data are used for generator training; only one retrieved reference image is used at inference. Downstream checkpoints follow the best-of-20 rule in Sec.~\ref{sec:app_downstream_eval}; a fixed-epoch rerun is in Sec.~\ref{sec:app_review_followup}.}
\label{tab:quantitative_classification_detection_generation_comparison}
\resizebox{\textwidth}{!}{
\begin{tabular}{lcccccccc}
\toprule
\multirow{2}{*}{\textbf{Metric}} & \multicolumn{4}{c}{\textbf{Few-shot Learning Anomaly Generation}} & \multicolumn{4}{c}{\textbf{Training-Free Few-shot Image Insertion}} \\
\cmidrule(lr){2-5} \cmidrule(lr){6-9}
 & AnoDiff & AnoGen & DualAnoDiff & SeaS & AnyDoor & InsertAny & \textbf{UniDG-SFT} & \textbf{UniDG-RFT} \\
\midrule
\rowcolor{gray!10} \multicolumn{9}{l}{\scriptsize\textit{Cls.}} \\
Acc-C    & 68.4 & 61.4 & 64.0 & 59.1 & 61.8 & 63.6 & \underline{74.9} & \textbf{77.4} \\
\midrule
\rowcolor{gray!10} \multicolumn{9}{l}{\scriptsize\textit{Det.\&Loc.}} \\
AUROC-I  & 97.9 & 97.0 & 94.9 & 96.2 & 80.3 & 95.3 & \underline{98.0} & \textbf{98.4} \\
AP-I     & \underline{98.9} & 95.0 & 96.6 & 95.5 & 80.7 & 95.5 & 98.6 & \textbf{99.0} \\
AUROC-P  & 96.6 & 98.3 & 97.5 & 98.0 & 88.1 & 97.9 & \underline{98.8} & \textbf{98.8} \\
AP-P     & 75.4 & 63.3 & 66.0 & 68.2 & 18.1 & 64.1 & \underline{77.0} & \textbf{77.3} \\
\midrule
\rowcolor{gray!10} \multicolumn{9}{l}{\scriptsize\textit{Syn.}} \\
IL       & 0.28 & 0.27 & \textbf{0.37} & \underline{0.34} & 0.33 & 0.26 & 0.29 & 0.28 \\
IL-a     & 0.14 & 0.11 & \underline{0.18} & \textbf{0.19} & 0.14 & 0.13 & \underline{0.18} & 0.16 \\
RefAd    & 3.84 & 3.50 & 3.32 & 3.26 & 3.48 & 3.81 & \underline{3.90} & \textbf{4.04} \\
Cons     & 3.97 & 3.87 & 3.57 & 3.76 & 3.76 & 3.98 & \underline{4.15} & \textbf{4.21} \\
Reas     & 4.18 & 3.91 & 3.67 & 3.74 & 3.83 & 4.12 & \underline{4.18} & \textbf{4.20} \\
Qual     & 4.01 & 3.77 & 3.55 & 3.58 & 3.70 & 3.99 & \underline{4.09} & \textbf{4.24} \\
\bottomrule
\end{tabular}
}
\end{table*}

\begin{table*}[!t]\small
\centering
\caption{Quantitative results for multi-class anomaly semantic segmentation on the MVTec-AD.}
\label{tab:quantitative_multi_class_comparison}
\resizebox{0.95\textwidth}{!}{
\begin{tabular}{lcccccccc}
\toprule
\multirow{2}{*}{Methods} & \multicolumn{5}{c}{Binary Anomaly Detection Metrics} & \multicolumn{3}{c}{Semantic Segmentation Metrics}\\
 & AUROC-I & AP-I & AUROC-P & AP-P & PRO-P & mIoU & BG-mIoU & FG-mIoU \\ \midrule

\multicolumn{9}{@{}l@{}}{\cellcolor{gray!10}\scriptsize\textit{Few-shot Anomaly Generation}} \\
AnoDiff & 92.79 & 96.24 & 95.90 & 61.41 & 81.82{\color{gray!60}\scriptsize$_{+0.0\%}$} & 37.87{\color{orange!60}\scriptsize$_{-4.7\%}$} & 98.47 & 23.83{\color{orange!60}\scriptsize$_{-9.9\%}$} \\
AnoGen & 93.10 & 96.20 & 94.85 & 63.32 & 79.54{\color{orange!60}\scriptsize$_{-2.8\%}$} & 39.72{\color{gray!60}\scriptsize$_{+0.0\%}$} & 98.54 & 26.45{\color{gray!60}\scriptsize$_{+0.0\%}$} \\
DualAnoDiff & 91.00 & 95.40 & 95.31 & 63.19 & 79.96{\color{orange!60}\scriptsize$_{-2.3\%}$} & 39.43{\color{orange!60}\scriptsize$_{-0.7\%}$} & 98.57 & 26.27{\color{orange!60}\scriptsize$_{-0.7\%}$} \\
SeaS & 89.29 & 93.76 & 95.89 & 60.98 & 81.01{\color{orange!60}\scriptsize$_{-1.0\%}$} & 35.41{\color{orange!60}\scriptsize$_{-10.9\%}$} & 98.33 & 20.48{\color{orange!60}\scriptsize$_{-22.6\%}$} \\

\multicolumn{9}{@{}l@{}}{\cellcolor{gray!10}\scriptsize\textit{Training-Free Few-shot Image Insertion}} \\
AnyDoor & 90.59 & 95.58 & 91.32 & 50.56 & 79.10{\color{orange!60}\scriptsize$_{-3.3\%}$} & 27.10{\color{orange!60}\scriptsize$_{-31.8\%}$} & 97.40 & 11.11{\color{orange!60}\scriptsize$_{-58.0\%}$} \\
InsertAnything & 84.97 & 91.76 & 93.96 & 54.63 & 77.45{\color{orange!60}\scriptsize$_{-5.3\%}$} & 31.14{\color{orange!60}\scriptsize$_{-21.6\%}$} & 97.77 & 15.63{\color{orange!60}\scriptsize$_{-40.9\%}$} \\
\textbf{UniDG-SFT}& \underline{95.13} & \underline{97.81} & \textbf{97.99} & \underline{71.32} & \underline{90.98}{\color{cyan!60}\scriptsize$_{+11.2\%}$} & \underline{41.81}{\color{cyan!60}\scriptsize$_{+5.3\%}$} & \underline{98.65} & \underline{28.52}{\color{cyan!60}\scriptsize$_{+7.8\%}$} \\
\textbf{UniDG-RFT} & \textbf{96.49} & \textbf{98.24} & \underline{97.83} & \textbf{72.53} & \textbf{91.12}{\color{cyan!60}\scriptsize$_{+11.4\%}$} & \textbf{44.21}{\color{cyan!60}\scriptsize$_{+11.3\%}$} & \textbf{98.66} & \textbf{31.87}{\color{cyan!60}\scriptsize$_{+20.5\%}$} \\ \bottomrule
\end{tabular}
}
\end{table*}

\subsection{Quantitative Results}

\begin{figure}[!t]
    \centering
    \begin{minipage}[c]{0.34\linewidth}
        \centering
        \resizebox{\linewidth}{!}{
        \begin{tabular}{lcccccc}
            \toprule
            \multirow{2}{*}{Arch.} & \multicolumn{3}{c}{Single-class Metrics}& \multicolumn{3}{c}{Multi-class Metrics} \\
            ~ & Acc & AP-I & AP-P & PRO-P & mIoU & FG-mIoU \\
            \midrule
            Baseline          & 57.8 & 92.5 & 43.1 & 75.3 & 29.4 & 13.3 \\
            + Diptych input   & 62.3 & 94.0 & 51.6 & 78.5 & 32.3 & 16.4 \\
            + Normal Reg      & 64.9 & 94.9 & 56.7 & 82.1 & 34.9 & 22.1 \\
            + Defect Attn     & 65.5 & 95.8 & 60.1 & 84.0 & 36.1 & 22.8 \\
            + Diversity-SFT   & 74.9 & 98.6 & 77.0 & 91.0 & 41.8 & 28.5 \\
            + Consistency-RFT & 77.4 & 99.0 & 77.3 & 91.1 & 44.2 & 31.9 \\
            \bottomrule
        \end{tabular}}
        \captionof{table}{Ablation studies on the components within UniDG.}
        \label{tab:overall_ablation}
    \end{minipage}
    \hfill
    \begin{minipage}[c]{0.26\linewidth}
        \centering
        \includegraphics[width=\linewidth]{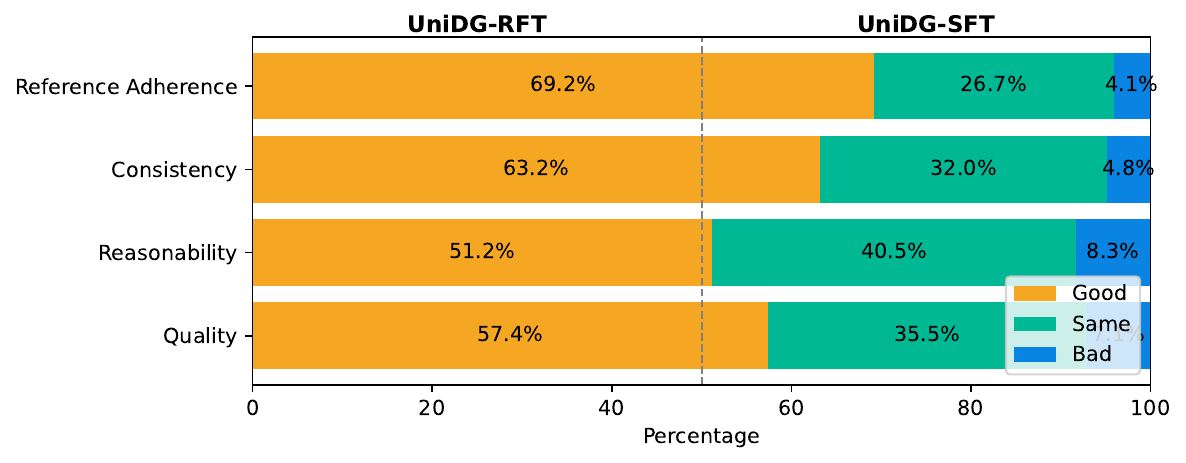}
        \captionof{figure}{Good-Same-Bad~(GSB) evaluation.}
        \label{fig:gsb}
    \end{minipage}
    \hfill
    \begin{minipage}[c]{0.36\linewidth}
        \centering
        \resizebox{\linewidth}{!}{
        \begin{tabular}{lllccccc}
            \toprule
            \multicolumn{3}{c}{Data Proportion} & \multicolumn{3}{c}{Multi-class Metrics} & \multicolumn{2}{c}{Synthesis Metrics} \\
            (a) & (b) & (c) & PRO-P & mIoU & FG-mIoU & Cons & Qual \\
            \midrule
            0 & 1 & 1 & 76.2 & 30.3 & 14.5 & 3.82 & 3.73 \\
            1 & 0 & 1 & 78.3 & 31.4 & 15.7 & 3.91 & 3.87 \\
            1 & 1 & 0 & 90.5 & 39.1 & 26.0 & 3.89 & 3.92 \\
            1 & 1 & 1 & 91.0 & 41.8 & 28.5 & 4.15 & 4.09 \\
            2 & 1 & 1 & 90.9 & 41.9 & 28.5 & 4.14 & 4.12 \\
            1 & 2 & 1 & 89.3 & 41.8 & 28.5 & 4.08 & 4.15 \\
            1 & 1 & 2 & 90.1 & 41.3 & 28.1 & 4.19 & 4.11 \\
            \bottomrule
        \end{tabular}}
        \captionof{table}{Ablation studies on the training data allocation.}
        \label{tab:data_allocation}
    \end{minipage}
    \vspace{-0.3cm}
\end{figure}

\noindent\textbf{Baseline.}
We compare with representative few-shot anomaly generation (FSAG) methods, including AnomalyDiffusion~\cite{hu2024anomalydiffusion}, AnoGen~\cite{gui2024anogen}, DualAnoDiff~\cite{jin2025dualanodiff}, and SeaS~\cite{dai2025seas}, as well as reference-based image insertion/editing methods, including AnyDoor~\cite{chen2024anydoor} and InsertAnything~\cite{song2025insertanything}.
We evaluate synthesis quality, anomaly classification, and anomaly detection/localization.
For fairness, we fine-tune the image insertion baselines on \textit{\textbf{UDG}} with the \emph{Diversity-SFT} recipe.
For FSAG methods, we use the released synthesis results and select 500 generated images per defect class.
For image insertion methods, we use four reference samples per defect class and randomly select one reference image at inference time~(details in Sec.~\ref{subsec:app_training_free_zero_shot}); we also reuse the top 500 source images and target-region masks from AnomalyDiffusion for consistency.
Unless otherwise specified, the same 500 generated images are used to train downstream classification and localization models, so our metrics may differ from those reported in the original papers.
Although some FSAG baselines rely on earlier diffusion backbones, we provide controlled comparisons to isolate this factor.
InsertAnything uses the same FLUX.1-Fill-dev and FLUX.1-Redux-dev backbones as UniDG and is fine-tuned on \textit{\textbf{UDG}} with the same SFT recipe; moreover, Tab.~\ref{tab:overall_ablation} ablates UniDG components under the same backbone and parameter scale.
These controlled comparisons help disentangle the effect of backbone capacity from the proposed defect-specific data, input formulation, and training strategy, with detailed analyses provided in Sec.~\ref{sec:app_arch_vs_training}.

\noindent\textbf{Downstream Performance.}
Following prior work, we conduct three downstream evaluations on MVTec-AD.
First, we train object-specific binary U-Net segmenters for anomaly detection and localization, using 500 generated anomaly images and masks per object category.
As shown in Tab.~\ref{tab:quantitative_classification_detection_generation_comparison}, UniDG achieves the best overall image- and pixel-level performance, with especially strong gains on AP-P.
The main role of \emph{Consistency-RFT} is not to substantially increase already-saturated AP/PRO scores, but to improve reference adherence, local realism, and defect-category consistency while preserving downstream utility.

Second, we train object-specific ResNet-34 classifiers with 500 generated anomaly images to evaluate within-object anomaly classification.
UniDG improves accuracy by \textbf{10\%} over prior methods, indicating better defect-category consistency while maintaining strong localization performance.

Third, to reflect deployment settings where a unified model is preferred, we train a multi-class semantic segmentation U-Net with 74 output channels (73 defect classes plus background) on all generated data.
We report segmentation metrics on the 73 foreground defect classes and compute anomaly detection metrics by merging all predicted defect classes into one foreground class.
As shown in Tab.~\ref{tab:quantitative_multi_class_comparison}, UniDG outperforms prior methods across all metrics.
UniDG-SFT improves PRO-P by \textbf{11.2\%} over AnomalyDiffusion, and UniDG-RFT improves FG-mIoU by \textbf{20.5\%} over AnoGen.
Because FG-mIoU measures multi-class defect recognition and localization, these gains indicate stronger defect-category adherence, further improved by \emph{Consistency-RFT}.

\begin{figure}[t]
    \centering
    \begin{minipage}[t]{0.39\linewidth}
        \centering
        \includegraphics[width=\linewidth]{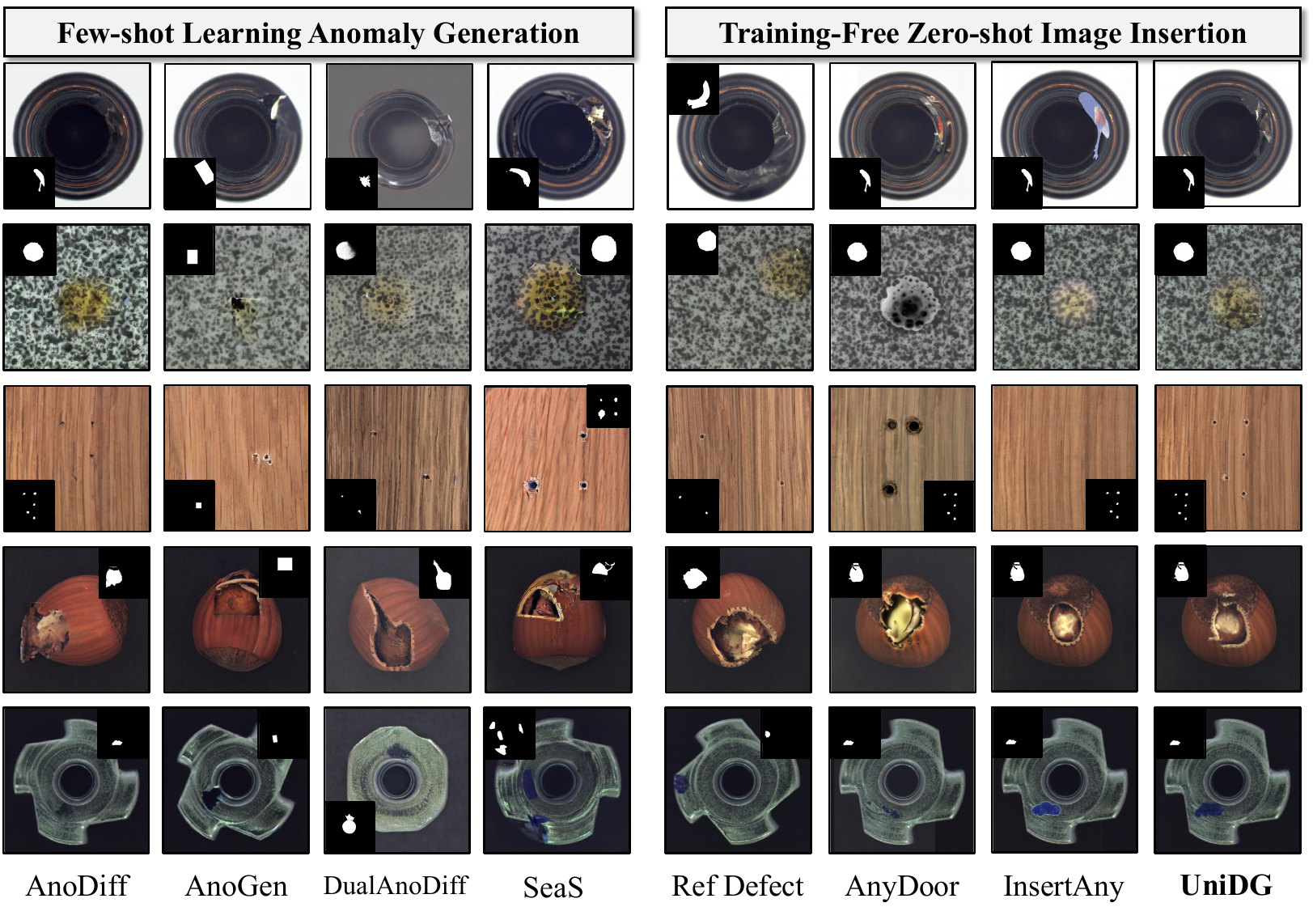}
        \\[2pt]{\small (a) Anomaly generation on MVTec-AD.}
    \end{minipage}
    \hfill
    \begin{minipage}[t]{0.60\linewidth}
        \centering
        \includegraphics[width=\linewidth]{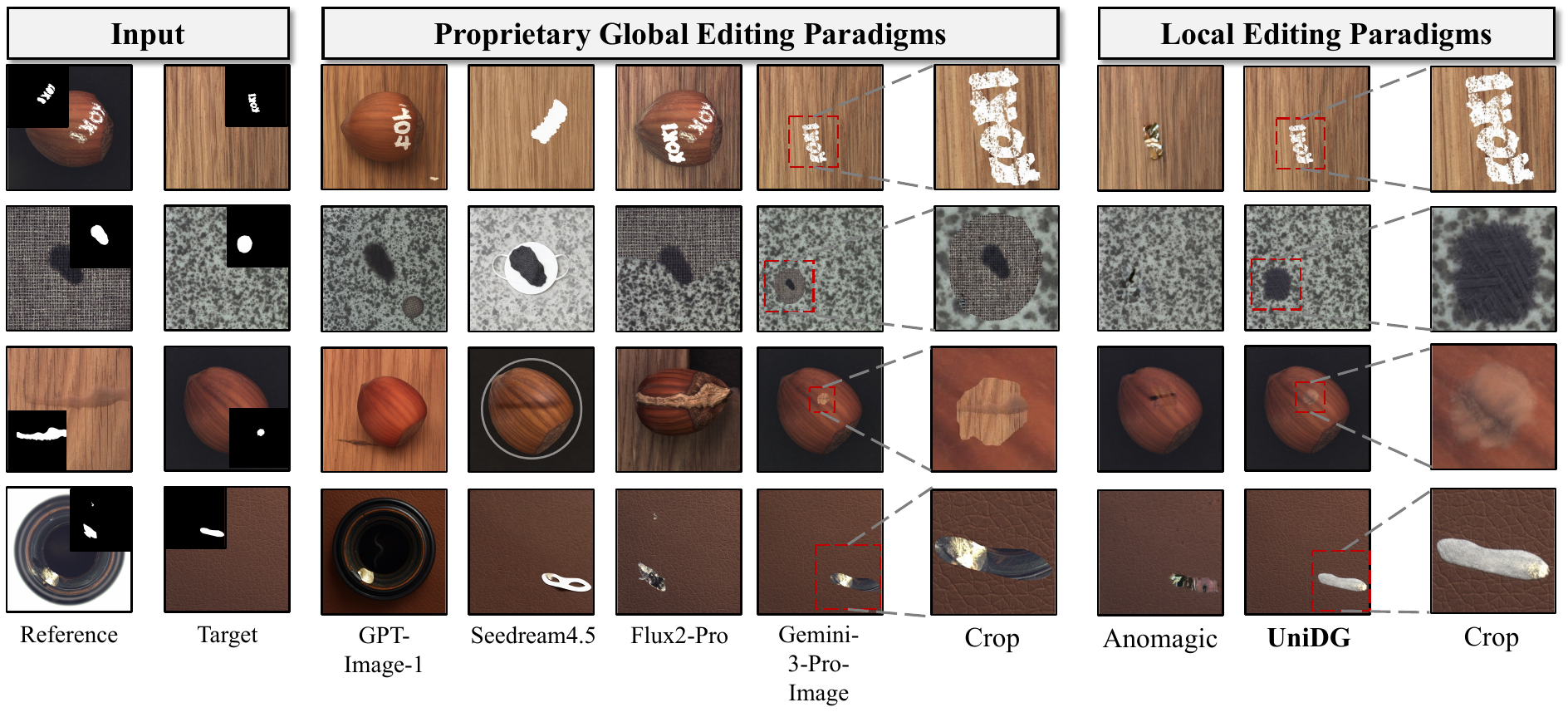}
        \\[2pt]{\small (b) Cross-object defect generation on MVTec-AD.}
    \end{minipage}
    \caption{Qualitative comparisons on MVTec-AD. (a) Anomaly generation under the standard single-object setting. (b) Cross-object defect generation, in which a defect from one object category is transferred to a different object category.}
    \label{fig:qualitative_main}
    \vspace{-0.4cm}
\end{figure}

\noindent\textbf{Generation Quality.}
We further evaluate generation quality with IL/IL-a and MLLM-based scores.
As shown in Tab.~\ref{tab:quantitative_classification_detection_generation_comparison}, UniDG-SFT and UniDG-RFT outperform prior methods by a large margin on MLLM-based metrics.
After \emph{Consistency-RFT}, IL and IL-a decrease because these metrics mainly reflect diversity, whereas RFT emphasizes reference adherence and within-category consistency.
A larger IL/IL-a therefore does not necessarily indicate more realistic defects.
In our setting, MLLM-based evaluation and downstream AD performance better reflect perceptual realism and practical utility.
The reward-weight ablation in Tab.~\ref{tab:app_reward_model} further shows that the two rewards trade off perceptual quality and downstream recognition, so we use balanced weights to improve realism and reference adherence while preserving downstream utility.
Overall, \emph{Consistency-RFT} should be interpreted as a perceptual and semantic alignment stage that improves reference adherence and perceived quality, rather than a stage designed to maximize already-saturated AP/PRO metrics.

\subsection{Qualitative Results}
\vspace{-0.2cm}

We visualize representative results in Fig.~\ref{fig:qualitative_main}(a).
UniDG produces more realistic defects and better preserves category consistency with the reference defect.
In the second row, UniDG follows the reference color, texture, and illumination more faithfully, whereas other methods show visible appearance or morphology shifts.
In the third row, SeaS fails to place the hole defect in the mask-specified region, while UniDG better preserves spatial accuracy under diverse reference conditions.

We further compare cross-object defect synthesis among UniDG, Anomagic, and advanced proprietary methods in Fig.~\ref{fig:qualitative_main}(b).
Anomagic is not trained for cross-object defect transfer and therefore underperforms in this setting.
UniDG also shows stronger fidelity and reference adherence than nano-banana-pro on rare defects, suggesting that \textit{\textbf{UDG}} and \emph{Diversity-SFT} provide useful cross-object transfer priors.

\subsection{Ablation Study}
We ablate UniDG components, evaluate the effect of \emph{Consistency-RFT}, and study data allocation in \emph{Diversity-SFT}.
Additional ablations are provided in Sec.~\ref{sec:app_ablation}.

\noindent\textbf{Overall ablation.}
To evaluate each component, we incrementally add modules to the baseline model (\ie, FLUX.1-Fill-dev\&FLUX.1-Redux-dev) and report results in Table~\ref{tab:overall_ablation}.
Without \emph{Diversity-SFT} (the first four rows), training uses only Single Object Defect-Fill sampling.
In this setting, diptych input, normal regularization, and defect attention each bring consistent gains.
The largest improvement comes from \emph{Diversity-SFT}, especially on FG-mIoU (+15.2 vs.\ baseline), showing that diverse training distributions improve reference-defect following and discrimination.
Defect attention is the weakest increment (+0.6 Acc, +0.7 FG-mIoU).
The data-formulation components account for most of the gain, and stopping after \emph{Diversity-SFT} recovers the large majority of it; the breakdown is in Sec.~\ref{sec:app_review_followup}.
\emph{Consistency-RFT} is an optional alignment stage on top of that recipe.

\noindent\textbf{Human Evaluation.}
To assess \emph{Consistency-RFT}, we conduct a user study with 30 participants.
Participants rate generated images on reference adherence, consistency, reasonability, and overall quality using a 1--5 scale.
For Fig.~\ref{fig:gsb}, we convert these ratings into a Good-Same-Bad~(GSB) comparison: for each evaluated case, we average the four dimension scores for UniDG-RFT and UniDG-SFT; RFT is counted as Good, Same, or Bad when its average score is higher than, equal to, or lower than the SFT score, respectively.
As shown in Fig.~\ref{fig:gsb}, \emph{Consistency-RFT} notably improves reference adherence, consistency, and overall quality over the SFT model, with a smaller gain in reasonability.
These results support the effectiveness of Defect-Und-Reward and Defect-Recog-Reward for improving perceived generation quality.

\noindent\textbf{Data allocation during Diversity-SFT.}
To study data allocation in \emph{Diversity-SFT}, we vary the proportions of the three training distributions and report results in Tab.~\ref{tab:data_allocation}.
(a), (b), and (c) denote Single Object Defect-Fill, Single Object Intra-Defect Cross-Sampling, and Multi-Object Intra-Defect Cross-Sampling, respectively.
Removing any distribution degrades downstream performance, especially removing Single Object Defect-Fill samples.
Removing Multi-Object Intra-Defect Cross-Sampling causes only mild detection/localization degradation but noticeably lowers visual quality.
When all distributions are present, increasing any single proportion yields limited gains, so we adopt a balanced allocation for later scaling experiments.

\section{Limitations and Future Work}
\label{sec:limit}
Although UniDG achieves strong generation quality and downstream performance without target-benchmark training, it still has limitations.
First, the default backbone has 12B parameters. A 3.2B FLUX-Mini variant and a 14-versus-28 NFE comparison are in Sec.~\ref{sec:app_review_followup}; W4A4 quantization reduces precision, not capacity.
Second, evaluation protocols for defect generation remain imperfect. Metrics such as IS and IL often correlate weakly with downstream detection performance and human judgment, and per-category baselines lead on KID (Sec.~\ref{sec:app_review_followup}).
Third, synthesis can still fail under extreme scale, lighting, or texture mismatch; failure rates are quantified in Sec.~\ref{sec:app_review_followup}.
Cross-domain results are transfer to the evaluated held-out benchmarks, not unrestricted generalization.

\section{Conclusion}
UniDG shows that defect generation benefits from treating defects as local, context-dependent visual phenomena rather than as generic text-guided edits.
The UDG dataset supplies paired normal-abnormal-mask-caption supervision at scale, while Defect-Context Editing preserves the reference defect and the target scene in one visual canvas.
The experiments suggest a practical division of labor: \emph{Diversity-SFT} learns transferable defect priors from heterogeneous data, and optional \emph{Consistency-RFT} can further refine reference adherence and local realism without sacrificing downstream utility.
Together, these components enable a single model to synthesize realistic, mask-aligned defects for MVTec-AD and VisA without per-category fine-tuning.
Code and released checkpoints are available at \url{https://github.com/RetoFan233/UniDG} under the Apache License 2.0.
The UDG dataset and UDG-Reward-Bench will be released.

\begin{ack}
This work was supported by Tencent Youtu Lab.
We thank the reviewers and the area chair for their comments.
\end{ack}

\bibliographystyle{plainnat}
\bibliography{main}

\newpage
\appendix

\section{Broader Impacts and Responsible Release}
\label{sec:app_broader_impacts}
UniDG can reduce the cost of collecting rare abnormal samples and may improve inspection systems in industrial, medical, and other safety-critical settings.
The same capability could be misused to fabricate defect evidence or manipulate quality-control records.
Code and released checkpoints are available at \url{https://github.com/RetoFan233/UniDG} under the Apache License 2.0.
The UDG dataset and UDG-Reward-Bench will be released.
FLUX.1-Fill-dev and FLUX.1-Redux-dev weights remain under the Black Forest Labs license and are not covered by the code license.
Medical images, quadruplets derived from them, and synthetic medical images are not redistributed (Sec.~\ref{sec:app_asset_documentation}).
Intended uses are research on defect generation and anomaly detection.
Out-of-scope uses include fabricating inspection evidence and making a safety-critical decision without human review.

\section{Details of the \textit{\textbf{UDG}} Dataset}
\label{sec:app_udg}

\subsection{Multi-Agent System for Dataset Curation}
Sec.~3.1 introduces the multi-agent system used to curate the \textit{\textbf{UDG}} dataset.
The system contains three agents: an Inpainting Agent, a Captioner Agent, and a Verifier Agent.

\textbf{Inpainting Agent.}
The Inpainting Agent generates normal counterparts given abnormal images and their corresponding masks.
We adopt the pre-trained inpainting model FLUX.1-Fill-dev as the base model.
We collect normal images and abnormal images (with masks) from 50 existing anomaly detection datasets, totaling approximately 600K defect-related scenario images.
The 600K defect-related scenario images exclude MVTec-AD and VisA to avoid evaluation leakage.
We then randomly mask parts of these images and train the model to reconstruct the normal content within the masked regions~(\ie, remove \emph{defects}).
For normal images, we use SAM~\cite{ravi2024samv2} to extract foreground regions and apply a pre-constructed mask template library to sample diverse masks (1--3 masks) for training.
For abnormal images, we ensure that the randomly sampled masks do not overlap with the annotated abnormal regions, and otherwise apply the same data construction procedure.
We follow the same training setup as UniDG (optimizer, resolution, and learning-rate schedule), and the total training iterations are 2.4M.

Why not directly retrieve the original normal images as the normal counterparts of abnormal images?
Because the original anomaly detection datasets may not provide perfectly matched normal images; differences in regions outside the mask can significantly degrade the quality of paired training data.

\begin{figure*}[!t]
    \centering
    \begin{minipage}{0.48\textwidth}
        \centering
        \includegraphics[width=\textwidth]{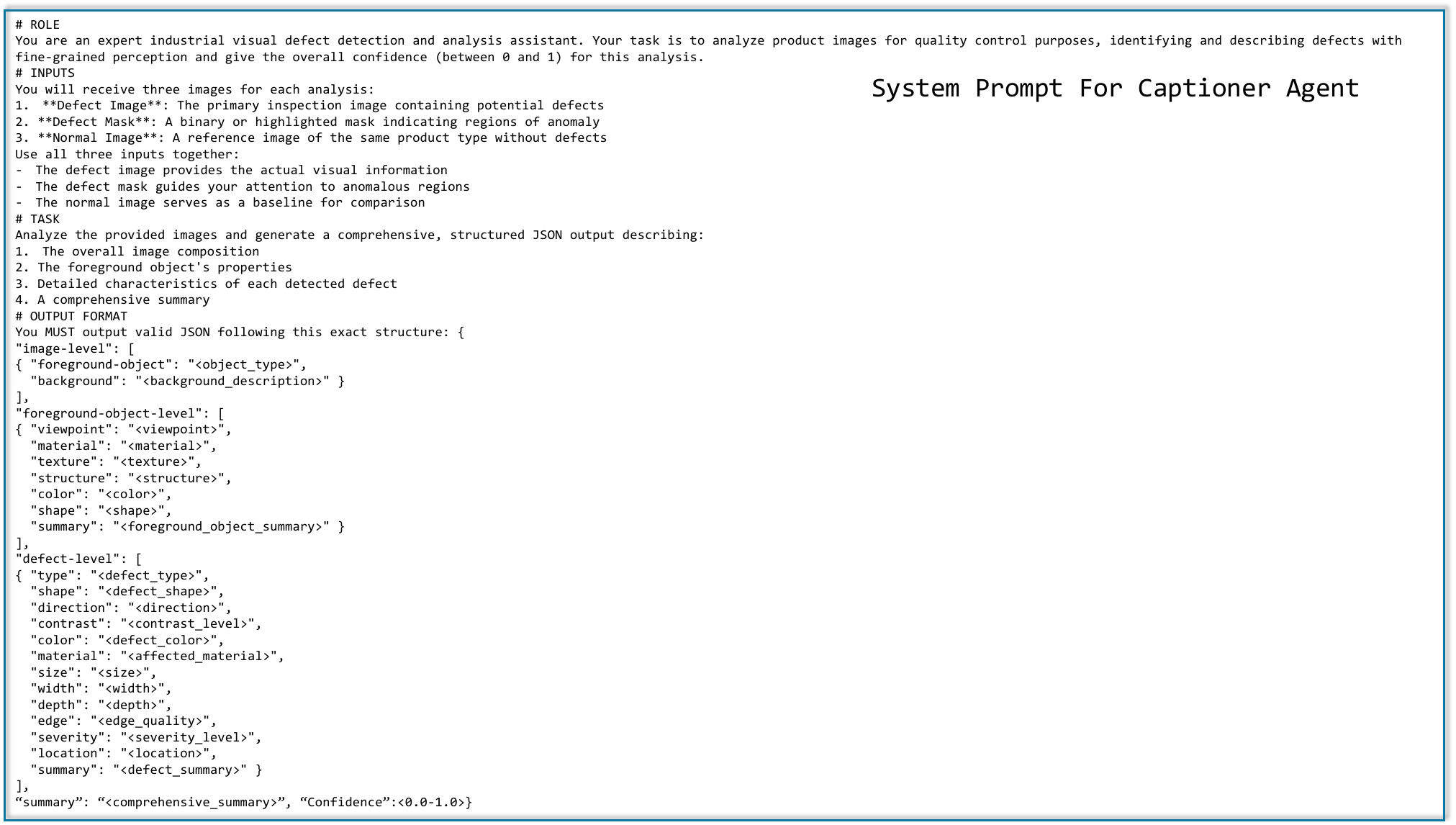}
        \caption{The system prompt for the Captioner Agent.}
        \label{fig:captioner_agent_system_prompt}
    \end{minipage}
    \hfill
    \begin{minipage}{0.48\textwidth}
        \centering
        \includegraphics[width=\textwidth]{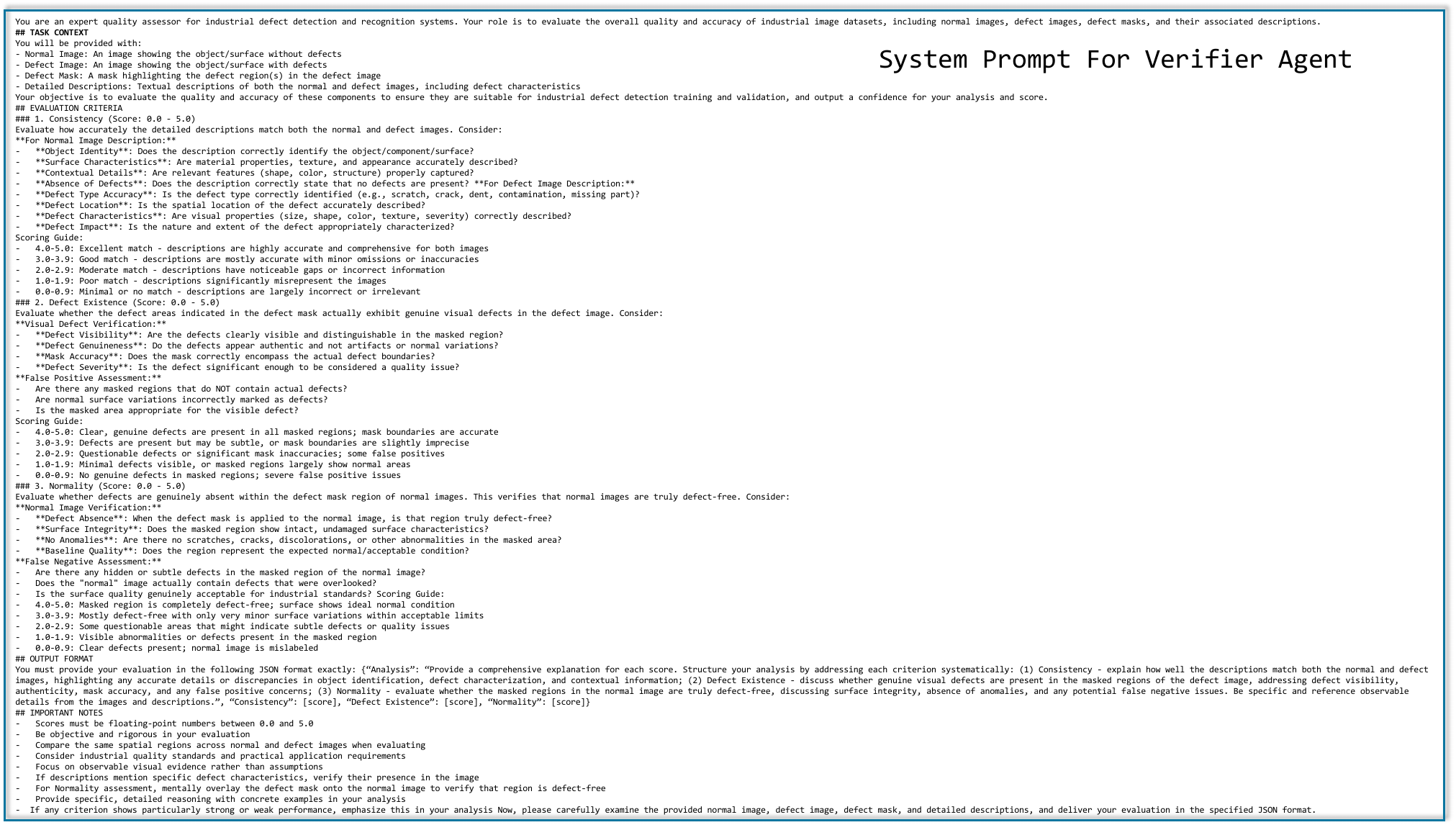}
        \caption{The system prompt for the Verifier Agent.}
        \label{fig:verifier_agent_system_prompt}
    \end{minipage}
\end{figure*}

\textbf{Captioner Agent.}
The Captioner Agent produces structured, fine-grained descriptions conditioned on the abnormal image, the abnormal mask, and the synthesized normal counterpart.
Inspired by FineGrainedAD~\cite{fan2026fgad}, we design a system prompt that guides the agent to describe defects from global to local perception, including image-level, foreground object-level, and defect-level information.
Since defects are typically located on foreground objects, this structured format helps capture defect distributions more reliably.
Moreover, providing the additional normal counterpart helps the agent better localize and describe the abnormal region.
The Captioner Agent integrates multiple advanced proprietary and open-source MLLMs, including proprietary models GPT-5.1~(\ie, gpt-5.1-2025-11-13) and Gemini-3-Pro~(\ie, gemini-3-pro-preview), as well as open-source models Qwen3-VL-235B-A22B-instruct and GLM-4.6V.
As shown in Fig.~\ref{fig:captioner_agent_system_prompt}, the prompt requires the agent to output both the structured description and a confidence score.

\textbf{Verifier Agent.}
The Verifier Agent filters low-quality data based on the abnormal image, abnormal mask, normal counterpart, and the candidate structured descriptions.
For each normal-abnormal-mask triplet, the Captioner Agent generates multiple candidate descriptions from different models, each associated with a confidence score in \([0,1]\).
Candidates with confidence below 0.8 are discarded upfront to reduce subsequent token consumption.
The remaining candidates are then fed into the Verifier Agent, which selects the most accurate description according to a predefined system prompt (Fig.~\ref{fig:verifier_agent_system_prompt}).
Specifically, if the average score of Verifier Agent's response to the candidate description is below 4.2, it is considered inaccurate.
If all remaining candidates are deemed inaccurate, the triplet is discarded; otherwise, we retain at most one final description~(\ie, the highest confidence candidate) for each triplet.

After this process, we obtain the real part of the \textit{\textbf{UDG}} dataset, containing 150K normal-abnormal-mask-caption quadruplets.
Subsequently, we construct the synthesized part as follows.
For each dataset, we take real normal images, extract foreground object regions with SAM, and retrieve mask templates that match the defect category and foreground region as synthesis targets.
The mask template repository is constructed by clustering the masks of the real part of the \textit{\textbf{UDG}} dataset into 100 clusters for each mapped defect category, resulting in 2800 templates across 28 categories.
During retrieval, we randomly select one mask from the cluster that is located within the foreground object region.
We then use \textbf{UniDG-SFT-Real}~(trained only once on the real part of \textit{\textbf{UDG}} with the same setup as \textbf{UniDG-SFT}) to synthesize defects conditioned on the retrieved masks and normal images.
After running the Captioner Agent and Verifier Agent again to generate descriptions and filter low-quality quadruplets, we obtain an additional 150K synthesized normal-abnormal-mask-caption quadruplets.
Finally, we combine the real and synthesized parts to form the full \textit{\textbf{UDG}} dataset with 300K samples.

\subsection{Data Distribution and Visual Cases of the \textit{\textbf{UDG}} Dataset}
In Sec.~3.2 of the main paper, we summarize the data distribution of the \textit{\textbf{UDG}} dataset, which spans industrial, natural, and medical scenarios.
\textit{\textbf{UDG}} contains 300K samples covering 269 original defect categories; we manually map these categories into 28 standardized categories, listed below in descending order by frequency:
missing, combined, deformation, discoloration, breakage, dirt, scratch, dehiscence, bruise, raised, foreign matter, abrasion, indentation, unknown, hole, black spot, fold, misprint, wrinkle, graze,
watermark, scraped, glue, knife grain, rust, bubble, unclear, bright line.

\begin{figure}[!t]
    \centering
    \includegraphics[width=0.98\textwidth]{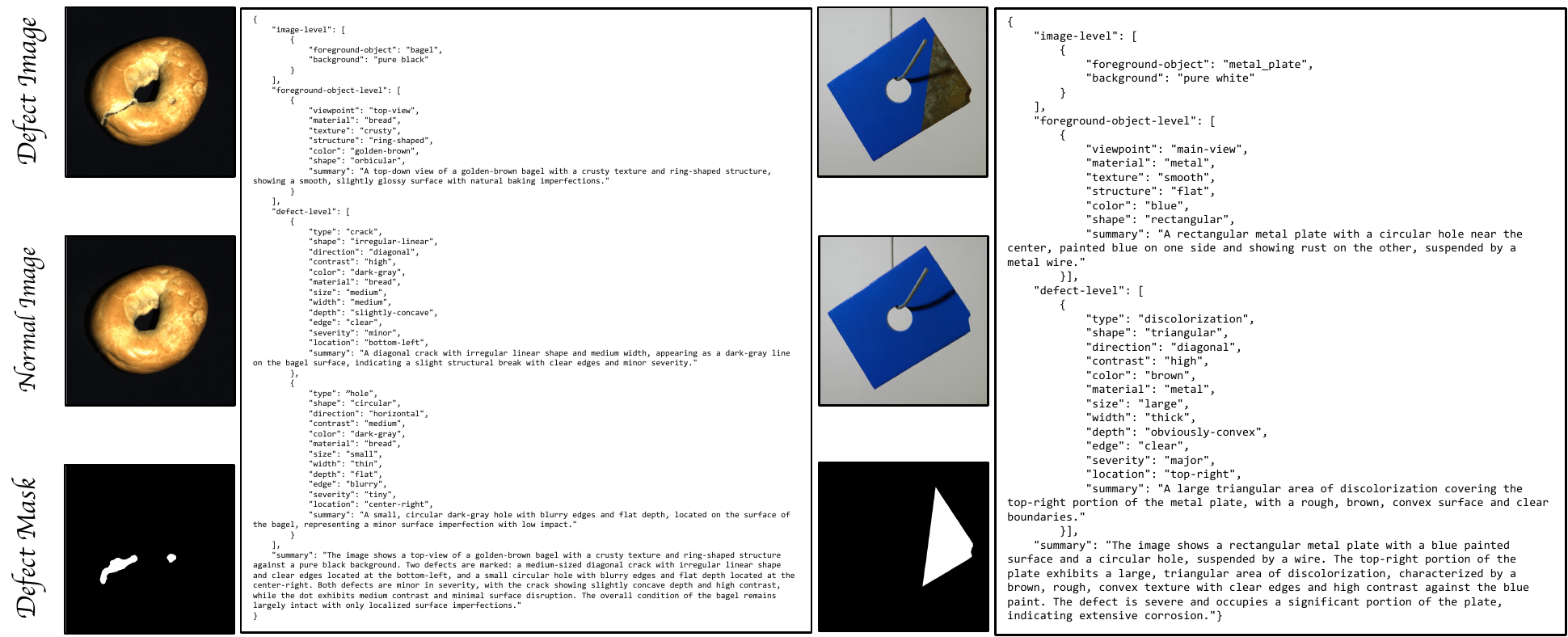}
    \caption{Examples of abnormal-normal-mask-caption quadruplets in the \textit{\textbf{UDG}} dataset.}
    \label{fig:dataset_visual_cases}
\end{figure}

Partial visualizations of \textit{\textbf{UDG}} are shown in Fig.~\ref{fig:dataset_visual_cases}.
The synthesized normal counterparts appear realistic, and the structured fine-grained descriptions precisely characterize the defects.
Such descriptions can serve as contextual signals and facilitate the performance of other MLLM-based anomaly detection models.
We further conduct additional ablation studies on the performance improvement with the descriptions for anomaly classification and detection in Tab.~\ref{tab:mmad_results}.

\section{Details of UniDG Models}
\subsection{Adaptive Defect Crop Strategy}
\label{subsec:app_adaptive_defect_crop}
In Sec.~5.2, we introduce that Defect-Context Editing comprises an Adaptive Defect Crop strategy, a diptych input format, and the multimodal attention mechanism in MM-DiT.
Due to space constraints, the main paper only describes the diptych input and multimodal attention; here we provide additional details of the Adaptive Defect Crop strategy.

As shown in Alg.~\ref{alg:adaptive_defect_crop}, during training and inference, if the reference defect occupies only a small region of the reference image, high-frequency defect details can be easily lost.
The Adaptive Defect Crop strategy first localizes the defect region using the mask, and then crops the defect region with an appropriate amount of surrounding context.
This increases the proportion of defect information while reducing irrelevant background noise.

The crop ratio $r$ also affects synthesis quality.
In our implementation, $r$ controls the \emph{inverse} expansion in Alg.~\ref{alg:adaptive_defect_crop}: \texttt{ExpandBBox}$(\cdot, r)$ expands the bounding box such that the expanded area is scaled by $1/r$ (equivalently, the side length is scaled by $\sqrt{1/r}$).
Therefore, smaller $r$ leads to larger expansion and includes more surrounding context, while larger $r$ results in less expansion.
Importantly, the bounding box is expanded \emph{before} slicing the reference image, so surrounding pixels remain available.

We adjust $r$ based on the defect size.
Specifically, let $s$ denote the area ratio of the reference defect region to the reference image, and let $T$ be a predefined threshold.
When $s \ge T$, we use the defect region without expansion~(\ie, $r = 1$).
When $s < T$, we use a smaller $r$ (thus a larger expansion factor $1/r$) to include more context for small defects.
The specific formula is:
\begin{equation}
r = \begin{cases}\beta+\frac{1-\beta}{T} s, & \text { if } s<T, \\ 1, & \text { if } s \geq T,\end{cases}
\end{equation}
where $s$ is the defect area ratio, $T$ is a predefined threshold, and $\beta$ is the minimum value of $r$ when $s = 0$.
In the default settings of UniDG, we set $\beta = 0.6$ and $T = 0.1$, so $r \in [0.6, 1]$ and the corresponding expansion factor $1/r \in [1, 1/0.6]$.

\begin{algorithm}[t]
\caption{Adaptive Defect Crop Strategy and Diptych Input Construction}
\label{alg:adaptive_defect_crop}
\begin{algorithmic}[1]
\REQUIRE Reference image $I_{\text{ref}}$, reference mask $M_{\text{ref}}$, source image $I_{\text{src}}$, target mask $M_{\text{src}}$, size $S$
\ENSURE Diptych triplet $(D_{\text{src}}, D_{\text{res}}, D_{\text{mask}})$

\STATE \textbf{// Reference Processing}
\STATE $(y_1, y_2, x_1, x_2) \gets \text{GetBBox}(M_{\text{ref}})$
\STATE $I_{\text{ref}}^{\text{masked}} \gets I_{\text{ref}} \odot M_{\text{ref}} + 255 \cdot (1 - M_{\text{ref}})$

\STATE \textbf{// Adaptive Expansion}
\STATE $s \gets \text{Area}(M_{\text{ref}}) / \text{Area}(I_{\text{ref}})$ \COMMENT{Defect area ratio}
\IF{$s < T$}
    \STATE $r \gets \beta + \frac{1-\beta}{T} \cdot s$
\ELSE
    \STATE $r \gets 1$
\ENDIF
\STATE $(y'_1, y'_2, x'_1, x'_2) \gets \text{ExpandBBox}((y_1,y_2,x_1,x_2), r, I_{\text{ref}})$
\STATE $I_{\text{ref}}^{\text{crop}} \gets I_{\text{ref}}^{\text{masked}}[y'_1:y'_2, x'_1:x'_2]$
\STATE $I_{\text{ref}}^{\text{final}} \gets \text{Resize}(\text{PadToSquare}(I_{\text{ref}}^{\text{crop}}), S)$

\STATE \textbf{// Target Mask Dilation}
\STATE $M_{\text{src}} \gets \text{Dilate}(M_{\text{src}} \times 255, \mathbf{1}_{7\times7}, 2)$

\STATE \textbf{// Source Image Processing}
\STATE $I_{\text{src}}^{\text{masked}} \gets \text{Resize}(\text{PadToSquare}(I_{\text{src}} \odot \neg M_{\text{src}}), S)$
\STATE $I_{\text{src}}^{\text{final}} \gets \text{Resize}(\text{PadToSquare}(I_{\text{src}}), S)$
\STATE $M_{\text{src}}^{\text{final}} \gets \text{Resize}(\text{PadToSquare}(M_{\text{src}}), S)$

\STATE \textbf{// Diptych Construction}
\STATE $D_{\text{mask}} \gets [\mathbf{0}_{S}, M_{\text{src}}^{\text{final}}]$
\STATE $D_{\text{src}} \gets [I_{\text{ref}}^{\text{final}}, I_{\text{src}}^{\text{masked}}]$
\STATE $D_{\text{res}} \gets [I_{\text{ref}}^{\text{final}}, I_{\text{src}}^{\text{final}}]$

\STATE \textbf{return} $(D_{\text{src}}, D_{\text{res}}, D_{\text{mask}})$
\end{algorithmic}
\end{algorithm}

\subsection{Model Structure and Parameters Analysis}
\label{subsec:app_model_structure}
Sec.~5.2 describes UniDG's MM-DiT backbone, where the diptych input is encoded by a VAE encoder into latent features.
We also use the SigLIP image encoder to extract reference-defect features for fusion with the target region.
Here, we summarize the model structure and parameters.

As shown in Tab.~\ref{tab:model_structure_params}, the adopted FLUX.1-Fill-dev consists of four components:
1) VAE Encoder\&Decoder, 2) CLIP Text Encoder, 3) T5-XXL Text Encoder, and 4) MM-DiT.
The VAE encoder compresses the dimensionality of MM-DiT latent features to reduce computational cost, following the design of LDM~\cite{rombach2022ldm}.
The CLIP and T5-XXL text encoders are used to encode text instructions; they are not activated for reference-based defect synthesis, and are only required for the instruction-based defect synthesis strategy described below.
MM-DiT is a 12B-parameter rectified-flow transformer with a velocity-prediction objective; it processes diverse conditioning signals through multimodal attention and iteratively generates high-quality target images.

For reference-feature extraction, the adopted FLUX.1-Redux-dev includes two components:
1) SigLIP Image Encoder and 2) SigLIP Image Embedder.
The SigLIP image encoder extracts high-level semantic information from the reference defect image, and the image embedder (an MLP projection) maps the extracted features into the feature space perceived by MM-DiT.

\begin{table}[!t]
\centering
\caption{Model Structure and Parameters. Enabled means that this component is enabled in the reference defect-based synthesis or text instruction-based synthesis. }
\label{tab:model_structure_params}
\begin{tabular}{llcc}
\toprule
\textbf{Component} & \textbf{Parameters} & \textbf{Enabled} & \textbf{Status} \\
\midrule
\multicolumn{4}{l}{\textit{FLUX.1-Fill-dev}} \\
\quad CLIP Text Encoder & $\sim$125M & \faTimes/\faCheck & \faSnowflake \\
\quad T5 XXL Text Encoder & $\sim$4.7B & \faTimes/\faCheck & \faSnowflake \\
\quad FLUX-VAE & $\sim$84M & \faCheck/\faCheck & \faSnowflake \\
\quad MM-DiT & $\sim$12B & \faCheck/\faCheck & \faFire \& LoRA \\
\midrule
\multicolumn{4}{l}{\textit{FLUX.1-Redux-dev}} \\
\quad SigLIP Image Encoder & $\sim$400M & \faCheck/\faCheck & \faSnowflake \\
\quad Image Embedder & $\sim$4.7M & \faCheck/\faCheck & \faSnowflake \\
\bottomrule
\end{tabular}
\end{table}

\subsection{Instruction Editing Capabilities of UniDG}
\label{subsec:app_instruction_editing}
Due to space constraints, the main paper focuses on reference-based defect synthesis.
In this section, we introduce UniDG's instruction-based defect editing capability.

In the reference-based defect synthesis framework, we use (i) VAE-encoded features to preserve reference defect textures and (ii) SigLIP-encoded features to capture high-level semantics, enabling high-fidelity defect synthesis.
To reduce the computational overhead during training and inference, we zero-pad the text-branch features of MM-DiT; therefore, the two text encoders do not need to be loaded for reference-based synthesis.

In the text instruction-based defect synthesis framework, we load two text encoders and encode the instruction (e.g., ``transfer the scratch from scene 1 to the specified area of scene 2'') as the text-branch features for MM-DiT.
Meanwhile, the instruction-editing branch optionally takes a reference defect image. Accordingly, we consider two instruction forms.
When a reference defect image is provided, a typical instruction is ``Migrate the defect from the reference image to the target scene.''
When no reference image is given, the instruction becomes ``Add the \{multiple defect attributes from caption\} \texttt{defect\_category} defect to the image,'' expecting the model to directly understand the specified defect category.
When no reference defect image is provided, we replace the left side of the diptych input with a pure-black image to maintain compatibility with existing models, and we zero-pad the reference-defect features from the SigLIP image encoder.

We do not directly fine-tune instruction-based defect synthesis from pre-trained models designed for natural-image editing.
Instead, we fine-tune on the two aforementioned data distributions (with and without a reference defect image) starting from the \textbf{UniDG-SFT} checkpoint.
All training data are constructed from the \textit{\textbf{UDG}} dataset, and the editing instructions are derived from the captions in the quadruplets.

\begin{figure*}[!t]
    \centering
    \includegraphics[width=0.98\textwidth]{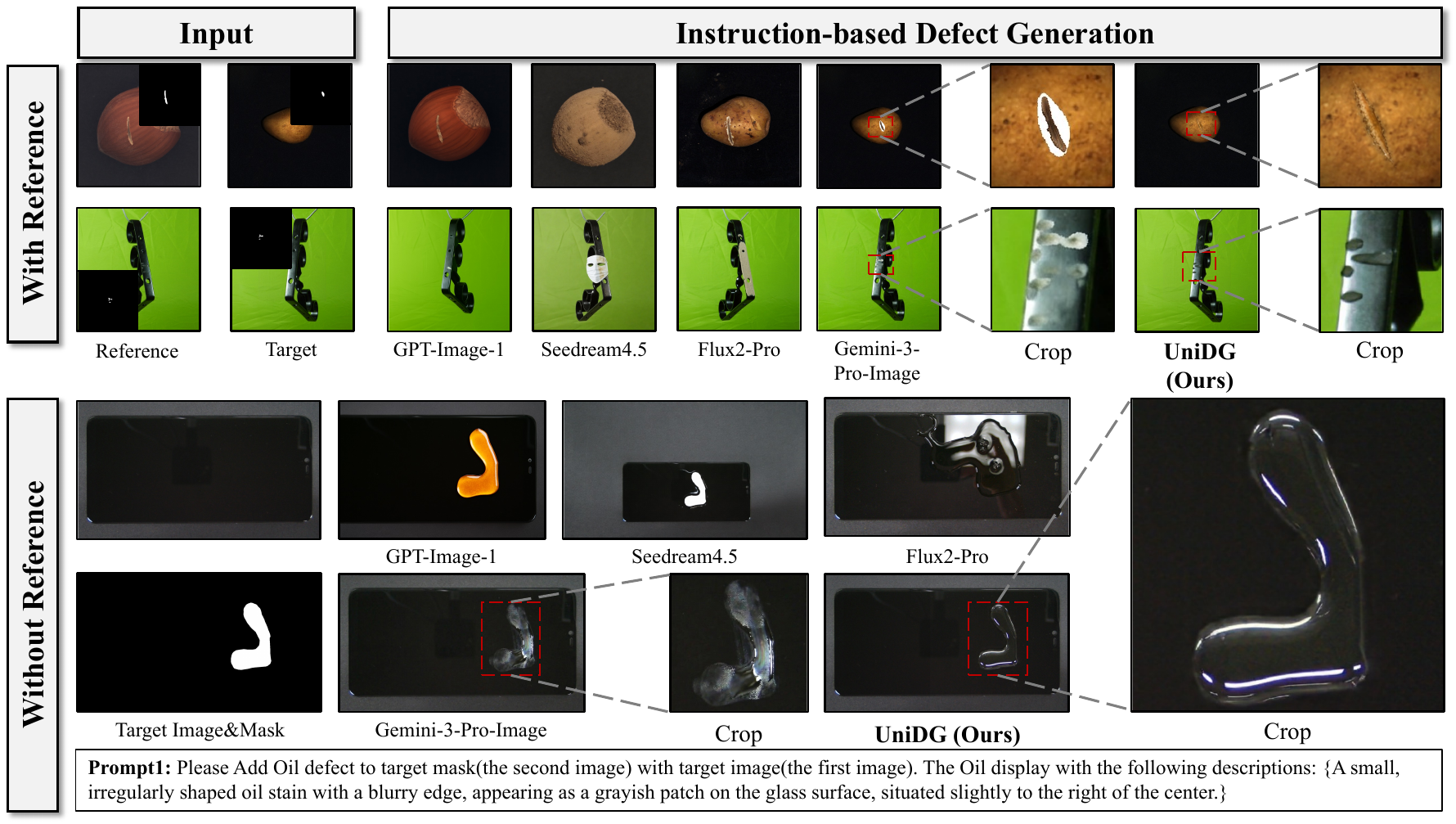}
    \caption{Qualitative examples of UniDG-Text and proprietary instruction-editing models. This figure is illustrative and does not support a superiority claim.}
    \label{fig:instruction_visualization}
\end{figure*}

Fig.~\ref{fig:instruction_visualization} shows qualitative examples of UniDG-Text alongside proprietary instruction-editing models.
We do not claim superiority over these models: the comparison is selected and qualitative, with no quantitative benchmark or systematic human study.
For the proprietary editors we use the prompt: ``Please add the \texttt{defect\_category} defect from the reference mask (the second image) within the reference image (the first image) to the target mask (the fourth image) within the target image (the third image).''
UniDG-Text only shows that Defect-Context Editing can be conditioned on text instructions.

\subsection{Training-Free Zero-Shot Pipeline}
\label{subsec:app_training_free_zero_shot}
In the experiments and visualizations in the main paper, UniDG adopts a training-free few-shot~(\ie, one-shot) setting for fair comparisons with other FSAG methods.
Specifically, the training-free one-shot strategy selects four defect samples per category and randomly chooses one as the reference defect during inference.
UniDG can also support zero-shot inference by following Anomagic~\cite{jiang2025anomagic}, \ie, pre-building an offline repository for each defect category and dynamically retrieving a same-category reference image during inference.
This zero-shot variant further improves the flexibility of UniDG's image-insertion framework and enhances its generality for downstream applications.

\section{Details of Diversity-SFT}

\begin{algorithm}[t]
\caption{Candidate Samples Quality Assessment}
\label{alg:mask_quality_calculation}
\begin{algorithmic}[1]
\REQUIRE Candidate reference image $I_{\text{ref}}$, reference mask $M_{\text{ref}}$
\REQUIRE Target image $I_{\text{tar}}$, target mask $M_{\text{tar}}$
\ENSURE Quality score $Q_{\text{score}}$

\STATE \textbf{// Area Ratio of Reference Mask}
\STATE $r_{\text{area}} \gets |\{p : M_{\text{ref}}(p) > 0\}| \,/\, (H \times W)$

\STATE \textbf{// Connectivity Score}
\STATE $N_{\text{cc}} \gets \text{ConnectedComponents}(M_{\text{ref}}) - 1$
\STATE $s_{\text{conn}} \gets \begin{cases} 1.0 & N_{\text{cc}} = 1 \\ 0.5 & N_{\text{cc}} \in [2,3] \\ 0.0 & \text{otherwise} \end{cases}$

\STATE \textbf{// Color Statistics in Masked Regions}
\STATE $\bar{I}_{\text{ref}}, \bar{I}_{\text{tar}} \gets \text{NormalizeTo}[0,1](I_{\text{ref}}, I_{\text{tar}})$
\STATE $\boldsymbol{\mu}_{\text{ref}} \gets \text{Mean}(\bar{I}_{\text{ref}}[M_{\text{ref}} > 0])$
\STATE $\boldsymbol{\mu}_{\text{tar}} \gets \text{Mean}(\bar{I}_{\text{tar}}[M_{\text{tar}} > 0])$

\STATE \textbf{// Color Similarity Score}
\STATE $d_{\text{color}} \gets \|\boldsymbol{\mu}_{\text{ref}} - \boldsymbol{\mu}_{\text{tar}}\|_2$
\STATE $s_{\text{color}} \gets \exp\left(-d_{\text{color}}^2 / 2\sigma^2\right)$

\STATE \textbf{// Area Score Mapping}
\STATE $s_{\text{area}} \gets \begin{cases} 1.0 & r_{\text{area}} \in [0.5\%, 1\%] \\ 0.7 & r_{\text{area}} \in (1\%, 5\%] \\ 0.5 & r_{\text{area}} \in [0.1\%, 0.5\%) \\ 0.1 & \text{otherwise} \end{cases}$

\STATE \textbf{// Final Quality Score}
\STATE $Q_{\text{score}} \gets \alpha \cdot s_{\text{area}} + \beta \cdot s_{\text{conn}} + \gamma \cdot s_{\text{color}}$

\STATE \textbf{return} $Q_{\text{score}}$
\end{algorithmic}
\end{algorithm}

\subsection{SFT Training Data Sampling}
\label{subsec:app_diversity_sft}
In Sec.~5.3 of the main paper, we introduce three training data distributions used in \emph{Diversity-SFT}:
(a) \textbf{Single Object Defect-Fill Sampling},
(b) \textbf{Single Object Intra-Defect Cross-Sampling}, and
(c) \textbf{Multi-Object Intra-Defect Cross-Sampling}.
In the following, we detail how these distributions are sampled from the \textit{\textbf{UDG}} dataset.

\textbf{(a) Single Object Defect-Fill Sampling.}
This distribution encourages the model to synthesize defects that are consistent with the reference defect category and identity features within the same object.
For each quadruplet in \textit{\textbf{UDG}}, we use the abnormal image as both the reference defect and the target defect, while using its normal counterpart as the source image (\ie, the right half of the diptych input).
Because we apply the Adaptive Defect Crop strategy, the left half of the diptych (reference side) is not identical to the right half, which mitigates trivial copying and discourages lazy learning.

\textbf{(b) Single Object Intra-Defect Cross-Sampling.}
This distribution improves diversity by learning to synthesize defects that match the reference defect category while exhibiting different identity features.
We first select one quadruplet and take its abnormal/normal images as the target defect and the source image.
We then retrieve another quadruplet with the same object type and defect type but different defect identity features as the reference defect.
The reference selection is based on mask quality and feature similarity between the target defect and the candidate reference defects; the detailed retrieval algorithm is provided in Alg.~\ref{alg:mask_quality_calculation}.
Specifically, the Connectivity Score filters masks that are overly disconnected, and the Area Score filters masks that are excessively large or small, both of which are not conducive to stable learning.
The Color Score measures color similarity between the target and reference defect regions after normalizing image tensors to $[0,1]$, where higher scores indicate higher similarity.
We set $\sigma=0.1$ for the calculation of the $s_\text{color}$.
We set the coefficients for the area, connectivity, and color scores to 0.3, 0.4, and 0.3, respectively.
Finally, we randomly choose the reference defect from the top-3 candidates ranked by the overall quality score.

\textbf{(c) Multi-Object Intra-Defect Cross-Sampling.}
This distribution enables cross-object defect synthesis by learning to transfer defects of the same category across different object types.
Since we map the original 269 defect categories into 28 standardized categories in \textit{\textbf{UDG}}, we select one quadruplet as the target (abnormal/normal images) and retrieve a reference quadruplet from a different object type but with the same mapped defect category.

We traverse \textit{\textbf{UDG}} only once to construct the training samples.
Within each data distribution, each quadruplet is used at most once as a reference defect and once as a target defect.
Overall, these three distributions substantially improve generation diversity, defect category consistency, and cross-product defect synthesis capability for our SFT model.

\subsection{Defect Enhancement Mechanism}
\label{subsec:app_defect_enhancement}

We introduce a defect enhancement mechanism to improve the visual fidelity and discriminability of synthesized defects, and validate its effectiveness in the ablation study.
This mechanism contains two components: (1) normal regularization loss and (2) defect attention.

\textbf{Normal regularization loss.}
The normal regularization loss encourages the synthesized target region to deviate from the normal counterpart in feature space, which helps counteract the inpainting prior of the pre-trained Fill model while maintaining reliable defect synthesis.
The loss is defined as:
\begin{equation}
\mathcal{L}_{\text{reg-cos}} = \frac{1}{|B|} \sum_{b=1}^{|B|} \text{ReLU}\left( \bar{s}_b - \tau \right),
\end{equation}
where $\bar{s}_b = \frac{1}{|\mathcal{M}_b|} \sum_{i \in \mathcal{M}_b} \cos(\mathbf{z}_{\text{target}}^{(i)}, \mathbf{z}_{\text{orig}}^{(i)})$, $\tau$ denotes the similarity margin (set to 0.5), and $B$ is the batch size.
Here, $\mathcal{M}_b$ contains only latent tokens inside the target mask.
$\mathbf{z}_{\text{target}}^{(i)}$ denotes the denoised latent representation of the synthesized target region, and $\mathbf{z}_{\text{orig}}^{(i)}$ denotes the VAE latent representation of the normal counterpart at the same masked location.

\textbf{Defect attention.}
The defect attention mechanism enhances fidelity by explicitly prioritizing information from the reference defect and the target region during generation.
Concretely, we bias the multimodal attention map toward the reference defect features (left side) and target-region features (right side).
Given the standard scaled dot-product attention:
\begin{equation}
\text{Attention}(\mathbf{Q}, \mathbf{K}, \mathbf{V}) = \text{softmax}\left( \frac{\mathbf{Q}\mathbf{K}^\top}{\sqrt{d_k}} \right) \mathbf{V}
\end{equation}

We construct a mask indicator set $\mathcal{M}$ containing the spatial indices of both the reference defect subjects (left side) and the target region (right side) in the latent space. An attention bias matrix $\mathbf{B} \in \mathbb{R}^{N \times N}$ is then defined as:
\begin{equation}
\mathbf{B}_{ij} =
\begin{cases}
\log \alpha, & \text{if } j \in \mathcal{M} \\
0, & \text{otherwise}
\end{cases}
\end{equation}
where $\alpha > 1$ is the attention amplification factor (default $\alpha = 2.0$).

The modified attention mechanism becomes:
\begin{equation}
\text{Attention}_{\text{weighted}}(\mathbf{Q}, \mathbf{K}, \mathbf{V}) = \text{softmax}\left( \frac{\mathbf{Q}\mathbf{K}^\top}{\sqrt{d_k}} + \mathbf{B} \right) \mathbf{V}
\end{equation}

This formulation amplifies attention weights for keys in the masked regions, guiding the model to emphasize these critical areas during generation.
Overall, these two strategies slightly improve both synthesis quality and downstream performance.

\subsection{Failure Case Visualizations}
\label{subsec:app_failure_cases_sft}
\begin{figure*}[!t]
    \centering
    \includegraphics[width=0.98\textwidth]{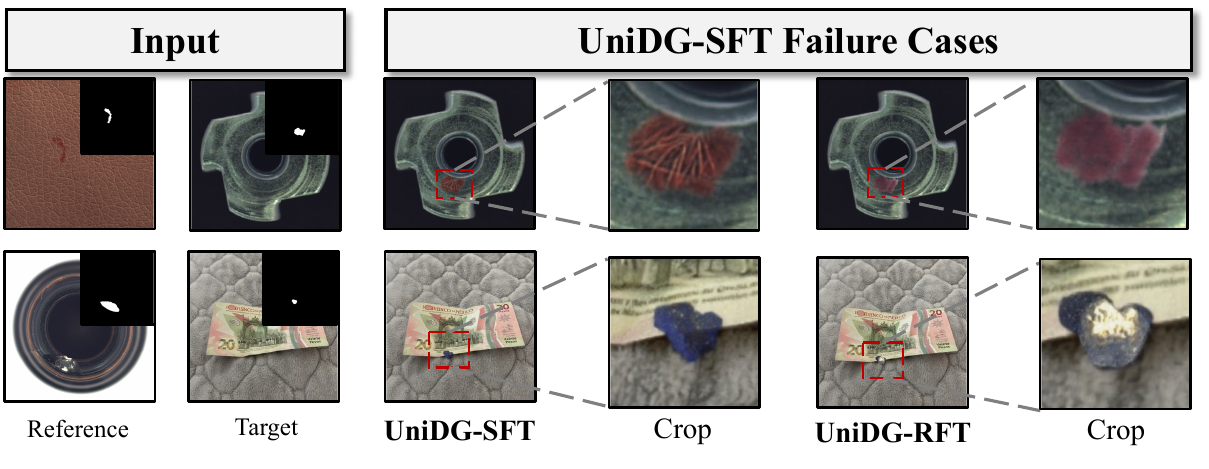}
    \caption{The failure cases of UniDG-SFT.}
    \label{fig:failure_cases_sft}
\end{figure*}

In Sec.~5.3 of the main paper, we note that \textbf{UniDG-SFT} achieves strong diversity, synthesizing high-quality and category-adherent defects across scenarios.
However, due to the diverse training distributions, the synthesized defects may occasionally deviate from the reference in certain attributes, such as color inconsistencies or changes in texture and lighting.
Although these discrepancies do not affect downstream performance, they indicate room for improvement in reference-consistent synthesis.
Representative failure cases of \textbf{UniDG-SFT} are shown in Fig.~\ref{fig:failure_cases_sft}.

\begin{figure*}[!t]
    \centering
    \includegraphics[width=0.98\textwidth]{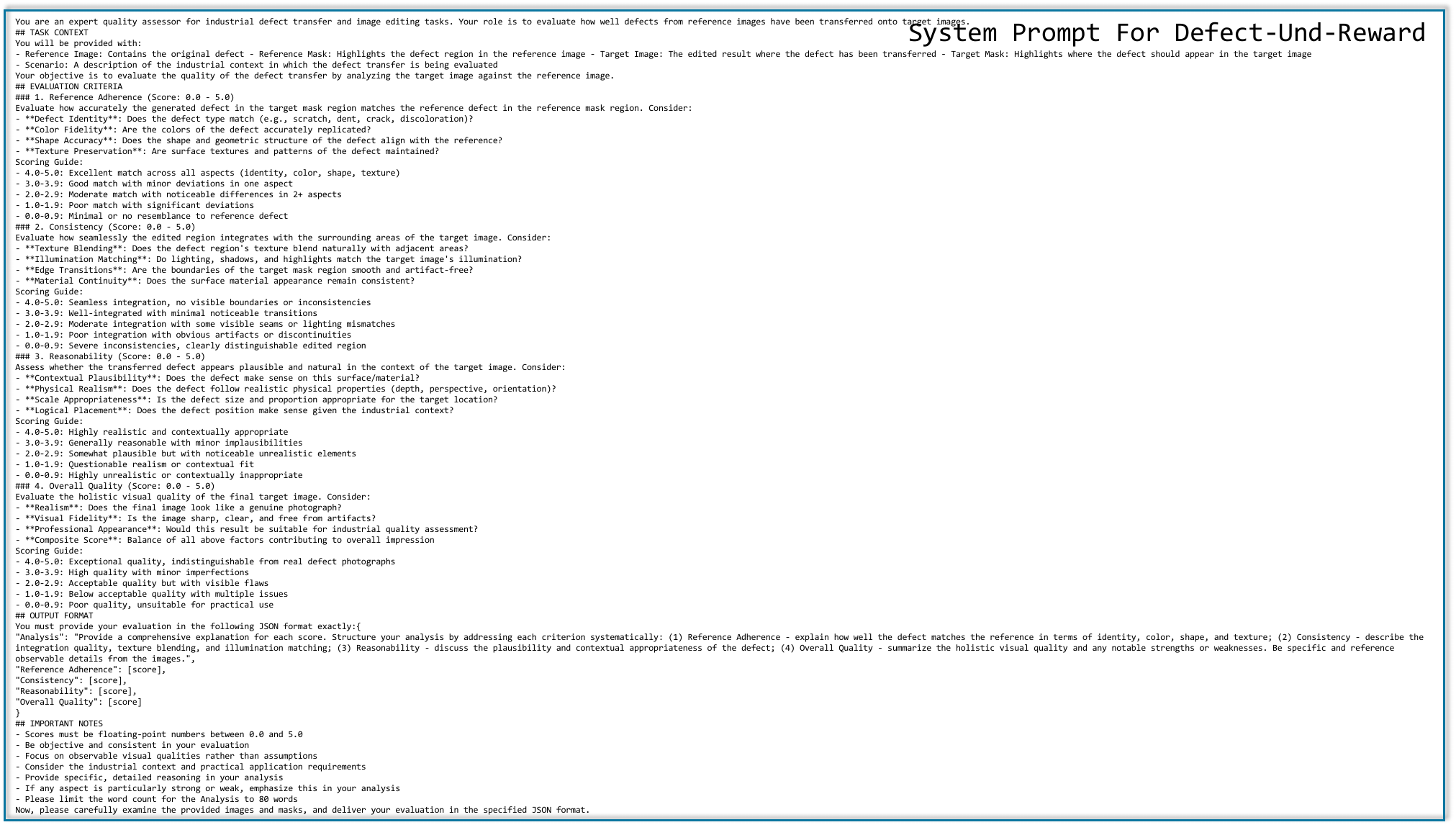}
    \caption{The system prompt for Defect-Und-Reward and MLLM-based comprehensive scores.}
    \label{fig:defect_und_reward_prompt}
\end{figure*}

\section{Details of Consistency-RFT}
\label{subsec:app_consistency_rft}
The main paper summarizes Consistency-RFT.
Here, we provide additional details on the reward models and online RFT framework.

\subsection{Reward Models}
We employ two reward models: Defect-Und-Reward and Defect-Recog-Reward.
Defect-Und-Reward leverages the multimodal perception capabilities of MLLMs to evaluate reference adherence, target-background consistency, the reasonability of defect occurrence, and overall generation quality.
Defect-Recog-Reward trains a unified defect classification and localization model using the large-scale \textit{\textbf{UDG}} dataset; it identifies the mapped defect category in generated images and localizes the corresponding regions (\ie, defect pseudo masks).

\textbf{Defect-Und-Reward.}
Defect-Und-Reward is built on the lightweight open-source model Qwen3VL-4B-Instruct~\cite{bai2025qwen3vl}.
Its training corpus comprises approximately 15K edited pairs (reference defect image, reference mask, target defect image, target mask) sampled from \textit{\textbf{UDG}}.
With a designed system prompt, proprietary models (including Gemini-3-Pro~(\ie, gemini-3-pro-preview) and GPT-5.1~(\ie, gpt-5.1-2025-11-13)) analyze and score each edited pair from four perspectives (\ie, reference adherence, consistency, reasonability, and overall quality).
The edited pairs, analysis, and scores are then used as supervision to train this reward model.
The detailed system prompt for Defect-Und-Reward is shown in Fig.~\ref{fig:defect_und_reward_prompt}.
The training pipeline is implemented with LLaMA-Factory~\cite{zheng2024llamafactory} using LoRA (rank 64), a learning rate of $1 \times 10^{-4}$, 20 epochs, and a cosine learning-rate scheduler.
We train both 4B and 8B variants with the same data, prompt, and LoRA setup; the 8B variant is reported only for reward-model benchmarking, while UniDG-RFT uses the 4B Defect-Und-Reward model for efficiency.

To evaluate its effectiveness, we manually construct UDG-Reward-Bench, which contains 500 edited pairs annotated by dozens of annotators.
Each pair is provided with detailed analyses and scores across the same four dimensions in Fig.~\ref{fig:defect_und_reward_prompt}.
An advanced MLLM (\ie, Gemini-3-Pro~\cite{comanici2025gemini}) then summarizes the annotations for each pair by discarding the highest and lowest scores.
Defect-Und-Reward achieves performance comparable to proprietary models on UDG-Reward-Bench, demonstrating its effectiveness and robustness.

\textbf{Defect-Recog-Reward.}
Defect-Recog-Reward trains a unified defect classification model and a defect instance segmentation model using the full training images and mapped defect categories from \textit{\textbf{UDG}}.
The classification model is built on a ResNet-34 backbone and predicts the defect category (28 defect types plus a normal category).
If the predicted category matches the ground truth, the score is 1.0; otherwise, it is 0.

The instance segmentation model is built on ViT-B/16 and predicts defect instance masks for input images.
Higher scores are assigned when the predicted masks better match the ground-truth masks.
The scoring is based on three metrics: pixel AUROC, pixel AP, and pixel PRO.
We normalize these metrics, average them with the category score, and use the resulting composite score as the overall evaluation for the current image; this robust preference signal guides the Online-RL process.
We train Defect-Recog-Reward for 5 epochs with a batch size of 128, a learning rate of $1 \times 10^{-4}$, and a cosine learning-rate scheduler.
The proposed defect instance segmentation model achieves an average score exceeding 0.92 on datasets such as MVTec-AD and VisA that are not included in the training samples, demonstrating the validity of its preference measurement.

\subsection{Consistency-RFT Framework}
\label{subsec:app_consistency_rft_framework}
Consistency-RFT builds on FLOW-GRPO~\cite{liu2025flowgrpo}, as summarized in Alg.~\ref{alg:flow_grpo_unidg}.
Inspired in part by Consistent-RFT~\cite{tan2026consistentrft}, we adopt Group Relative Policy Optimization (GRPO) to optimize the generation model with reward signals.
For each defect generation task, we sample a group of $G$ generated images $\{I_{\text{tar}}^{(i)}\}_{i=1}^G$ and compute rewards $\{r^{(i)}\}_{i=1}^G$.
Defect-Und-Reward outputs four scores (reference adherence, consistency, reasonability, and quality), which we average to obtain $r_{\text{gen}}$.
The final reward is a weighted average of $r_{\text{gen}}$ and the Defect-Recog-Reward score $r_{\text{det}}$.
Here, $\tilde{I}_{\text{ref}}$ denotes reference-defect features extracted from ${I}_{\text{ref}}$ via Adaptive Defect Crop.

\begin{algorithm}[t]
\caption{Flow-GRPO for UniDG}
\label{alg:flow_grpo_unidg}
\begin{algorithmic}[1]
\REQUIRE UniDG-SFT model $\Phi_\theta$, Redux encoder $\mathcal{E}_{\text{redux}}$, reward models $\mathcal{R}_{\text{gen}}$, $\mathcal{R}_{\text{det}}$, dataset $\mathcal{D}$, group size $G$
\FOR{epoch $= 1$ to $N$}
    \FOR{each batch of prompts $\{(I_{\text{ref}}^j, M_{\text{ref}}^j, M_{\text{tar}}^j)\}_{j=1}^B$}
        \FOR{$j = 1$ to $B$}
            \STATE \textcolor{gray}{\textit{// Sampling Phase}}
            \FOR{$i = 1$ to $G$}
                \STATE Extract features: $(\mathbf{e}_{\text{img}}^j, \mathbf{e}_{\text{pool}}^j) \leftarrow \mathcal{E}_{\text{redux}}(\tilde{I}_{\text{ref}}^j)$
                \STATE Construct diptych: $I_{\text{diptych}}^j, M_{\text{diptych}}^j$
                \STATE Generate: $I_{\text{tar}}^{(j,i)}, \{z_t^{(j,i)}\}, \{\log \pi_\theta^{(j,i)}\} \leftarrow \Phi_\theta$
            \ENDFOR
            \STATE \textcolor{gray}{\textit{// Reward Computation Phase}}
            \FOR{$i = 1$ to $G$}
                \STATE $r_{\text{gen}}^{(j,i)} \leftarrow \mathcal{R}_{\text{gen}}(I_{\text{ref}}^j, M_{\text{ref}}^j, I_{\text{tar}}^{(j,i)}, M_{\text{tar}}^j)$
                \STATE $r_{\text{det}}^{(j,i)} \leftarrow \mathcal{R}_{\text{det}}(I_{\text{tar}}^{(j,i)}, M_{\text{tar}}^j)$
                \STATE $r^{(j,i)} \leftarrow w_{\text{gen}} \cdot r_{\text{gen}}^{(j,i)} + w_{\text{det}} \cdot r_{\text{det}}^{(j,i)}$
            \ENDFOR
            \STATE \textcolor{gray}{\textit{// Prompt-wise Advantage Estimation}}
            \STATE $\mu_G^{(j)} \leftarrow \text{mean}(\{r^{(j,i)}\}_{i=1}^G)$, $\sigma_G^{(j)} \leftarrow \text{std}(\{r^{(j,i)}\}_{i=1}^G)$
            \STATE $A^{(j,i)} \leftarrow (r^{(j,i)} - \mu_G^{(j)}) / (\sigma_G^{(j)} + \epsilon_{\mathrm{adv}})$
        \ENDFOR
        \STATE \textcolor{gray}{\textit{// Policy Update Phase}}
        \FOR{inner epoch $= 1$ to $K$}
            \STATE Compute $\mathcal{L}_{\text{GRPO}}$ using Eq. (15) and (17)
            \STATE Update $\theta \leftarrow \theta - \eta \nabla_\theta \mathcal{L}$
        \ENDFOR
        \STATE Update EMA parameters
    \ENDFOR
\ENDFOR
\end{algorithmic}
\end{algorithm}

\noindent\textbf{Advantage Estimation.}
The advantage for each sample is computed relative to the group mean of the same prompt:
\begin{equation}
    A^{(j,i)} = \frac{r^{(j,i)} - \mu_G^{(j)}}{\sigma_G^{(j)} + \epsilon_{\mathrm{adv}}}
\end{equation}
where $\mu_G^{(j)} = \frac{1}{G}\sum_{i=1}^G r^{(j,i)}$ and $\sigma_G^{(j)} = \sqrt{\frac{1}{G}\sum_{i=1}^G (r^{(j,i)} - \mu_G^{(j)})^2}$ are the prompt-wise group mean and standard deviation, respectively.

\noindent\textbf{Log Probability Computation.}
For each denoising step $t$, we follow the FLOW-GRPO implementation and compute the log probability with a Gaussian transition:
\begin{equation}
    \log \pi_\theta(z_{t-1} | z_t) = -\frac{\|z_{t-1} - \mu_\theta(z_t, t)\|^2}{2\sigma_t^2} + C
\end{equation}
where $\mu_\theta(z_t, t)$ is the denoising-step mean induced by the rectified-flow velocity prediction, and $\sigma_t$ follows the default FLOW-GRPO noise schedule.

\noindent\textbf{Policy Gradient Loss.}
The GRPO loss with clipping is computed as:
\begin{equation}
    \mathcal{L}_{\text{GRPO}} = -\mathbb{E}\left[\sum_{t=1}^{T} \min\left(\rho_t A, \text{clip}(\rho_t, 1-\epsilon_{\mathrm{clip}}, 1+\epsilon_{\mathrm{clip}}) A\right)\right]
\end{equation}
where $\rho_t = \exp(\log \pi_\theta(z_{t-1}|z_t) - \log \pi_{\theta_{\text{old}}}(z_{t-1}|z_t))$ is the importance sampling ratio.

\noindent\textbf{KL Regularization.}
To prevent the policy from deviating too far from the reference model, we optionally add a KL divergence penalty:
\begin{equation}
    \mathcal{L}_{\text{KL}} = \mathbb{E}_{t}\left[\frac{\|\mu_\theta(z_t, t) - \mu_{\theta_{\text{ref}}}(z_t, t)\|^2}{2\sigma_t^2}\right]
\end{equation}

The final training objective combines the policy loss and KL regularization:
\begin{equation}
    \mathcal{L} = \mathcal{L}_{\text{GRPO}} + \beta \mathcal{L}_{\text{KL}}
\end{equation}
where $\beta$ controls the strength of the KL penalty.
We use FLUX.1-Fill-dev as the base inpainting model and FLUX.1-Redux-dev for reference defect image feature extraction. The MM-DiT is fine-tuned using LoRA with rank $r=64$ and $\alpha=128$, targeting attention and feed-forward layers.
We train with $G=8$ samples per prompt, $T=8$ denoising steps during training, and guidance scale of 3.5. The learning rate is set to $1 \times 10^{-4}$ with AdamW optimizer.
We set $w_{\text{gen}} = 0.5$, $w_{\text{det}} = 0.5$, $\beta=0.001$, $\epsilon_{\mathrm{adv}}=1\times 10^{-5}$, and $\epsilon_{\mathrm{clip}}=0.2$.
We use mixed precision training with DeepSpeed ZeRO-2 for distributed training across 8 GPUs.

\section{The Additional Ablation Studies}
\label{sec:app_ablation}

In this section, we present additional ablation studies on (i) the performance evaluation of the Defect-Und-Reward model, (ii) the iteration scaling behavior during \emph{Diversity-SFT}, and (iii) the comparison between applying only \emph{Consistency-RFT} and the full two-stage training recipe.

\noindent\textbf{The performance evaluation of Defect-Und-Reward Model.}
We evaluate defect-understanding capability on UDG-Reward-Bench, which contains 500 manually annotated editing pairs with detailed analyses and scores.
Each pair consists of a reference defect image/mask and a target defect image/mask.
We compare Defect-Und-Reward with proprietary models (\eg, Gemini-3-Pro) and open-source models (\eg, the Qwen3-VL series), where Qwen3-VL is the pre-trained backbone of Defect-Und-Reward.
The results are reported in Tab.~\ref{tab:reward_benchmark_comparison}.

The metric is computed as follows.
Under the same system prompt, each model outputs a score in \([0,5]\) for each dimension.
A prediction is considered correct if the absolute difference between the predicted score and the ground-truth score is within 0.3.
We then compute the proportion of correct predictions over 500 samples and normalize it to obtain the final accuracy score (\ie, $Acc \in [0,100]$).
Overall, proprietary models exhibit relatively small performance gaps on this benchmark, and our 4B model surpasses Qwen3-VL-32B while approaching proprietary models.

\begin{figure*}[!t]
    \centering
    \includegraphics[width=0.98\textwidth]{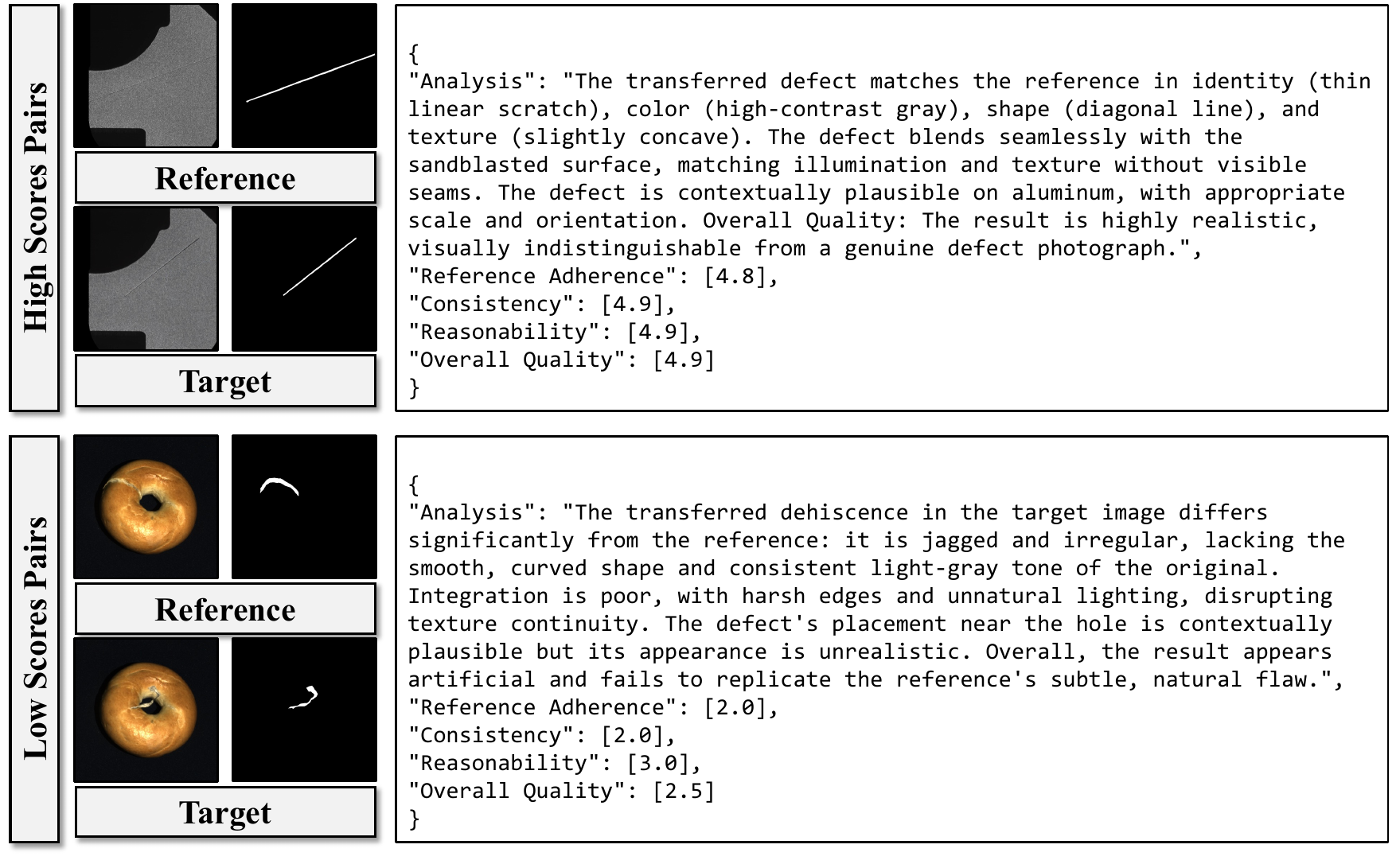}
    \caption{Examples from UDG-Reward-Bench.}
    \label{fig:app_udg_reward_bench}
\end{figure*}

We also visualize examples from UDG-Reward-Bench in Fig.~\ref{fig:app_udg_reward_bench}.

\begin{table*}[h]
\centering
\caption{Benchmark results on UDG-Reward-Bench, reporting the accuracy of prediction with advanced MLLMs on reference adherence, consistency, reasonability, overall quality.}
\label{tab:reward_benchmark_comparison}
\resizebox{0.95\textwidth}{!}{
\begin{tabular}{lcccccccc}
\toprule
\multirow{2}{*}{\textbf{Accuracy}} & \multirow{2}{*}{{GPT-4.1}} & \multirow{2}{*}{{GPT-5}} & \multirow{2}{*}{{Gemini-3-Pro}} & \multicolumn{3}{c}{{Qwen3-VL}} & \multirow{2}{*}{\textbf{Defect-Und-Reward-4B}} & \multirow{2}{*}{\textbf{Defect-Und-Reward-8B}} \\
 &  &  &  & \textbf{4B} & \textbf{8B} & \textbf{32B}   \\
\midrule
RefAd & 84.2 & 82.6 & 86.4 & 74.6 & 72.4 & 80.0 & 82.4 &  84.8  \\
Cons  & 81.6 & 83.8 & 88.8 & 75.2 & 78.4 & 79.8 & 82.0 &  86.0  \\
Reas  & 86.4 & 88.4 & 87.6 & 62.8 & 68.0 & 73.2 & 84.8 &  85.4  \\
Qual  & 92.0 & 92.4 & 94.8 & 84.2 & 86.8 & 87.4 & 91.6 &  93.4  \\
\bottomrule
\end{tabular}
}
\end{table*}

\noindent\textbf{The data scaling experiment during Diversity-SFT.}
We further study how increasing the training iterations in \emph{Diversity-SFT} affects overall performance (Tab.~\ref{tab:scaling_iterations}).
Generation quality improves consistently as training proceeds, whereas downstream performance largely saturates after 14K iterations.
The default Diversity-SFT setting corresponds to 14K iterations.
This suggests that generation quality exhibits a clearer scaling behavior, while downstream task performance converges and remains stable with additional training.
Tab.~\ref{tab:scaling_iterations} varies optimizer iterations on the full 300K set; unique-data scale at 50K/100K/200K/300K is reported separately in Sec.~\ref{sec:app_review_followup}.

\begin{figure*}[t]
    \centering
    \begin{minipage}{0.48\textwidth}
        \centering
        \captionof{table}{Scaling the training iterations improves generation quality during Diversity-SFT.}
        \label{tab:scaling_iterations}
        \resizebox{\textwidth}{!}{
            \begin{tabular}{cccccc}
            \toprule
            \multirow{2}{*}{Training Iterations} & \multicolumn{3}{c}{Multi-class Metrics} & \multicolumn{2}{c}{Synthesis Metrics} \\
            & PRO-P & mIoU & FG-mIoU & Cons & Qual \\ \midrule
            2K & 77.2 & 27.0 & 11.0 & 3.03 & 3.10 \\
            6K & 80.1 & 29.1 & 13.3 & 3.44 & 3.57 \\
            10K & 88.9 & 38.4 & 25.4 & 3.72 & 3.83 \\
            14K & 91.0 & 41.8 & 28.5 & 4.15 & 4.09 \\
            20K & 90.5 & 40.1 & 27.4 & 4.22 & 4.27 \\
            24K & 90.1 & 39.7 & 26.5 & 4.24 & 4.33 \\ \bottomrule
        \end{tabular}}
    \end{minipage}
    \hfill
    \begin{minipage}{0.48\textwidth}
        \centering
        \captionof{table}{Ablation studies on the Consistency-RFT within UniDG. -Diversity-SFT denotes only with Consistency-RFT, while -Consistency-RFT means only with Diversity-SFT.}
        \label{tab:ablation_consistency_rft}
        \resizebox{\textwidth}{!}{
            \begin{tabular}{lcccccccc}
            \toprule
                \multirow{2}{*}{Arch.} & \multicolumn{3}{c}{Single-class Metrics}& \multicolumn{3}{c}{Multi-class Metrics} & \multicolumn{2}{c}{Synthesis Metrics}\\
                ~ & Acc & AP-I & AP-P & PRO-P & mIoU & FG-mIoU & Cons & Qual \\
                \midrule
                UniDG-RFT    & 77.4 & 99.0 & 77.3 & 91.1 & 44.2 & 31.9 & 4.21 & 4.24 \\
                -Diversity-SFT   & 53.1 & 92.9 & 57.2 & 71.4 & 20.4 & 4.7  & 3.11 & 3.07 \\
                -Consistency-RFT & 74.9 & 98.6 & 77.0 & 91.0 & 41.8 & 28.5 & 4.15 & 4.09 \\
                \bottomrule
            \end{tabular}}
    \end{minipage}
\end{figure*}

\noindent\textbf{Performance comparison only with Consistency-RFT.}
To assess whether \emph{Consistency-RFT} alone is effective when applied to a pre-trained Fill model, we perform an ablation that applies \emph{Consistency-RFT} directly to the pre-trained MM-DiT weights, as reported in Tab.~\ref{tab:ablation_consistency_rft}.
Directly applying \emph{Consistency-RFT} leads to inferior downstream performance and slightly worse generation quality compared with the full recipe.
In contrast, applying \emph{Consistency-RFT} on top of \emph{Diversity-SFT} yields better downstream performance and higher generation quality.
These results indicate that reinforcement learning benefits from a strong initialization that already produces sufficiently faithful reference-based defect synthesis, which in turn supports higher-quality online batch generation and enables GRPO to better adjust the model distribution.

\noindent\textbf{Performance improvement with \textit{\textbf{UDG}} quadruplet context.}
As mentioned in the main paper, the normal-abnormal-mask-caption quadruplets in \textit{\textbf{UDG}} can improve MLLM-based anomaly detection by providing robust contextual information.
Following MMAD~\cite{jiang2024mmad}, which leverages defect-related RAG and expert agents to enhance MLLM performance on anomaly classification and localization, we augment the MMAD context with \textit{\textbf{UDG}} quadruplets from the same defect category.
The results are reported in Tab.~\ref{tab:mmad_results}.
These quadruplets explicitly provide defect-region cues and aligned textual descriptions, leading to improved performance for Qwen2.5-VL-7B-Instruct, which becomes comparable to Qwen2.5-VL-72B-Instruct without quadruplet context.
Beyond MLLM-based approaches, these contextual signals may also benefit vision-language alignment-based anomaly detection methods~\cite{gao2026adaptclip}.

\begin{table}[t]\small
\centering
\caption{\textit{\textbf{UDG}} quadruplets improve MLLM-based anomaly classification and detection.}
\label{tab:mmad_results}
\resizebox{\textwidth}{!}{
\begin{tabular}{l c c c c c c c c c}
\toprule
\multirow{2}{*}{Model} & \multirow{2}{*}{Scale} &
\multicolumn{1}{c}{Anomaly} &
\multicolumn{4}{c}{Defect} &
\multicolumn{2}{c}{Object} &
\multirow{2}{*}{Average} \\
\cmidrule(lr){3-3}\cmidrule(lr){4-7}\cmidrule(lr){8-9}
& & Discrimination & Classification & Localization & Description & Analysis & Classification & Analysis & \\
\midrule
Random Chance      & -- & 50.00 & 25.00 & 25.00 & 25.00 & 25.00 & 25.00 & 25.00 & 28.57 \\
Human (expert)     & -- & 95.24 & 75.00 & 92.31 & 83.33 & 94.20 & 86.11 & 80.37 & 86.65 \\
Human (ordinary)   & -- & 86.90 & 66.25 & 85.58 & 71.25 & 81.52 & 89.58 & 69.72 & 78.69 \\
\midrule
Claude-3.5-sonnet  & -- & 60.14 & 60.14 & 48.81 & 67.13 & 79.11 & 85.19 & 79.83 & 68.36 \\
Gemini-1.5-pro     & -- & {68.63} & 60.12 & {58.56} & 70.38 & 82.46 & 89.20 & 82.25 & 73.09 \\
GPT-4o             & -- & 68.63 & 65.80 & 55.62 & {73.21} & {83.41} & {94.98}s & {82.80} & {74.92} \\
\midrule
Qwen2.5-VL-Instruct  & 7B & 71.10 & 56.02 & 60.69 & 64.13 & 78.26 & 91.49 & 83.67 & 72.19 \\
Qwen2.5-VL-Instruct + Quadruplet  & 7B & 72.48 & 60.83 & 64.43 & 68.37 & 79.82 & 92.11 & 85.29 & 74.76 \\
Qwen2.5-VL-Instruct  & 72B & 72.66 & 62.31 & 67.16 & 73.56 & 81.95 & 94.30 & 86.78 & 76.96 \\
\bottomrule
\end{tabular}
}
\end{table}

\subsection{Inference Efficiency Comparison}
To verify the inference efficiency differences between different methods, we conduct inference on the same GPU device with the default resolution and iterations settings.
The results are shown in Tab.~\ref{tab:app_inference_efficiency}.
UniDG remains competitive with FSAG and image insertion methods because Rectified Flow produces strong results with fewer iterations (28 NFE).

\begin{table*}[htbp]
\centering
\caption{Inference time of different methods.}
\label{tab:app_inference_efficiency}
\resizebox{0.95\textwidth}{!}{
\begin{tabular}{cccccccc}
\toprule
Method & AnoDiff & AnoGen & DualAnoDiff & SeaS & AnyDoor & InsertAny & \textbf{UniDG}  \\ \midrule
Paradigm & \multicolumn{5}{>{\columncolor{gray!15}}c}{DDIM}  & \multicolumn{2}{>{\columncolor{cyan!15}}c}{Rectified Flow} \\
Resolution & $256 \times 256$  & $256 \times 256$  & $512 \times 512$  & $512 \times 512$  & $512 \times 512$  & $768 \times 768$  & $768 \times 768$   \\
Iterations & 50  & 50  & 100  & 25  & 50  & 50  & 28   \\
\textbf{Inference Time~(s)} &  3.47  &  3.21  &  19.68  &  6.73  &  7.59  &  28.04  &  11.59  \\ \bottomrule
\end{tabular} }
\end{table*}

\subsection{Impact of Reward Models during Consistency-RFT}
Consistency-RFT uses two reward models, Defect-Und-Reward and Defect-Recog-Reward, described in Sec.~\ref{subsec:app_consistency_rft}.
We evaluate their impact by varying the reward weights.
When a weight is set to 0, the corresponding reward model is disabled during Consistency-RFT.

\begin{table}[ht]\small
\centering
\caption{Ablation studies on reward models during Consistency-RFT. $w_{gen}$ and $w_{det}$ denote the coefficients of Defect-Und-Reward and Defect-Recog-Reward, respectively.}
\label{tab:app_reward_model}
\resizebox{\textwidth}{!}{
\begin{tabular}{llcccccccc}
\toprule
\multicolumn{2}{c}{Coefficient} & \multicolumn{3}{c}{Single-class Metrics} & \multicolumn{3}{c}{Multi-class Metrics} & \multicolumn{2}{c}{Synthesis Metrics}\\
$w_{gen}$ & $w_{det}$ & Acc & AP-I & AP-P & PRO-P & mIoU & FG-mIoU & Cons & Qual \\
\midrule
0.0 & 1.0 & 78.3 & 99.1 & 78.1 & 91.3 & 45.3 & 32.5 & 4.01 & 3.97 \\
0.2 & 0.8 & 77.6 & 99.0 & 77.6 & 91.2 & 44.7 & 32.2 & 4.15 & 4.17 \\
\rowcolor{cyan!25}
0.5 & 0.5 & 77.4 & 99.0 & 77.3 & 91.1 & 44.2 & 31.9 & 4.21 & 4.24 \\
0.8 & 0.2 & 76.7 & 98.8 & 76.7 & 91.0 & 43.7 & 30.7 & 4.34 & 4.39 \\
1.0 & 0.0 & 71.6 & 97.3 & 75.8 & 88.3 & 39.8 & 27.1 & 4.59 & 4.67 \\
\bottomrule
\end{tabular}}
\end{table}

\section{Downstream Evaluation Training Details}
\label{sec:app_downstream_eval}

\paragraph{Binary U-Net localization.}
For object-specific anomaly detection and localization, we use the same standard U-Net evaluator for all methods.
The network is a discriminative encoder--decoder U-Net with Conv--BN--ReLU blocks, five downsampling stages, bilinear upsampling, skip connections, base channel width $64$, input channels $3$, and output channels $2$.
Each object-specific model is trained with $500$ generated anomaly images and masks resized to $256\times256$.
We use Adam with learning rate $1\times10^{-4}$, batch size $128$, Focal Loss on the softmax mask prediction, and $200$ epochs.
The learning rate follows a MultiStep schedule with milestones at $0.8$ and $0.9$ of total epochs and decay factor $0.2$.
Convolution layers are initialized from $\mathcal{N}(0,0.02)$, BatchNorm weights from $\mathcal{N}(1,0.02)$, and BatchNorm biases to zero.
We evaluate every $10$ epochs and select the checkpoint with the best sum of AUROC-I, AUROC-P, AP-P, and PRO-P.
This best-of-20 rule is applied identically to every compared method.
There is no validation split separate from the evaluation split.
A fixed-epoch rerun without selection is reported in Sec.~\ref{sec:app_review_followup}; it lowers every method by $0.6$--$1.4$ points and does not change the ranking.
The tables in the main paper keep the originally reported best-of-20 numbers. The rerun published aggregate Acc-C, AP-P, PRO-P, and FG-mIoU only, so the per-category tables are left unchanged.

\paragraph{Multi-class U-Net segmentation.}
The multi-class semantic segmentation evaluator uses the same U-Net architecture and training recipe as the binary U-Net, except that the output layer is changed to $74$ channels, corresponding to $73$ MVTec-AD foreground defect classes plus background.
It is trained once on all generated data, rather than separately per object category.
During evaluation, semantic metrics are computed over the foreground defect classes, while anomaly detection metrics merge all foreground predictions into a single anomaly class.

\paragraph{ResNet-34 classification.}
For object-specific anomaly classification, we use a ResNet-34 classifier initialized with ImageNet pre-trained weights.
Images are resized to $512\times512$ and trained with batch size $32$ for $100$ epochs.
We use AdamW with learning rate $6\times10^{-5}$, weight decay $0.01$, betas $(0.9,0.999)$, a cosine learning-rate schedule with $5$ warmup epochs and minimum learning rate $1\times10^{-6}$, mixed-precision training, and gradient clipping at $1.0$.
Training augmentation includes random crop, horizontal flip with probability $0.5$, vertical flip with probability $0.3$, rotation within $15^\circ$, and brightness/contrast jitter of $0.2$; validation uses center crop.

\section{The Additional Quantitative Experiments}
\label{sec:app_quantitative_experiments}
Due to space constraints in the main paper, we only report averaged metrics for single-class anomaly detection and classification.
Here, we additionally provide per-category results.
The detailed single-class anomaly detection metrics, anomaly classification metrics, and generation quality metrics on MVTec-AD are shown in Tab.~\ref{tab:quantitative_detection_comparison}, \ref{tab:quantitative_classification_comparison}, and \ref{tab:quantitative_generation_comparison}, respectively.
We also compare UniDG with SeaS on VisA for single-class anomaly detection, anomaly classification, and multi-class anomaly segmentation, as shown in Tab.~\ref{tab:quantitative_detection_comparison_visa}, \ref{tab:quantitative_classification_comparison_visa}, and \ref{tab:quantitative_multi_class_comparison_visa}, respectively.
UniDG outperforms SeaS across all VisA tasks.
These results also highlight the potential of unified defect generation for addressing continual adaptation challenges in industrial anomaly detection~\cite{li2022towards}, where new product categories are incrementally introduced.

\begin{table*}[htbp]
\centering
\caption{Quantitative results for anomaly detection and localization on MVTec-AD dataset. Bold and underline represent optimal and sub-optimal average results, respectively.}
\label{tab:quantitative_detection_comparison}
\resizebox{0.95\textwidth}{!}{
\begin{tabular}{cccccccccccccccccc}
\toprule
\multirow{2}{*}{\textbf{Method}} & \multirow{2}{*}{\textbf{Metric}} & \multicolumn{15}{c}{\textbf{Category}} & \multirow{2}{*}{\textbf{Average}} \\
 & & bottle & cable & capsule & carpet & grid & hazelnut & leather & metal\_nut & pill & screw & tile & toothbrush & transistor & wood & zipper & \\
\midrule
\multicolumn{18}{l}{\textbf{Few-shot Learning Anomaly Generation}} \\
\multirow{4}{*}{AnoDiff}
 & AUROC-I   & 100.0 & 99.8 & 91.3 & 93.2 & 97.5 & 99.6 & 99.9 & 100.0 & 99.1 & 88.0 & 100.0 & 100.0 & 99.8 & 100.0 & 100.0 & 97.9 \\
 & AP-I      & 100.0 & 99.4 & 97.4 & 95.4 & 99.0 & 99.6 & 100.0 & 100.0 & 98.9 & 93.6 & 100.0 & 100.0 & 99.6 & 100.0 & 100.0 & 98.9 \\
 & AUROC-P   & 99.1  & 82.2 & 97.7 & 97.2 & 91.8 & 99.7 & 99.8 & 99.6 & 99.8 & 93.6 & 99.1 & 98.1 & 94.1 & 98.3 & 99.4 & 96.6 \\
 & AP-P      & 87.5  & 82.2 & 34.5 & 78.6 & 30.3 & 95.5 & 77.5 & 98.2 & 96.7 & 37.2 & 92.9 & 60.3 & 94.1 & 78.4 & 87.3 & 75.4 \\
\midrule
\multirow{4}{*}{AnoGen}
 & AUROC-I   & 98.4  & 98.2  & 95.6  & 93.8  & 99.3  & 99.4  & 99.0  & 98.7  & 98.1  & 78.7  & 100.0  & 100.0  & 97.2  & 98.6  & 100.0 & 97.0 \\
 & AP-I    & 96.7  & 96.1  & 91.9  & 96.5  & 96.0  & 98.5  & 98.8  & 90.6  & 99.1  & 84.8  & 99.1  & 90.3  & 89.8 & 97.8  & 98.5  & 95.0 \\
 & AUROC-P & 99.3  & 98.9  & 98.6  & 97.7  & 99.6  & 99.5  & 99.5  & 99.6  & 99.5  & 87.2  & 100.0  & 100.0  & 96.0  & 99.3  & 100.0 & 98.3 \\
 & AP-P    & 79.0  & 70.2  & 36.2  & 72.5  & 39.6  & 83.2  & 60.7  & 70.1  & 87.5  & 19.4  & 90.3  & 26.8  & 56.8  & 80.1  & 77.1  & 63.3 \\
\midrule
\multirow{4}{*}{DualAnoDiff}
 & AUROC-I   & 96.3 & 95.9 & 81.4 & 87.0 & 89.3 & 98.1 & 99.4 & 99.9 & 93.3 & 90.2 & 99.8 & 100.0 & 93.1 & 100 & 99.9 & 94.9 \\
 & AP-I    & 97.5 & 96.6 & 93.7 & 97.3 & 79.4 & 97.8 & 99.3 & 99.0 & 99.6 & 96.5 & 99.1 & 98.5 & 97.9 & 96.8 & 99.4 & 96.6 \\
 & AUROC-P & 98.7 & 97.2 & 93.9 & 94.4 & 95.0 & 98.7 & 99.7 & 100.0 & 98.2 & 95.5 & 99.9 & 100.0 & 91.4 & 100.0 & 100.0 & 97.5 \\
 & AP-P   & 79.3 & 60.4 & 34.5 & 71.8 & 9.9 & 68.3 & 68.1 & 91.8 & 94.5 & 12.0 & 93.0 & 62.7 & 83.1 & 75.4 & 84.9 & 66.0 \\
\midrule
\multirow{4}{*}{SeaS}
 & AUROC-I   & 99.4 & 92.0 & 89.7 & 95.9 & 97.1 & 99.3 & 97.8 & 100.0 & 95.2 & 97.7 & 99.6 & 85.4 & 95.1 & 100.0 & 98.9 & 96.2 \\
 & AP-I    & 98.1 & 91.8 & 93.1 & 97.8 & 97.3 & 99.7 & 98.8 & 99.0 & 99.3 & 98.3 & 99.2 & 76.2 & 90.1 & 97.7 & 96.4 & 95.5 \\
 & AUROC-P & 99.8 & 94.9 & 97.0 & 98.5 & 98.8 & 99.4 & 99.0 & 100 & 98.8 & 98.9 & 99.8 & 92.2 & 93.7 & 100 & 99.6 & 98.0 \\
 & AP-P    & 87.3 & 71.5 & 33.0 & 78.0 & 48.7 & 92.8 & 60.7 & 95.4 & 86.3 & 39.0 & 91.7 & 26.5 & 57.8 & 79.1 & 75.6 & 68.2 \\
\midrule
\multicolumn{18}{l}{\textbf{Training-Free Few-shot Image Insertion}} \\
\multirow{4}{*}{AnyDoor}
 & AUROC-I   & 88.0 & 68.7 & 52.8 & 73.3 & 85.2 & 96.2 & 83.0 & 82.0 & 87.5 & 83.0 & 96.4 & 80.4 & 72.4 & 93.7 & 61.9 & 80.3 \\
 & AP-I    & 72.7 & 73.8 & 91.9 & 85.6 & 65.2 & 86.9 & 92.0 & 88.0 & 89.3 & 79.9 & 77.1 & 85.9 & 69.8 & 74.1 & 78.4 & 80.7 \\
 & AUROC-P & 94.7 & 70.2 & 78.6 & 87.7 & 92.1 & 97.2 & 92.1 & 93.8 & 96.4 & 91.7 & 97.8 & 87.5 & 63.5 & 97.3 & 81.4 & 88.1 \\
 & AP-P    & 16.2 & 10.9 & 9.6 & 10.3 & 1.3 & 8.2 & 28.9 & 60.8 & 50.7 & 0.6 & 18.6 & 4.8 & 22.9 & 19.4 & 8.7 & 18.1 \\
\midrule
\multirow{4}{*}{InsertAny}
 & AUROC-I   & 97.6 & 96.1 & 83.3 & 97.1 & 99.5 & 98.2 & 99.9 & 99.6 & 91.8 & 90.3 & 100 & 97.5 & 96.7 & 100.0 & 82.5 & 95.3 \\
 & AP-I    & 94.2 & 97.3 & 87.8 & 96.3 & 95.3 & 99.5 & 98.7 & 98.0 & 92.7 & 95.4 & 97.1 & 97.1 & 95.8 & 96.7 & 90.3 & 95.5 \\
 & AUROC-P & 99.0 & 96.8 & 94.6 & 98.8 & 99.8 & 98.9 & 100.0 & 99.9 & 97.7 & 95.0 & 100.0 & 98.8 & 95.3 & 100.0 & 93.3 & 97.9 \\
 & AP-P    & 73.5 & 73.8 & 38.2 & 68.6 & 34.4 & 92.8 & 61.5 & 88.3 & 77.5 & 38.3 & 85.5 & 58.9 & 73.9 & 73.4 & 23.0 & 64.1 \\
\midrule
\multirow{4}{*}{\textbf{UniDG-SFT}}
 & AUROC-I   & 99.8 & 98.3 & 94.4 & 90.8 & 100.0 & 99.7 & 100.0 & 99.6 & 98.3 & 92.3 & 100.0 & 100.0 & 96.5 & 100.0 & 99.9 & 98.0 \\
 & AP-I    & 98.5 & 97.3 & 97.2 & 97.9 & 99.8 & 99.0 & 99.8 & 99.1 & 99.6 & 93.9 & 99.7 & 99.1 & 98.9 & 99.3 & 99.2 & 98.6 \\
 & AUROC-P & 99.9 & 98.3 & 98.2 & 96.2 & 99.8 & 99.8 & 100 & 99.9 & 99.5 & 96.2 & 100.0 & 100.0 & 94.5 & 100.0 & 100.0 & 98.8 \\
 & AP-P    & 88.3 & 63.6 & 54.7 & 76.2 & 57.5 & 77.7 & 79.2 & 96.9 & 95.3 & 40.0 & 96.9 & 68.7 & 87.2 & 86.8 & 85.3 & 77.0 \\
 \midrule
\multirow{4}{*}{\textbf{UniDG-RFT}}
 & AUROC-I & 100.0  & 98.3  & 96.5  & 92.1  & 100.0  & 99.7  & 100.0  & 99.6  & 98.3  & 94.7  & 100.0  & 100.0  & 96.5  & 100.0  & 99.9  & 98.4  \\
 & AP-I    & 99.9  & 98.7  & 98.1  & 97.4  & 99.8  & 99.8  & 100.0  & 99.9  & 99.5  & 96.3  & 100.0  & 100.0  & 95.7  & 100.0  & 100.0  & 99.0  \\
 & AUROC-P & 98.7  & 97.5  & 97.8  & 97.9  & 99.8  & 99.0  & 99.8  & 99.1  & 99.6  & 96.8  & 99.7  & 99.1  & 98.9  & 99.3  & 99.2  & 98.8  \\
 & AP-P    & 88.3  & 63.6  & 56.8  & 76.2  & 58.9  & 77.7  & 79.2  & 96.9  & 95.3  & 42.0  & 96.9  & 68.7  & 87.2  & 86.8  & 85.3  & 77.3  \\
\bottomrule
\end{tabular}
}
\end{table*}

\begin{table*}[htbp]
\centering
\caption{Quantitative results for anomaly classification on MVTec-AD. Higher accuracy indicates that the generated data better matches the real defect categories.}
\label{tab:quantitative_classification_comparison}
\resizebox{0.95\textwidth}{!}{
\begin{tabular}{ccccccccc}
\toprule
Category & AnoDiff & AnoGen & DualAnoDiff & SeaS & AnyDoor & InsertAny & \textbf{UniDG-SFT} & \textbf{UniDG-RFT} \\ \midrule
bottle     & 90.70  & 53.49  & 55.81  & 83.72  & 65.12  & 79.07  & 88.37  & 83.72  \\
cable      & 67.19  & 60.94  & 48.44  & 50.00  & 62.50  & 76.56  & 79.69  & 87.88  \\
capsule    & 66.67  & 24.00  & 49.33  & 30.67  & 65.33  & 41.33  & 68.00  & 73.33  \\
carpet     & 58.06  & 40.32  & 69.35  & 41.94  & 53.23  & 77.42  & 64.52  & 72.58  \\
grid       & 42.50  & 57.50  & 62.50  & 32.50  & 60.00  & 32.50  & 50.00  & 40.00  \\
hazel nut  & 85.42  & 85.42  & 79.17  & 85.42  & 81.25  & 95.83  & 95.83  & 95.83  \\
leather    & 61.90  & 57.14  & 90.48  & 52.38  & 53.97  & 65.08  & 79.37  & 69.84  \\
metal nut  & 59.38  & 62.50  & 57.81  & 85.25  & 39.06  & 57.81  & 90.62  & 85.42  \\
pill       & 59.38  & 47.92  & 37.50  & 26.04  & 48.96  & 36.46  & 64.58  & 76.06  \\
screw      & 48.15  & 39.51  & 32.10  & 59.26  & 49.38  & 30.86  & 34.57  & 29.63  \\
tile       & 84.21  & 96.49  & 100.00  & 91.23  & 61.40  & 80.70  & 96.49  & 89.47  \\
toothbrush & 100.00  & 100.00  & 100.00  & 100.00  & 100.00  & 100.00  & 100.00  & 100.00  \\
transistor & 60.71  & 67.86  & 75.00  & 50.00  & 78.57  & 67.86  & 82.14  & 80.95  \\
wood       & 71.43  & 73.81  & 85.71  & 57.14  & 52.38  & 76.19  & 73.81  & 91.18  \\
zipper     & 69.51  & 53.66  & 17.07  & 41.46  & 56.10  & 36.59  & 54.88  & 84.51  \\
\rowcolor{cyan!25}
\textbf{Average} & 68.35  & 61.37  & 64.02  & 59.13  & 61.82  & 63.62  & 74.86  & 77.36  \\ \bottomrule
\end{tabular} }
\end{table*}

\begin{table*}[htbp]
\centering
\caption{Comparison of synthesis data on IL, IL-a, Reference Adherence, Consistency, Reasonability, Quality on MVTec-AD dataset.}
\label{tab:quantitative_generation_comparison}
\resizebox{0.95\textwidth}{!}{
\begin{tabular}{cccccccccccccccccc}
\toprule
\multirow{2}{*}{\textbf{Method}} & \multirow{2}{*}{\textbf{Metric}} & \multicolumn{15}{c}{\textbf{Category}} & \multirow{2}{*}{\textbf{Average}} \\
 & & bottle & cable & capsule & carpet & grid & hazelnut & leather & metal\_nut & pill & screw & tile & toothbrush & transistor & wood & zipper & \\
\midrule
\multirow{6}{*}{AnoDiff}
 & IL    & 0.16 & 0.37 & 0.16 & 0.20 & 0.39 & 0.29 & 0.39 & 0.24 & 0.23 & 0.23 & 0.50 & 0.15 & 0.27 & 0.34 & 0.23 & 0.28 \\
 & IL-a  & 0.17 & 0.13 & 0.08 & 0.14 & 0.11 & 0.17 & 0.10 & 0.22 & 0.14 & 0.07 & 0.21 & 0.14 & 0.16 & 0.18 & 0.12 & 0.14 \\
 & RefAd & 3.76 & 4.00 & 3.69 & 3.96 & 4.11 & 3.43 & 3.82 & 3.86 & 3.97 & 3.62 & 3.87 & 2.63 & 3.74 & 3.49 & 4.25 & 3.84 \\
 & Cons  & 3.80 & 3.93 & 4.13 & 4.29 & 3.62 & 3.54 & 4.21 & 3.92 & 4.16 & 3.75 & 4.13 & 2.99 & 4.01 & 3.82 & 4.00 & 3.97 \\
 & Reas  & 4.29 & 4.39 & 4.48 & 4.69 & 4.41 & 3.58 & 4.07 & 4.19 & 3.85 & 4.00 & 3.89 & 3.06 & 4.38 & 3.85 & 4.26 & 4.18 \\
 & Qual  & 4.01 & 3.97 & 4.03 & 4.23 & 4.25 & 3.57 & 4.03 & 3.98 & 4.11 & 3.80 & 4.10 & 3.01 & 4.03 & 3.80 & 3.98 & 4.01 \\
\midrule
\multirow{6}{*}{AnoGen}
 & IL    & 0.15 & 0.37 & 0.18 & 0.18 & 0.38 & 0.26 & 0.37 & 0.21 & 0.20 & 0.22 & 0.48 & 0.12 & 0.36 & 0.32 & 0.22 & 0.27 \\
 & IL-a  & 0.13 & 0.11 & 0.08 & 0.08 & 0.10 & 0.10 & 0.08 & 0.11 & 0.11 & 0.07 & 0.14 & 0.07 & 0.27 & 0.12 & 0.07 & 0.11 \\
 & RefAd & 3.73 & 3.90 & 3.23 & 3.82 & 3.27 & 3.93 & 3.48 & 3.83 & 3.47 & 2.86 & 3.42 & 2.66 & 3.60 & 2.73 & 3.88 & 3.50 \\
 & Cons  & 3.66 & 4.06 & 4.15 & 4.36 & 3.49 & 4.12 & 4.11 & 3.97 & 3.63 & 3.39 & 3.53 & 2.94 & 4.05 & 3.58 & 4.17 & 3.87 \\
 & Reas  & 3.99 & 4.19 & 3.67 & 4.38 & 3.68 & 4.27 & 4.16 & 4.12 & 3.76 & 3.31 & 3.66 & 3.03 & 4.07 & 3.44 & 4.20 & 3.91 \\
 & Qual  & 3.79 & 4.05 & 3.63 & 4.23 & 3.49 & 4.10 & 3.97 & 3.98 & 3.67 & 3.16 & 3.57 & 2.94 & 3.83 & 3.27 & 4.09 & 3.77 \\
\midrule
\multirow{6}{*}{DualAnoDiff}
 & IL    & 0.30 & 0.44 & 0.31 & 0.26 & 0.44 & 0.35 & 0.35 & 0.32 & 0.40 & 0.36 & 0.48 & 0.43 & 0.30 & 0.38 & 0.40 & 0.37 \\
 & IL-a  & 0.18 & 0.27 & 0.11 & 0.11 & 0.13 & 0.19 & 0.11 & 0.22 & 0.18 & 0.11 & 0.21 & 0.27 & 0.23 & 0.20 & 0.14 & 0.18 \\
 & RefAd & 3.34 & 3.01 & 3.13 & 4.04 & 3.24 & 3.52 & 3.36 & 3.57 & 3.16 & 3.53 & 3.73 & 2.35 & 3.41 & 3.01 & 3.07 & 3.32 \\
 & Cons  & 3.41 & 3.09 & 3.30 & 4.44 & 3.53 & 3.80 & 3.90 & 3.54 & 3.46 & 3.78 & 4.09 & 2.45 & 3.45 & 3.60 & 3.23 & 3.57 \\
 & Reas  & 3.73 & 3.29 & 3.42 & 4.41 & 3.61 & 3.86 & 3.85 & 3.72 & 3.46 & 3.98 & 4.14 & 2.45 & 3.57 & 3.53 & 3.51 & 3.67 \\
 & Qual  & 3.55 & 3.14 & 3.30 & 4.33 & 3.46 & 3.78 & 3.73 & 3.66 & 3.42 & 3.76 & 4.01 & 2.35 & 3.51 & 3.43 & 3.29 & 3.55 \\
\midrule
\multirow{6}{*}{SeaS}
 & IL    & 0.20 & 0.42 & 0.25 & 0.25 & 0.44 & 0.30 & 0.39 & 0.31 & 0.32 & 0.30 & 0.50 & 0.25 & 0.34 & 0.47 & 0.30 & 0.34 \\
 & IL-a  & 0.23 & 0.16 & 0.13 & 0.19 & 0.26 & 0.13 & 0.17 & 0.26 & 0.17 & 0.10 & 0.28 & 0.18 & 0.17 & 0.29 & 0.19 & 0.19 \\
 & RefAd & 3.06 & 3.84 & 2.84 & 3.48 & 3.30 & 3.56 & 2.84 & 2.83 & 3.18 & 2.91 & 3.34 & 3.13 & 3.14 & 3.10 & 3.61 & 3.26 \\
 & Cons  & 3.26 & 4.05 & 3.78 & 4.15 & 3.93 & 3.84 & 3.75 & 2.92 & 3.70 & 3.57 & 3.64 & 3.29 & 3.70 & 3.71 & 4.08 & 3.76 \\
 & Reas  & 3.50 & 4.15 & 3.37 & 4.16 & 3.83 & 4.10 & 3.59 & 3.20 & 3.62 & 3.50 & 3.76 & 3.46 & 3.64 & 3.65 & 3.95 & 3.74 \\
 & Qual  & 3.28 & 4.03 & 3.28 & 3.97 & 3.74 & 3.87 & 3.32 & 2.99 & 3.48 & 3.28 & 3.56 & 3.38 & 3.51 & 3.52 & 3.86 & 3.58 \\
\midrule
\multirow{6}{*}{AnyDoor}
 & IL    & 0.27 & 0.41 & 0.20 & 0.26 & 0.37 & 0.35 & 0.37 & 0.34 & 0.32 & 0.21 & 0.47 & 0.30 & 0.32 & 0.42 & 0.33 & 0.33 \\
 & IL-a  & 0.16 & 0.13 & 0.07 & 0.14 & 0.14 & 0.14 & 0.10 & 0.21 & 0.14 & 0.05 & 0.21 & 0.12 & 0.15 & 0.18 & 0.14 & 0.14 \\
 & RefAd & 3.16 & 4.14 & 3.37 & 3.97 & 3.29 & 3.84 & 3.65 & 3.04 & 3.22 & 3.94 & 3.43 & 2.65 & 3.69 & 2.91 & 2.99 & 3.48 \\
 & Cons  & 3.09 & 4.22 & 3.56 & 4.48 & 3.73 & 4.01 & 4.09 & 3.13 & 3.55 & 4.21 & 3.60 & 2.75 & 3.79 & 3.30 & 3.65 & 3.76 \\
 & Reas  & 3.54 & 4.38 & 3.69 & 4.52 & 3.75 & 4.23 & 4.11 & 3.12 & 3.53 & 4.41 & 3.83 & 3.01 & 3.82 & 3.23 & 3.38 & 3.83 \\
 & Qual  & 3.27 & 4.25 & 3.55 & 4.36 & 3.61 & 4.03 & 3.98 & 3.06 & 3.48 & 4.18 & 3.66 & 2.96 & 3.76 & 3.19 & 3.27 & 3.70 \\
\midrule
\multirow{6}{*}{InsertAny}
 & IL    & 0.15 & 0.35 & 0.18 & 0.24 & 0.32 & 0.32 & 0.27 & 0.26 & 0.25 & 0.22 & 0.42 & 0.22 & 0.25 & 0.33 & 0.19 & 0.26 \\
 & IL-a  & 0.14 & 0.11 & 0.07 & 0.13 & 0.14 & 0.15 & 0.10 & 0.20 & 0.12 & 0.05 & 0.21 & 0.12 & 0.15 & 0.17 & 0.14 & 0.13 \\
 & RefAd & 3.4 & 4.03 & 3.47 & 4.21 & 3.76 & 3.63 & 3.69 & 3.6 & 3.72 & 3.85 & 3.65 & 3.62 & 3.97 & 3.73 & 4.24 & 3.81 \\
 & Cons & 3.29 & 4.08 & 3.71 & 4.47 & 4.01 & 3.64 & 4.19 & 3.67 & 3.87 & 4.01 & 3.78 & 3.69 & 4.01 & 3.94 & 4.49 & 3.98 \\
 & Reas & 3.67 & 4.29 & 3.89 & 4.57 & 4.07 & 3.8 & 4.24 & 3.9 & 3.89 & 4.17 & 3.93 & 4.1 & 4.15 & 4.08 & 4.55 & 4.12 \\
 & Qual & 3.46 & 4.14 & 3.72 & 4.43 & 4.01 & 3.74 & 4.02 & 3.7 & 3.84 & 4.04 & 3.81 & 3.84 & 4.06 & 3.96 & 4.44 & 3.99 \\
\midrule
\multirow{6}{*}{\textbf{UniDG-SFT}}
 & IL    & 0.15 & 0.40 & 0.19 & 0.27 & 0.35 & 0.33 & 0.32 & 0.27 & 0.26 & 0.26 & 0.46 & 0.18 & 0.30 & 0.38 & 0.23 & 0.29 \\
 & IL-a  & 0.18 & 0.21 & 0.11 & 0.18 & 0.18 & 0.20 & 0.16 & 0.23 & 0.16 & 0.13 & 0.23 & 0.16 & 0.18 & 0.20 & 0.16 & 0.18 \\
 & RefAd & 4.07 & 4.16 & 3.35 & 4.27 & 3.44 & 3.72 & 3.80 & 3.75 & 3.95 & 4.08 & 3.61 & 3.09 & 3.97 & 4.07 & 4.24 & 3.90 \\
 & Cons  & 4.14 & 4.17 & 3.64 & 4.60 & 3.69 & 3.92 & 4.30 & 4.02 & 4.24 & 4.33 & 3.82 & 3.45 & 4.32 & 4.29 & 4.51 & 4.15 \\
 & Reas  & 4.30 & 4.40 & 3.67 & 4.62 & 3.74 & 3.98 & 4.18 & 4.11 & 4.18 & 4.39 & 3.89 & 3.38 & 4.14 & 4.36 & 4.49 & 4.18 \\
 & Qual  & 4.19 & 4.27 & 3.60 & 4.47 & 3.66 & 3.92 & 4.11 & 3.98 & 4.13 & 4.26 & 3.79 & 3.40 & 4.11 & 4.26 & 4.40 & 4.09 \\
 \midrule
\multirow{6}{*}{\textbf{UniDG-RFT}}
 & IL    & 0.15 & 0.38 & 0.19 & 0.24 & 0.34 & 0.32 & 0.30 & 0.27 & 0.26 & 0.25 & 0.46 & 0.21 & 0.28 & 0.38 & 0.21 & 0.28 \\
 & IL-a  & 0.18 & 0.14 & 0.10 & 0.16 & 0.17 & 0.17 & 0.13 & 0.22 & 0.16 & 0.09 & 0.23 & 0.17 & 0.16 & 0.19 & 0.15 & 0.16 \\
 & RefAd & 4.19 & 4.45 & 3.77 & 4.48 & 3.48 & 3.92 & 3.81 & 3.83 & 4.11 & 4.41 & 3.81 & 3.43 & 4.33 & 4.21 & 4.38 & 4.04 \\
 & Cons  & 4.27 & 4.42 & 3.91 & 4.54 & 4.02 & 4.08 & 4.12 & 3.99 & 4.38 & 4.52 & 3.92 & 3.51 & 4.27 & 4.55 & 4.60 & 4.21 \\
 & Reas  & 4.26 & 4.55 & 3.82 & 4.51 & 3.79 & 4.10 & 4.03 & 4.01 & 4.21 & 4.55 & 3.97 & 3.68 & 4.41 & 4.42 & 4.65 & 4.20 \\
 & Qual  & 4.39 & 4.67 & 3.88 & 4.62 & 3.81 & 4.14 & 4.17 & 3.92 & 4.37 & 4.57 & 3.98 & 3.52 & 4.48 & 4.47 & 4.57 & 4.24 \\
\bottomrule
\end{tabular}
}
\end{table*}

\begin{table*}[htbp]
\centering
\caption{Quantitative results for anomaly detection and localization on VisA dataset. Bold and underline represent optimal and sub-optimal average results, respectively.}
\label{tab:quantitative_detection_comparison_visa}
\resizebox{0.95\textwidth}{!}{
\begin{tabular}{ccccccccccccccc}
\toprule
\multirow{2}{*}{\textbf{Method}} & \multirow{2}{*}{\textbf{Metric}} & \multicolumn{12}{c}{\textbf{Category}} & \multirow{2}{*}{\textbf{Average}} \\
 & & candle & capsules & cashew & chewinggum & fryum & macaroni1 & macaroni2 & pcb1 & pcb2 & pcb3 & pcb4 & pipe\_fryum  & \\
\midrule
\multicolumn{15}{l}{\textbf{Few-shot Learning Anomaly Generation}} \\
\multirow{4}{*}{SeaS}
 & AUROC-I    & 73.70  & 78.20  & 85.90  & 97.00  & 90.40  & 81.30  & 63.80  & 96.80  & 88.00  & 91.50  & 98.10  & 71.30  & 84.67 \\
 & AP-I     & 71.60  & 82.70  & 90.50  & 98.10  & 93.60  & 70.70  & 52.10  & 95.90  & 89.00  & 88.40  & 94.80  & 75.50  & 83.58 \\
 & AUROC-P  & 84.10  & 98.10  & 98.30  & 98.80  & 96.90  & 98.10  & 91.40  & 97.60  & 94.70  & 90.50  & 95.60  & 99.10  & 95.27 \\
 & AP-P     & 1.60  & 50.70  & 83.60  & 77.80  & 67.10  & 7.60  & 0.10  & 86.70  & 21.40  & 38.30  & 49.50  & 78.70  & 46.93 \\
\midrule
\multicolumn{15}{l}{\textbf{Training-Free Few-shot Image Insertion}} \\
\multirow{4}{*}{\textbf{UniDG-SFT}}
 & AUROC-I    & 83.10  & 79.40  & 95.30  & 98.10  & 91.60  & 88.50  & 67.20  & 93.80  & 93.10  & 90.00  & 96.40  & 96.20  & 89.39 \\
 & AP-I     & 77.4  & 84.50  & 96.10  & 95.30  & 91.60  & 82.80  & 56.10  & 91.20  & 92.80  & 85.30  & 93.30  & 96.90  & 86.94 \\
 & AUROC-P  & 88.60  & 96.40  & 99.60  & 98.40  & 97.10  & 96.40  & 92.70  & 98.40  & 97.30  & 92.60  & 93.40  & 99.50  & 95.87 \\
 & AP-P     & 17.50  & 52.30  & 85.70  & 73.20  & 66.80  & 13.20  & 7.80 & 88.30  & 38.40  & 42.70  & 58.20  & 83.80  & 52.33 \\
 \midrule
\multirow{4}{*}{\textbf{UniDG-RFT}}
 & AUROC-I    & 89.20  & 83.10  & 94.90  & 98.20  & 90.70  & 87.70  & 72.10  & 95.30  & 96.60  & 89.70  & 96.40  & 96.20  & 90.84 \\
 & AP-I     & 81.20  & 85.70  & 96.80  & 95.50  & 92.30  & 81.60  & 58.70  & 93.40  & 95.20  & 85.80  & 92.70  & 96.80  & 87.98 \\
 & AUROC-P  & 89.30  & 97.90  & 99.90  & 99.20  & 96.80  & 97.20  & 94.50  & 98.60  & 97.10  & 94.90  & 93.80  & 99.20  & 96.53 \\
 & AP-P     & 28.40  & 66.70  & 87.20  & 75.40  & 68.20  & 24.60  & 15.90 & 87.60  & 42.80  & 47.60  & 58.90  & 82.10  & 57.12 \\
\bottomrule
\end{tabular}
}
\end{table*}

\begin{table}[htbp]\small
\centering
\caption{Quantitative results for anomaly classification on VisA. Higher accuracy indicates that the generated data better matches the real defect categories.}
\label{tab:quantitative_classification_comparison_visa}
\resizebox{0.48\textwidth}{!}{
\begin{tabular}{cccccccc}
\toprule
Category & SeaS & \textbf{UniDG-SFT} & \textbf{UniDG-RFT} \\ \midrule
candle        & 73.70  & 87.50  & 91.20  \\
capsules      & 78.20  & 91.20  & 90.40  \\
cashew        & 85.90  & 84.30  & 85.20  \\
chewinggum    & 97.00  & 96.40  & 96.70  \\
fryum         & 90.40  & 92.30  & 91.50  \\
macaroni1     & 81.30  & 89.60  & 92.30  \\
macaroni2     & 63.80  & 78.70  & 82.70  \\
pcb1          & 96.80  & 96.40  & 95.40  \\
pcb2          & 88.00  & 92.10  & 94.30  \\
pcb3          & 91.50  & 90.70  & 92.10  \\
pcb4          & 98.10  & 98.40  & 97.90  \\
pipe\_fryum   & 71.30  & 82.50  & 84.20  \\
\rowcolor{cyan!25}
\textbf{Average} & 84.67  & 90.01  & 91.16  \\ \bottomrule
\end{tabular} }
\end{table}

\begin{table*}[htbp]\small
\centering
\caption{Quantitative results for multi-class anomaly semantic segmentation on the VisA dataset.}
\label{tab:quantitative_multi_class_comparison_visa}
\resizebox{0.95\textwidth}{!}{
\begin{tabular}{lcccccccc}
\toprule
\multirow{2}{*}{Methods} & \multicolumn{5}{c}{Binary Anomaly Detection Metrics} & \multicolumn{3}{c}{Semantic Segmentation Metrics}\\
 & AUROC-I & AP-I & AUROC-P & AP-P & PRO-P & mIoU & BG-mIoU & FG-mIoU \\ \midrule

\multicolumn{9}{@{}l@{}}{\cellcolor{gray!10}\scriptsize\textit{Few-shot Anomaly Generation}} \\
SeaS & 81.72 & 82.87 & 95.83 & 45.85 & 78.73 & 24.99 & 99.58 & 11.83 \\

\multicolumn{9}{@{}l@{}}{\cellcolor{gray!10}\scriptsize\textit{Training-Free Few-shot Image Insertion}} \\
\textbf{UniDG-SFT}& \underline{89.21} & \underline{91.34} & \textbf{97.88} & \underline{58.14} & \underline{88.45} & \underline{31.79} & \underline{99.54} & \underline{18.89} \\
\textbf{UniDG-RFT} & \textbf{91.37} & \textbf{92.96} & \underline{97.59} & \textbf{63.20} & \textbf{90.92} & \textbf{36.23} & \textbf{99.55} & \textbf{23.47} \\ \bottomrule
\end{tabular}
}
\end{table*}

\clearpage
\section{Deployment Efficiency: Post-Training Quantization}
\label{sec:app_deployment_efficiency}

A natural concern about UniDG is the cost of running a \textbf{12B}-parameter Rectified-Flow DiT at inference time.
We further validate that UniDG is in fact \emph{practical to deploy on a single consumer GPU} without modifying its training recipe.

\paragraph{Why LoRA + post-training quantization composes cleanly.}
Both \emph{Diversity-SFT} and \emph{Consistency-RFT} are trained as LoRA adapters on a frozen FLUX.1-Fill-dev backbone.
This enables a plug-and-play deployment path: we apply \textbf{SVDQuant}, a post-training W4A4 quantization method, to the frozen backbone, and then \emph{directly merge the already-trained LoRA weights onto the pre-quantized base model}.
No retraining, no re-quantization, and no mode-collapse / instability typically observed when fine-tuning quantized weights.
The BF16$\rightarrow$W4A4 transition is purely a deployment-time step.

\paragraph{Quantitative results.}
Tab.~\ref{tab:app_w4a4_quantization} compares the BF16 checkpoint to the SVDQuant-W4A4 variant on four metrics spanning both generation quality (RefAd, Cons) and downstream anomaly detection (SC-Acc, MC-FG-mIoU).
The W4A4 variant delivers $3.6\times$ model-size reduction ($12$B$\rightarrow 3.3$B effective), $3.1\times$ memory reduction ($28$GB$\rightarrow 9$GB), and $3.0\times$ inference speedup ($11.59$s$\rightarrow 3.9$s), while maintaining near-identical generation quality and downstream AD performance.
W4A4 lowers the numerical precision of the trained 12B model; it is not evidence about a smaller backbone. The 3.2B FLUX-Mini comparison is in Sec.~\ref{sec:app_review_followup}.
Coupled with UniDG's 28-step Rectified-Flow inference (already markedly faster than 50--100-step DDIM baselines, cf.\ Tab.~\ref{tab:app_inference_efficiency}), the W4A4 checkpoint fits on a single \texttt{RTX 4060Ti 16GB} or \texttt{Jetson Orin}, making UniDG genuinely deployable outside research clusters.

\begin{table}[htbp]
\centering
\small
\caption{BF16 versus SVDQuant-W4A4 UniDG: size, memory, latency, and end-to-end quality.}
\label{tab:app_w4a4_quantization}
\begin{tabular}{lccccccc}
\toprule
\textbf{Model} & \textbf{Params} & \textbf{VRAM} & \textbf{Time (s)} & \textbf{RefAd}$\uparrow$ & \textbf{Cons}$\uparrow$ & \textbf{SC-Acc}$\uparrow$ & \textbf{MC-FG-mIoU}$\uparrow$ \\
\midrule
UniDG (BF16) & 12B  & 28\,GB & 11.59 & 4.04 & 4.21 & 77.4 & 31.87 \\
UniDG (W4A4) & 3.3B & 9\,GB  & \phantom{0}3.90  & 4.01 & 4.17 & 76.8 & 31.28 \\
\bottomrule
\end{tabular}
\end{table}

\paragraph{Efficiency--performance knobs.}
Two further knobs give practitioners additional flexibility:
(i) \emph{LoRA rank} --- we use rank $256$ for SFT and $64$ for RFT; smaller ranks may be used for resource-constrained settings.
(ii) \emph{NFE} --- halving the number of denoising steps from $28$ to $14$ is quantified in Sec.~\ref{sec:app_review_followup}.
In all reported comparisons we keep the default 28-NFE BF16 configuration for fairness.

\section{Why General-Purpose Generative Models Struggle with Rare Defects}
\label{sec:app_rare_defects}

Fig.~\ref{fig:teaser_and_dataset}(a) and our qualitative comparisons show that even the most advanced proprietary image generators (e.g., Gemini-3-Pro-Image~\cite{comanici2025gemini}, GPT-Image-1) fail to synthesize realistic rare defects.
To complement the empirical observation, we discuss three factors that explain this failure mode and motivate why a defect-specific generative model is needed.

\paragraph{(i) Defect-image scarcity in web-scale pretraining.}
While web-scale corpora contain billions of industrial, medical, and natural images, these are overwhelmingly \emph{normal} samples.
Genuine defect images are rarely publicly available due to privacy, safety, and intellectual-property concerns, so general-purpose generators receive limited supervision about what realistic defects look like during pretraining.
UDG is, to our knowledge, the first large-scale (300K quadruplets) paired dataset that directly exposes a model to the normal--abnormal--mask--caption structure specific to defects.

\paragraph{(ii) Fine-grained texture understanding vs.\ high-level semantics.}
Defects typically manifest as subtle texture-level perturbations --- micro-cracks, discolorations, surface porosity, filament warpage --- rather than the high-level semantic edits that general-purpose generators are optimized for.
Standard conditioning pathways (text embeddings, IP-Adapter-style DINO/CLIP tokens) compress visual information into low-dimensional semantic vectors and lose exactly the high-frequency details that define a defect.
UniDG's diptych in-context design preserves full spatial fidelity of the reference, and its \emph{Defect Attention} concentrates MM-DiT's joint attention on the defect tokens.

\paragraph{(iii) Spatial precision of synthesis.}
Defect synthesis requires pixel-accurate control over \emph{where} and \emph{how} a defect appears, so the surrounding normal context remains untouched.
Instruction-based generators have coarse spatial control and often bleed edits into non-masked regions.
UniDG inherits explicit fill-mask conditioning from FLUX.1-Fill-dev and further constrains it with \emph{Normal Regularization} on the target-mask region, yielding clean, bounded edits.

In summary, the gap between general-purpose generators and UniDG is not one of raw capacity but of (a) \emph{the data distribution the model has been exposed to} and (b) \emph{the conditioning granularity} available at inference time.
UDG addresses the former, and UniDG's architecture addresses the latter.

\section{Data Contamination Analysis: Defect-Type-Level Holdout}
\label{sec:app_contamination}

Because UDG aggregates data from 50 source datasets, a natural concern is whether MVTec-AD and VisA evaluations could benefit from implicit leakage of related samples into the training set.
We further validate UniDG's generalization claim in two complementary ways.

\paragraph{Image-level isolation (already stated).}
All MVTec-AD and VisA images are \emph{explicitly excluded} from UDG (Sec.~\ref{sec:app_udg}).
The evaluation setting is therefore already zero-shot at the image level.

\paragraph{Concept-level isolation: leave-overlapping-defect-types-out.}
To further probe concept-level transfer, we map the 269 raw defect types in UDG into 28 coarse categories (scratch, hole, breakage, discoloration, \textit{etc.}), remove every UDG quadruplet whose defect category overlaps with those appearing in MVTec-AD / VisA, and retrain UniDG from scratch with the same SFT$\rightarrow$RFT pipeline.
We refer to this setting as \emph{UDG\textsubscript{no-overlap}}.

\paragraph{Why defect-type-level, not domain-level, holdout?}
Industrial data constitute roughly $60\%$ of UDG.
Removing the entire industrial domain would collapse training scale and conflate data-volume effects with distribution shift.
A defect-type-level holdout removes the specific concepts that overlap with the evaluation targets while preserving overall scale --- a more controlled leakage test.

\paragraph{Results.}
Tab.~\ref{tab:app_contamination} reports anomaly-detection performance on MVTec-AD for the full-UDG model and for \emph{UDG\textsubscript{no-overlap}}.
The concept-removed variant loses only $\sim 3\%$ across all four metrics.
We remove real and synthetic samples symmetrically to keep the original 1:1 real/synthetic ratio in \emph{UDG\textsubscript{no-overlap}}.
Combined with Tab.~\ref{tab:scaling_iterations}, which shows downstream AD saturating at 14K SFT iterations while generation quality continues to improve, the evidence strongly supports the hypothesis that UniDG learns \emph{abstract, transferable defect priors} rather than memorizing overlapping category statistics.

\begin{table}[htbp]
\centering
\small
\caption{Concept-level contamination test. Removing UDG samples whose defect categories overlap with MVTec-AD / VisA and retraining from scratch yields only $\sim 3\%$ degradation, indicating that UniDG's performance reflects transferable priors rather than concept leakage.}
\label{tab:app_contamination}
\begin{tabular}{lcccc}
\toprule
\textbf{Training set} & \textbf{SC-Acc}$\uparrow$ & \textbf{SC-AUROC-I}$\uparrow$ & \textbf{MC-AUROC-I}$\uparrow$ & \textbf{MC-FG-mIoU}$\uparrow$ \\
\midrule
UDG (full, 300K)                      & 77.4 & 98.4 & 96.49 & 31.87 \\
UDG\textsubscript{no-overlap} ($\sim$210K)   & 75.3 & 97.6 & 95.83 & 30.15 \\
\midrule
$\Delta$ & $-2.1$ & $-0.8$ & $-0.66$ & $-1.72$ \\
\bottomrule
\end{tabular}
\end{table}

\section{Cross-Domain Generalization: Natural and Medical Benchmarks}
\label{sec:app_cross_domain}

To address whether UniDG's gains are tied to the industrial domain of MVTec-AD / VisA, we further evaluate on two unseen benchmarks from natural and medical regimes that are represented in UDG but differ from the training datasets and visual statistics.

\paragraph{Benchmarks.}
\begin{itemize}\setlength\itemsep{2pt}
  \item \textbf{GoodsAD}~\cite{zhang2024goodsad} (natural / commodity): a subset of the test split covering all $6$ defect types (\emph{broken, dirty, foreign, leakage, stain, crease}) with $330$ defect images.
  \item \textbf{Kvasir-SEG}~\cite{jha2020kvasirseg} (medical / endoscopy): the official test split of $120$ colonoscopy polyp segmentation images with pixel-level masks. Because Kvasir-SEG contains only one defect class (polyp), its multi-class metrics are reported as ``---''. Kvasir-SEG is held out of UDG training.
\end{itemize}

\paragraph{Protocol.}
We follow the same evaluation protocol as in Sec.~\ref{sec:exp}.
Because few-shot anomaly generation baselines (AnoDiff, SeaS, \textit{etc.}) do not have official implementations on GoodsAD or Kvasir-SEG, we focus on \emph{training-free few-shot image-insertion} baselines (AnyDoor, InsertAnything) that apply cleanly across domains.
AnyDoor and InsertAnything are fine-tuned on UDG with the same SFT recipe used for UniDG; UniDG additionally undergoes the \emph{Consistency-RFT} stage.
All three models are then evaluated on GoodsAD and Kvasir-SEG \emph{without any domain-specific fine-tuning}.

\begin{figure*}[htbp]
\centering
\setlength{\tabcolsep}{1pt}
\renewcommand{\arraystretch}{0}
\newcommand{\crossdomcell}[1]{\includegraphics[width=0.135\linewidth,height=0.105\linewidth,keepaspectratio]{#1}}
\newcommand{\crossdomhead}[1]{\makebox[0.135\linewidth]{\scriptsize\bf #1}}
\newcommand{\crossdomlabel}[1]{\makebox[0.135\linewidth]{\scriptsize #1}}

\newcommand{\crossdomrow}[2]{
  \crossdomcell{#1/small_ref.png}\hfill
  \crossdomcell{#1/small_ref_mask.png}\hfill
  \crossdomcell{#1/small_tar.png}\hfill
  \crossdomcell{#1/small_tar_mask.png}\hfill
  \crossdomcell{#1/small_anydoor.png}\hfill
  \crossdomcell{#1/small_ia.png}\hfill
  \crossdomcell{#1/small_unidg.png}
  \\[1pt]
  \crossdomlabel{#2}\hfill\crossdomlabel{}\hfill\crossdomlabel{}\hfill
  \crossdomlabel{}\hfill\crossdomlabel{}\hfill\crossdomlabel{}\hfill\crossdomlabel{}
  \\[4pt]
}

\crossdomhead{Reference}\hfill
\crossdomhead{Ref.\ Mask}\hfill
\crossdomhead{Target}\hfill
\crossdomhead{Tar.\ Mask}\hfill
\crossdomhead{AnyDoor}\hfill
\crossdomhead{InsertAnything}\hfill
\crossdomhead{UniDG (Ours)}
\\[2pt]
\hrule
\vspace{3pt}

\crossdomrow{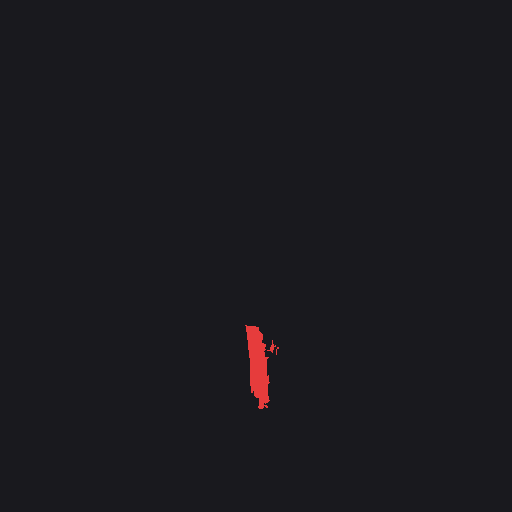}{GoodsAD~\#1}
\crossdomrow{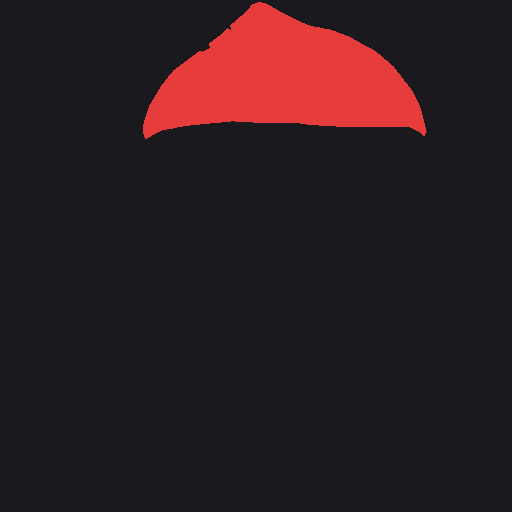}{GoodsAD~\#2}
\crossdomrow{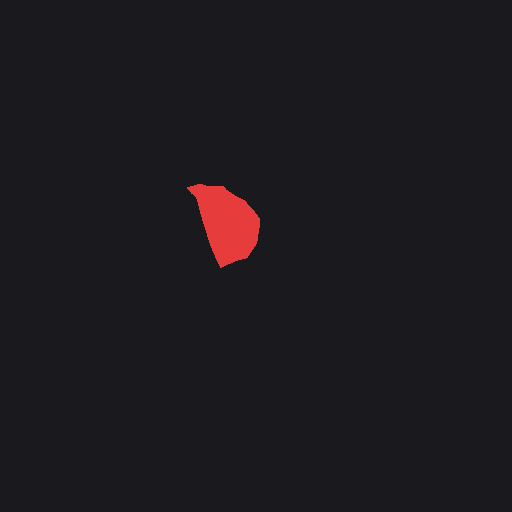}{GoodsAD~\#3}
\crossdomrow{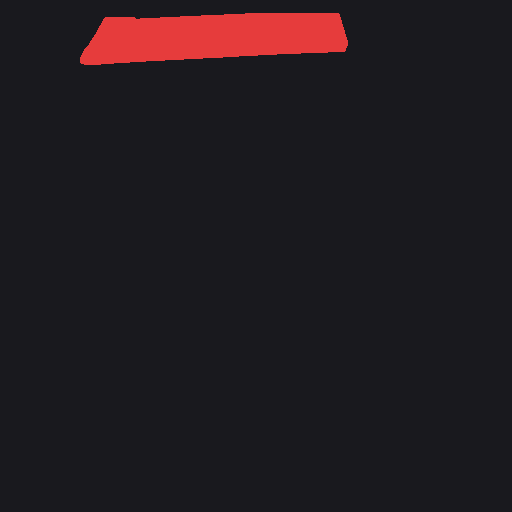}{GoodsAD~\#4}

\vspace{2pt}
\hrule
\vspace{3pt}

\crossdomrow{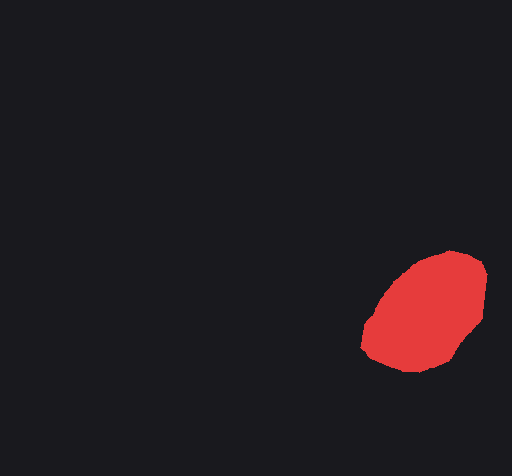}{Kvasir-SEG~\#1}
\crossdomrow{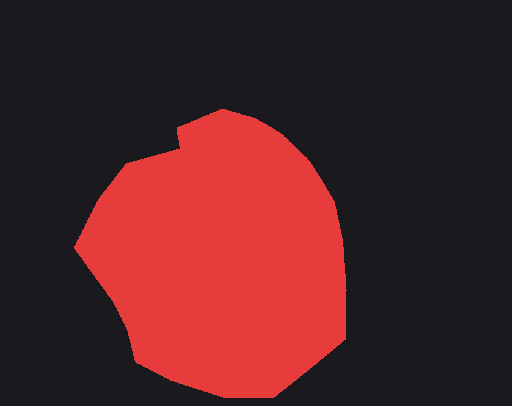}{Kvasir-SEG~\#2}

\caption{
Qualitative cross-domain comparison on natural (GoodsAD) and medical (Kvasir-SEG) benchmarks.
Each row shows one test case with seven columns:
(1) the reference defect image,
(2) the binary mask of the reference defect region,
(3) the target image,
(4) the binary mask of the target region to be edited,
and the defect-synthesis results from
(5) AnyDoor,
(6) InsertAnything, and
(7) UniDG (ours).
Masks are visualized as red foreground on a dark background.
All three methods are trained on UDG and evaluated without any per-domain fine-tuning.
AnyDoor tends to produce incoherent textures that bleed outside the masked region, while InsertAnything often fails to localize the defect.
UniDG consistently produces realistic and spatially well-aligned defects across both natural and medical domains.
}
\label{fig:app_cross_domain_qual}
\end{figure*}

\paragraph{Results.}
Tab.~\ref{tab:app_cross_domain} reports four anomaly-detection metrics across the three benchmarks.
AnyDoor and InsertAnything exhibit substantial degradation when transferred to natural and medical domains, indicating that their representations do not carry transferable cross-scenario defect-synthesis capabilities.
UniDG, by contrast, preserves a large performance margin on these held-out benchmarks without target-benchmark or category-specific fine-tuning.
Few-shot anomaly generation baselines cannot be run here without retraining on the target anomalies, so the comparison is against insertion methods.
We do not claim unrestricted domain-agnostic generalization beyond the evaluated benchmarks.
Qualitative side-by-side comparisons are provided in Fig.~\ref{fig:app_cross_domain_qual}, where UniDG consistently produces higher-fidelity defect synthesis with more realistic appearance and more accurate spatial alignment than both baselines.

\begin{table}[htbp]
\centering
\small
\caption{Cross-domain evaluation on natural (GoodsAD) and medical (Kvasir-SEG) benchmarks. All three methods are trained on UDG and evaluated without any per-domain fine-tuning.}
\label{tab:app_cross_domain}
\resizebox{0.98\textwidth}{!}{
\begin{tabular}{llcccc}
\toprule
\textbf{Benchmark} & \textbf{Method} & \textbf{SC-Acc}$\uparrow$ & \textbf{SC-AP-P}$\uparrow$ & \textbf{MC-PRO-P}$\uparrow$ & \textbf{MC-FG-mIoU}$\uparrow$ \\
\midrule
\multirow{3}{*}{MVTec-AD (Industrial)}
 & AnyDoor         & 61.8 & 18.1 & 79.10 & 11.11 \\
 & InsertAnything  & 63.6 & 64.1 & 77.45 & 15.63 \\
 & \textbf{UniDG-RFT} & \textbf{77.4} & \textbf{77.3} & \textbf{91.12} & \textbf{31.87} \\
\midrule
\multirow{3}{*}{GoodsAD (Natural)}
 & AnyDoor         & 55.2 & 14.3 & 71.84 & \phantom{0}8.26 \\
 & InsertAnything  & 56.8 & 55.7 & 69.53 & 11.42 \\
 & \textbf{UniDG-RFT} & \textbf{74.6} & \textbf{73.8} & \textbf{88.37} & \textbf{28.94} \\
\midrule
\multirow{3}{*}{Kvasir-SEG (Medical)}
 & AnyDoor         & 51.4 & 11.7 & --- & --- \\
 & InsertAnything  & 53.1 & 49.6 & --- & --- \\
 & \textbf{UniDG-RFT} & \textbf{73.2} & \textbf{71.5} & --- & --- \\
\bottomrule
\end{tabular}
}
\end{table}

\section{Architecture vs.\ Training Contribution}
\label{sec:app_arch_vs_training}

Our ablations in the main paper (Tab.~\ref{tab:overall_ablation}) separately study the contribution of architectural components and training-stage components.
To directly disentangle \emph{architecture} from \emph{training strategy}, we apply exactly the same \emph{Consistency-RFT} objective to a strong in-context insertion baseline (InsertAnything) and compare it to UniDG under matched training conditions. InsertAnything uses the same FLUX.1-Fill-dev and FLUX.1-Redux-dev backbones as UniDG, so both models have comparable 12B-scale capacity.

\begin{table}[htbp]
\centering
\small
\caption{Disentangling architecture from training strategy. The same Consistency-RFT is applied on top of InsertAnything (ControlNet-style injection) and UniDG (diptych MM-DiT). UniDG's SFT-only model already surpasses InsertAnything's SFT+RFT model, indicating that architecture is the dominant factor; RFT provides consistent complementary gains on top.}
\label{tab:app_arch_vs_training}
\begin{tabular}{lcccc}
\toprule
\textbf{Method} & \textbf{SC-Acc}$\uparrow$ & \textbf{SC-AP-P}$\uparrow$ & \textbf{MC-PRO-P}$\uparrow$ & \textbf{MC-FG-mIoU}$\uparrow$ \\
\midrule
InsertAnything (SFT)      & 63.6 & 64.1 & 77.5 & 15.6 \\
InsertAnything (SFT+RFT)  & 65.9 & 67.8 & 81.2 & 18.3 \\
\midrule
UniDG (SFT)               & 74.9 & 77.0 & 91.0 & 28.5 \\
\textbf{UniDG (SFT+RFT)}  & \textbf{77.4} & \textbf{77.3} & \textbf{91.1} & \textbf{31.9} \\
\bottomrule
\end{tabular}
\end{table}

Two observations follow.
(i) UniDG SFT-only already exceeds InsertAnything SFT+RFT by a large margin (MC-FG-mIoU $28.5$ vs.\ $18.3$); \emph{architecture is the dominant factor} for defect synthesis quality.
(ii) RFT contributes consistent, complementary gains of similar magnitude ($+3.4$ for UniDG, $+2.7$ for InsertAnything on MC-FG-mIoU), confirming that our two-stage recipe is broadly useful and is not specifically tailored to UniDG's architecture.

\section{Data Curation Thresholds: Calibration and Ablation}
\label{sec:app_curation_thresholds}

The Multi-Agent curation pipeline (Sec.~\ref{sec:app_udg}) retains a caption only when the Captioner's self-reported confidence is $\ge 0.8$ \emph{and} the Verifier's average Likert score is $\ge 4.2$.
Because the confidence signal is self-reported, we further validate that the two thresholds are well-calibrated and robust.

\paragraph{Captioner confidence is well-calibrated.}
We sampled $500$ candidate captions stratified by confidence bin and had human annotators judge each caption as accurate or inaccurate.
The resulting calibration curve is monotonic (Tab.~\ref{tab:app_captioner_calibration}).
At the operating point $\tau_{\mathrm{conf}}=0.8$, precision reaches $92.3\%$ with recall $68.0\%$, confirming that the self-reported score is a meaningful quality signal rather than an arbitrary cutoff.
Beyond calibration, the Captioner Agent integrates multiple independent MLLMs (GPT-5.1, Gemini-3-Pro, Qwen3-VL-235B, GLM-4.6V), and high-confidence captions exhibit higher inter-model agreement in practice, providing an additional cross-model consensus signal on top of the per-model confidence.

\begin{table}[htbp]
\centering
\small
\caption{Captioner confidence calibration on $500$ stratified samples. Human accuracy increases monotonically with self-reported confidence, validating the threshold $\tau_{\mathrm{conf}}=0.8$.}
\label{tab:app_captioner_calibration}
\begin{tabular}{lcc}
\toprule
\textbf{Confidence bin} & \textbf{\#Samples} & \textbf{Human accuracy (\%)} \\
\midrule
$[0.0, 0.5)$ & \phantom{0}38  & 42.1 \\
$[0.5, 0.7)$ & \phantom{0}76  & 61.8 \\
$[0.7, 0.8)$ & \phantom{0}89  & 74.2 \\
$[0.8, 0.9)$ & 142            & 88.7 \\
$[0.9, 1.0]$ & 155            & 95.5 \\
\bottomrule
\end{tabular}
\end{table}

\paragraph{Threshold selection via a pilot study.}
Thresholds were not hand-tuned.
We selected them from a pilot subset of $\approx 1/10$ of the annotation pool with the following protocol.
(i) \emph{Confidence threshold.} We tested filtering ratios of $\{10\%, 20\%, 40\%, 60\%\}$; the operating point $\tau_{\mathrm{conf}}=0.8$ naturally filters out approximately $19.84\%$ of candidates, and manual inspection showed that captions below this bar had substantially higher rates of hallucinated defect descriptions or mis-identified defect types.
(ii) \emph{Verifier threshold.} Among candidates passing $\tau_{\mathrm{conf}}$, we swept the verifier threshold in $\{3.8, 4.0, 4.2, 4.4\}$; thresholds below $4.0$ retained too many borderline-quality samples, $4.4$ was overly aggressive, and $\tau_{\mathrm{ver}}=4.2$ discards approximately $30\%$ of triplets and offered the best precision-retention trade-off.
(iii) \emph{Boundary spot-check.} After applying both thresholds we manually inspected $\approx 5{,}000$ retained quadruplets near the decision boundary; the spot-check confirmed that the chosen operating point produces a high-precision training set with minimal false retentions.

\paragraph{End-to-end validation.}
The quality of the filtered UDG dataset is also validated by three downstream signals that do not depend on the thresholds themselves: the 30-participant human evaluation reported in the main paper, MLLM-independent AD metrics (AUROC, AP, PRO, mIoU) on MVTec-AD / VisA, and cross-domain transfer to GoodsAD and Kvasir-SEG (Sec.~\ref{sec:app_cross_domain}).
Degraded filtering would manifest as weaker downstream performance, which is not observed.

\section{Compute Resources}
\label{sec:app_compute}

The full UniDG pipeline is trained once on an 8$\times$A800 cluster.
Tab.~\ref{tab:app_compute} reports the precise wall-clock and GPU-hour cost of each training stage, including the two reward models used inside \emph{Consistency-RFT}.

\begin{table}[htbp]
\centering
\small
\caption{Total training compute for UniDG. SFT and RFT use an 8$\times$A800 cluster; reward-model runs use $2\times$ and $1\times$ A800 respectively. Inference cost is a single A800 (BF16, 11.59\,s / image) or a single consumer GPU (W4A4, 3.9\,s / image; see Sec.~\ref{sec:app_deployment_efficiency}).}
\label{tab:app_compute}
\begin{tabular}{lcccc}
\toprule
\textbf{Stage} & \textbf{GPUs} & \textbf{Iters / Epochs} & \textbf{Wall-clock} & \textbf{GPU-h} \\
\midrule
Diversity-SFT        & 8$\times$A800 & 14K iters & \phantom{0}47.97\,h  & 383.72 \\
Consistency-RFT      & 8$\times$A800 & 37K iters & \phantom{0}71.98\,h  & 575.81 \\
Defect-Und-Reward    & 2$\times$A800 & 20 epochs & \phantom{0}10.21\,h  & \phantom{00}20.42 \\
Defect-Recog-Reward  & 1$\times$A800 & \phantom{0}5 epochs & \phantom{00}3.34\,h  & \phantom{000}3.34 \\
\midrule
\textbf{Total}       &               &           & \phantom{0}$\approx$\,133\,h & \textbf{983.29} \\
\bottomrule
\end{tabular}
\end{table}

The one-time training cost is approximately $984$ GPU-h for a 12B-parameter rectified-flow model.
All stages share the same cluster and storage; the reward-model runs together account for less than $3\%$ of total compute.
The reported compute does not include preliminary or failed experiments beyond those documented here.
Inference at deployment time runs on a single GPU (single A800 in BF16 or a consumer-grade 16\,GB GPU after W4A4 quantization).

\section{Asset Documentation and Release Plan}
\label{sec:app_asset_documentation}

\paragraph{Existing assets.}
UDG is curated from publicly accessible defect-related datasets and uses existing models and tools, including SAM~\cite{ravi2024samv2}, FLUX.1-Fill-dev, FLUX.1-Redux-dev, and the MLLMs used by the Captioner, Verifier, and independent evaluator.
Tab.~\ref{tab:app_sources} lists the public sources recorded in the curation sheet.
MVTec-AD, VisA, GoodsAD, and Kvasir-SEG are held out of training.
Industrial and natural sources are used under each provider's research terms.
We do not place those upstream datasets under the Apache License 2.0 of the code.
Where a source license does not permit redistribution, the UDG release will contain derived metadata and curation scripts rather than the raw images.
A dash in the citation column means the curation record names the dataset and we did not attach a separate canonical paper.

\begin{longtable}{p{3.3cm}p{1.7cm}p{1.6cm}p{6.4cm}}
\caption{Public source datasets used to curate UDG or held out for evaluation. Citations are the canonical dataset papers we could verify.}
\label{tab:app_sources}\\
\toprule
Dataset & Domain & Use & Citation \\
\midrule
\endfirsthead
\toprule
Dataset & Domain & Use & Citation \\
\midrule
\endhead
\bottomrule
\endfoot
Real-IAD & Industrial & Train & \cite{wang2024realiad} \\
3CAD & Industrial & Train & \cite{yang20253cad} \\
MANTA & Industrial & Train & \cite{fan2025manta} \\
MVTec 3D-AD & Industrial & Train & \cite{bergmann2022mvtec3d} \\
MVTec AD 2 & Industrial & Train & \cite{heckler2026mvtecad2} \\
Magnetic tile (MTD) & Industrial & Train & \cite{huang2020magnetictile} \\
BSData & Industrial & Train & \cite{schlagenhauf2021bsdata} \\
MPDD & Industrial & Train & \cite{jezek2021mpdd} \\
BTAD & Industrial & Train & \cite{mishra2021btad} \\
MIAD & Industrial & Train & \cite{bao2023miad} \\
Defect Spectrum & Industrial & Train & \cite{yang2024defectspectrum} \\
KolektorSDD2 & Industrial & Train & \cite{bozic2021kolektorsdd2} \\
RSDDs & Industrial & Train & \cite{gan2017rsdds} \\
Crack500 & Industrial & Train & \cite{yang2020crack500} \\
GAPs & Industrial & Train & \cite{eisenbach2017gaps} \\
DAGM & Industrial & Train & \cite{wieler2007dagm} \\
Anomaly-ShapeNet & Industrial & Train & \cite{li2024anomalyshapenet} \\
NanoTWICE & Industrial & Train & \cite{carrera2017nanotwice} \\
Robust AD & Industrial & Train & \cite{pemula2025robustad} \\
CSDD & Industrial & Train & \cite{mao2025csdd} \\
PD-REAL & Industrial & Train & \cite{qin2023pdreal} \\
WFDD & Industrial & Train & -- \\
RAD & Industrial & Train & -- \\
VISION & Industrial & Train & -- \\
Insulator & Industrial & Train & -- \\
MSD & Industrial & Train & -- \\
PIAD & Industrial & Train & -- \\
MulSen-AD & Industrial & Train & -- \\
Variety & Industrial & Train & -- \\
MAD & Industrial & Train & -- \\
AFID & Industrial & Train & -- \\
AutoVI & Industrial & Train & -- \\
AeBAD & Industrial & Train & -- \\
Texture-AD & Industrial & Train & -- \\
ISP-AD & Industrial & Train & -- \\
IPAS & Industrial & Train & -- \\
ContinualAD & Industrial & Train & -- \\
Coffee leaves & Natural & Train & -- \\
ISIC & Medical & Train & \cite{codella2018isic,bao2024bmad} \\
CVC-ClinicDB & Medical & Train & \cite{bernal2015clinicdb} \\
CVC-ColonDB & Medical & Train & \cite{bernal2012colondb} \\
EndoScene & Medical & Train & \cite{vazquez2017endoscene} \\
TN3K & Medical & Train & \cite{gong2021tn3k} \\
RESC & Medical & Train & \cite{bao2024bmad} \\
LiTS & Medical & Train & \cite{bilic2023lits,bao2024bmad} \\
BTCV / liver & Medical & Train & \cite{landman2015btcv,bao2024bmad} \\
BraTS 2018--2020 & Medical & Train & \cite{menze2015brats,bakas2017brats,bakas2018brats} \\
MVTec-AD & Industrial & Held out & \cite{bergmann2019mvtec} \\
VisA & Industrial & Held out & \cite{zou2022visa} \\
GoodsAD & Natural & Held out & \cite{zhang2024goodsad} \\
Kvasir-SEG & Medical & Held out & \cite{jha2020kvasirseg} \\
\end{longtable}

\paragraph{Medical sources.}
We do not hold patient-level consent documents for any medical source, and we do not claim to have verified consent upstream.
Each source was released for research reuse by the institution that collected it, and we use it through that provider's access route.
De-identification is largely a property of the images rather than a check we performed: colonoscopy and endoscopy frames show the lumen, OCT shows retinal cross-sections, the BraTS volumes are distributed skull-stripped and template-registered, the liver scans are abdominal, and the TN3K frames are cropped to the ultrasound field.
ISIC is the exception.
Dermoscopic close-ups could in principle contain a distinguishing mark.
Contributors warrant that uploads contain no protected health information, and the archive states that it does not independently verify that warranty.
We therefore do not describe the ISIC images as de-identified by us.
BraTS is used under its data-use agreement, which prohibits re-identification and any reconstruction of the face, head, or body.
The publications required by that agreement are cited in Tab.~\ref{tab:app_sources}.
CVC-ClinicDB, CVC-ColonDB, EndoScene, and Kvasir-SEG are limited to research and education and do not permit commercial use.
Kvasir-SEG is an evaluation set only.
The retinal OCT and liver splits follow the BMAD reorganization~\cite{bao2024bmad}.
The code license does not cover these images.
We do not redistribute medical images, the quadruplets built from them, the masks or captions derived from them, or synthetic medical images.
For each medical record, the UDG release will include the source dataset, the original image identifier, and the mask, caption, and generation parameters needed to rebuild the sample, together with the curation scripts.
A user who obtains the source data on the provider's own terms can reconstruct that subset.
Generated medical abnormalities are not evidence of clinical validity and must not enter diagnostic-model training or clinical decisions without specialist review.

\paragraph{UDG dataset card.}
UDG contains 300K normal-abnormal-mask-caption quadruplets spanning industrial, natural, and medical defect-related scenarios.
It is intended for research on defect generation, anomaly detection/localization, multimodal defect understanding, and robustness analysis.
It is not intended for fabricating visual evidence, manipulating inspection records, or making safety-critical deployment decisions without human validation.
Known limitations include imperfect captions or masks that may survive automated verification, uneven coverage across rare defect types, and potential biases inherited from the public source datasets.
The UDG dataset and UDG-Reward-Bench will be released with documentation, curation code, and source attribution, subject to the upstream constraints above.

\paragraph{UniDG model card.}
UniDG is a 12B-parameter reference-conditioned defect generation model based on FLUX.1-Fill-dev and FLUX.1-Redux-dev.
Code and released checkpoints are available at \url{https://github.com/RetoFan233/UniDG} under the Apache License 2.0.
FLUX.1-Fill-dev and FLUX.1-Redux-dev weights remain under the Black Forest Labs license.
The released checkpoints are for benign defect synthesis, data augmentation, and research benchmarking.
Users should not use the model to create deceptive inspection evidence or to bypass quality-control processes.
The model may fail under large reference-target scale, lighting, or texture mismatch and should be evaluated in the target deployment domain before use.

\section{Human Subjects}
\label{sec:app_human_subjects}
The 30-participant user study, the UDG-Reward-Bench annotations, and the caption-calibration check were done by students recruited within the authors' institution.
They were not hired through an external crowd platform.
Participation was voluntary.
Raters were told the purpose of the task and that the ratings would be used for research, and they agreed before starting.
The tasks ask people to score already collected or synthesized defect images, or to mark a caption as correct or incorrect.
They do not collect personal images and they do not make clinical decisions.
The study was reviewed internally under the institution's procedure for minimal-risk image-rating studies.

Annotators were paid per completed task.
The effective rate is approximately RMB 60 per hour, corresponding to about USD 8--10 per hour, and is above the applicable statutory minimum hourly wage.
The median time is approximately 45 seconds for one paired user-study case, in which the rater gives four 1--5 scores for the UniDG-SFT image and the UniDG-RFT image.
It is approximately 3 minutes for one UDG-Reward-Bench pair when the annotator also writes the four-dimension analysis, and approximately 15 seconds for one caption judged correct or incorrect.
A user-study participant completed one rating session, on the order of one hour.
A reward-bench annotator completed on the order of 100 pairs, about five hours.
The 500-caption calibration is about two hours in total.
These times are rounded from the work each task requires.
The full instruction sheets and rating-interface screenshots are not included here.

\section{Additional Analyses from the Review Discussion}
\label{sec:app_review_followup}

This section records the measurements added in response to the reviews: unique data scale, failure rates, KID/IS, inference steps, a smaller backbone, mask sensitivity, split isolation, and checkpoint selection.
A short glossary is included because several table symbols were defined far from where they are used.

\subsection{Metric glossary}
\begin{itemize}
    \item RefAd, Cons, Reas, Qual: MLLM scores in $[1,5]$ for reference adherence, region--background consistency, reasonability, and overall quality.
    \item SC-* / MC-*: single-class (one model per object) / multi-class (one model over all categories).
    \item Acc / Acc-C: anomaly classification accuracy of the downstream ResNet-34.
    \item AUROC-I/P, AP-I/P, PRO-P: image-/pixel-level AUROC, average precision, and per-region overlap.
    \item mIoU / FG-mIoU / BG-mIoU: mean IoU over all 74 classes / the 73 foreground defect classes / background.
    \item IL / IL-a: intra-cluster LPIPS over the whole image / inside the defect region.
    \item NFE: number of denoising steps at inference.
\end{itemize}

\subsection{Which components carry the result}
Reading Tab.~\ref{tab:overall_ablation} as successive FG-mIoU increments ($13.3\rightarrow 31.9$) attributes 17\% to diptych Defect-Context Editing, 31\% to normal regularization, 4\% to defect attention, 31\% to Diversity-SFT, and 18\% to Consistency-RFT.
The three components that implement the paired-data formulation account for 79\% of the FG-mIoU gain.
Stopping after Diversity-SFT recovers 15.2 of 18.6 FG-mIoU points and 17.1 of 19.6 Acc points.
Removing Single Object Defect-Fill or Intra-Defect Cross-Sampling from Tab.~\ref{tab:data_allocation} drops FG-mIoU from 28.5 to 14.5 and 15.7; removing Multi-Object Cross-Sampling costs 2.5 points.
Consistency-RFT without Diversity-SFT collapses to FG-mIoU 4.7 (Tab.~\ref{tab:ablation_consistency_rft}).
The 28-way taxonomy enters training almost only through Multi-Object Intra-Defect Cross-Sampling (Sec.~\ref{subsec:app_diversity_sft}), so that 2.5-point drop upper-bounds sensitivity to remapping the 269 raw names.
Evaluation keeps 73 fine-grained classes.

\subsection{Unique data scale versus training iterations}
The default Diversity-SFT recipe traverses UDG once: 14K steps at global batch 64.
Holding the step count fixed while shrinking the unique set would repeat the smaller subsets, and Tab.~\ref{tab:scaling_iterations} shows that over-training already costs FG-mIoU.
We use nested, stratified subsets, checkpoint each size at 6K/10K/14K steps, and report the best.
Each size is a single run.

\begin{table}[htbp]
\centering
\small
\caption{Unique UDG size under an otherwise identical Diversity-SFT recipe. The 300K row is the submitted UniDG-SFT checkpoint.}
\label{tab:app_unique_data_scale}
\begin{tabular}{rccccc}
\toprule
UDG size & PRO-P & FG-mIoU & RefAd & Cons & Qual \\
\midrule
50K  & 86.4  & 23.2  & 3.79 & 4.05 & 4.01 \\
100K & 88.6  & 25.8  & 3.84 & 4.09 & 4.04 \\
200K & 90.1  & 27.4  & 3.88 & 4.12 & 4.07 \\
300K & 90.98 & 28.52 & 3.90 & 4.15 & 4.09 \\
\bottomrule
\end{tabular}
\end{table}

\subsection{Blinded failure audit}
Two annotators independently labelled UniDG-SFT and UniDG-RFT on the same 365 MVTec-AD inputs (five per defect class).
The failure types overlap and do not sum to the any-failure rate.

\begin{table}[htbp]
\centering
\small
\caption{Blinded two-annotator failure audit on 365 paired MVTec-AD outputs.}
\label{tab:app_failure_audit}
\begin{tabular}{lccccc}
\toprule
Model & Any failure $\downarrow$ & Mask violation $\downarrow$ & Ref.\ mismatch $\downarrow$ & Context mismatch $\downarrow$ & Agreement \\
\midrule
UniDG-SFT & 14.2\% & 2.7\% & 7.4\% & 6.3\% & $\kappa=0.70$ \\
UniDG-RFT & 11.0\% & 2.6\% & 5.7\% & 4.9\% & $\kappa=0.70$ \\
\bottomrule
\end{tabular}
\end{table}

By defect type, UniDG-RFT fails on 8.7\% of scratch/abrasion, 10.2\% breakage/hole, 13.2\% discoloration/stain, 15.3\% deformation/fold/wrinkle, and 7.8\% foreign matter (SFT: 11.8 / 13.1 / 16.4 / 19.2 / 10.7\%).
The 575.81 GPU-hours of Consistency-RFT (Tab.~\ref{tab:app_compute}) buy a modest reduction in reference and context mismatch, almost no change in mask violations, and essentially nothing on already-saturated AP/PRO.

\subsection{KID and IS}
Unbiased KID and IS use Inception-V3 pool3 features, 100 subsets of size 100, and identical per-category counts over 15 MVTec-AD categories.
Every entry is recomputed on our generated sets.
AnomalyDiffusion and DualAnoDiff decline FID/KID on anomaly images because overfitted per-category models score better; SeaS reports KID only for generated normal images.
Those per-category baselines lead on KID.

\begin{table}[htbp]
\centering
\small
\caption{Distribution-level synthesis metrics on MVTec-AD.}
\label{tab:app_kid_is}
\begin{tabular}{lcc}
\toprule
Method & KID $\times 10^{3}$ $\downarrow$ & IS $\uparrow$ \\
\midrule
AnomalyDiffusion & 6.8 & 1.79 \\
DualAnoDiff      & 6.1 & 1.91 \\
SeaS             & 7.4 & 1.86 \\
InsertAnything   & 14.3 & 1.71 \\
UniDG-SFT        & 11.5 & 1.84 \\
UniDG-RFT        & 10.4 & 1.87 \\
\bottomrule
\end{tabular}
\end{table}

The clean comparison is InsertAnything, which also uses no per-category training.
DualAnoDiff also leads on IL (0.37 versus 0.28), so realism and diversity are a trade-off rather than a joint win.

\subsection{Inference steps: 14 versus 28 NFE}
With the UniDG-RFT checkpoint, references, masks, and seeds held fixed, 14 NFE takes 6.10s per image versus 11.59s at 28 NFE (Tab.~\ref{tab:app_inference_efficiency}), with RefAd/Cons/Qual/FG-mIoU of 3.98/4.15/4.17/30.6 against 4.04/4.21/4.24/31.87.
Reported comparisons stay at 28 NFE.

\subsection{Retained-sample quality}
On 500 stratified retained UDG samples, image/mask validity is 96.4\%, caption correctness 95.6\%, and 97.1\% of recovered normals are judged defect-free and artifact-free.
The Inpainting Agent builds training pairs only.
At inference the user supplies a real normal photograph.
We do not claim performance on proprietary distributions whose overlap with pretraining cannot be audited.

\subsection{Smaller backbone}
FLUX-Mini (3.2B; depth $5{+}10$ rather than $19{+}38$) keeps hidden size 3072, condition dimension 4096, the RoPE configuration, and the FLUX VAE.
It is distilled from FLUX.1-dev (64 input channels) rather than Fill (384), so we widen the stem $64\rightarrow 384$ and zero-initialize the new channels.
The 3.2B model must learn Fill's verbatim-reproduction behaviour during SFT, so the gap is an upper bound on the capacity cost.
Held fixed: UDG data and the three-way mixture, LoRA rank 256, Prodigy, batch 64 at $768\times 768$, normal regularization, defect attention, and the evaluation protocol.
The 12B rows are the existing Tab.~\ref{tab:scaling_iterations} entries at the same step budgets.

\begin{table}[htbp]
\centering
\small
\caption{FLUX-Mini 3.2B versus FLUX.1-Fill 12B at matched Diversity-SFT step budgets.}
\label{tab:app_flux_mini}
\begin{tabular}{llccccc}
\toprule
Backbone & Steps & PRO-P & mIoU & FG-mIoU & Cons & Qual \\
\midrule
FLUX.1-Fill 12B & 2K & 77.2 & 27.0 & 11.0 & 3.03 & 3.10 \\
FLUX-Mini 3.2B  & 2K & 68.3 & 20.6 & 5.0  & 2.36 & 2.45 \\
FLUX.1-Fill 12B & 6K & 80.1 & 29.1 & 13.3 & 3.44 & 3.57 \\
FLUX-Mini 3.2B  & 6K & 73.4 & 24.0 & 7.8  & 2.88 & 2.99 \\
\bottomrule
\end{tabular}
\end{table}

The 3.2B backbone retains 45\% of FG-mIoU at 2K steps and 59\% at 6K.
On the same 12B backbone, changing only the formulation (InsertAnything, same UDG data, Tab.~\ref{tab:app_arch_vs_training}) retains 55\% after Diversity-SFT and 57\% after the full recipe.
The capacity gap narrows from 2K to 6K steps; the formulation gap does not.

\subsection{Mask-quality sensitivity}
Adaptive Defect Crop (Alg.~\ref{alg:adaptive_defect_crop}) dilates the target mask with a $7\times 7$ kernel for two iterations, and every compared method receives the same AnomalyDiffusion masks.
The sweep below perturbs only UniDG-RFT target masks at inference.

\begin{table}[htbp]
\centering
\small
\caption{UniDG-RFT on MVTec-AD when the target mask is perturbed before generation.}
\label{tab:app_mask_sensitivity}
\begin{tabular}{lcc}
\toprule
Target mask & FG-mIoU & RefAd \\
\midrule
Unperturbed & 31.87 & 4.04 \\
Dilate 3 / 7 / 11 px & 31.4 / 30.3 / 28.2 & 4.02 / 3.98 / 3.89 \\
Erode 3 / 7 / 11 px & 30.5 / 28.1 / 24.6 & 3.99 / 3.91 / 3.72 \\
Translate 5 / 10 px & 29.6 / 26.5 & 3.95 / 3.82 \\
\bottomrule
\end{tabular}
\end{table}

RefAd moves less than FG-mIoU because mask geometry does not touch the reference stream.
Erosion hurts more than dilation.
The worst setting, erosion by 11 px, still leaves FG-mIoU at 24.6, above InsertAnything's 15.63 with unperturbed masks.

\subsection{Split isolation}
We hash the final UDG manifest with SHA-256 over decoded pixels and with pHash (Hamming distance $\le 5$ as a candidate; every candidate was inspected).

\begin{table}[htbp]
\centering
\small
\caption{Overlap between UDG training, generation references, and evaluation images.}
\label{tab:app_split_isolation}
\begin{tabular}{lc}
\toprule
Intersection & Overlaps \\
\midrule
UDG training vs.\ MVTec-AD evaluation images & 0 \\
UDG training vs.\ VisA evaluation images & 0 \\
Generation references vs.\ downstream evaluation images & 0 \\
\bottomrule
\end{tabular}
\end{table}

\subsection{Fixed-epoch downstream evaluation}
Every downstream model was re-evaluated at epoch 200 with the best-of-20 rule removed.
The 500-image budget, architectures, and schedule are unchanged.
Only the aggregate metrics below were remeasured.

\begin{table}[htbp]
\centering
\small
\caption{Single-class metrics under best-of-20 selection versus a fixed final epoch.}
\label{tab:app_fixed_epoch_sc}
\resizebox{\linewidth}{!}{
\begin{tabular}{lcc}
\toprule
Method & Acc-C best-of-20 $\rightarrow$ fixed & AP-P best-of-20 $\rightarrow$ fixed \\
\midrule
AnoDiff        & $68.4 \rightarrow 67.1$ & $75.4 \rightarrow 74.2$ \\
AnoGen         & $61.4 \rightarrow 60.2$ & $63.3 \rightarrow 62.1$ \\
DualAnoDiff    & $64.0 \rightarrow 62.7$ & $66.0 \rightarrow 64.6$ \\
SeaS           & $59.1 \rightarrow 57.9$ & $68.2 \rightarrow 66.9$ \\
AnyDoor        & $61.8 \rightarrow 60.4$ & $18.1 \rightarrow 17.2$ \\
InsertAnything & $63.6 \rightarrow 62.4$ & $64.1 \rightarrow 62.8$ \\
UniDG-SFT      & $74.9 \rightarrow 73.6$ & $77.0 \rightarrow 75.7$ \\
UniDG-RFT      & $77.4 \rightarrow 76.1$ & $77.3 \rightarrow 76.1$ \\
\bottomrule
\end{tabular}}
\end{table}

\begin{table}[htbp]
\centering
\small
\caption{Multi-class metrics under best-of-20 selection versus a fixed final epoch.}
\label{tab:app_fixed_epoch_mc}
\resizebox{\linewidth}{!}{
\begin{tabular}{lcc}
\toprule
Method & PRO-P best-of-20 $\rightarrow$ fixed & FG-mIoU best-of-20 $\rightarrow$ fixed \\
\midrule
AnoDiff        & $81.82 \rightarrow 80.7$ & $23.83 \rightarrow 23.0$ \\
AnoGen         & $79.54 \rightarrow 78.4$ & $26.45 \rightarrow 25.6$ \\
DualAnoDiff    & $79.96 \rightarrow 78.8$ & $26.27 \rightarrow 25.4$ \\
SeaS           & $81.01 \rightarrow 79.9$ & $20.48 \rightarrow 19.7$ \\
AnyDoor        & $79.10 \rightarrow 78.0$ & $11.11 \rightarrow 10.5$ \\
InsertAnything & $77.45 \rightarrow 76.3$ & $15.63 \rightarrow 14.8$ \\
UniDG-SFT      & $90.98 \rightarrow 89.9$ & $28.52 \rightarrow 27.6$ \\
UniDG-RFT      & $91.12 \rightarrow 90.1$ & $31.87 \rightarrow 30.9$ \\
\bottomrule
\end{tabular}}
\end{table}

Removing selection lowers every method by $0.6$--$1.4$ points and leaves the ranking unchanged.
UniDG-RFT over InsertAnything is 13.7 Acc-C and 16.1 FG-mIoU under the fixed protocol, against 13.8 and 16.24 under best-of-20.

\newpage
\section*{NeurIPS Paper Checklist}

\begin{enumerate}

\item {\bf Claims}
    \item[] Question: Do the main claims made in the abstract and introduction accurately reflect the paper's contributions and scope?
    \item[] Answer: \answerYes{}
    \item[] Justification: The abstract and Section~\ref{sec:intro} explicitly list our two contributions (the UDG dataset of 300K quadruplets and the UniDG foundation model with Diversity-SFT/Consistency-RFT), which match the experimental results in Sections on MVTec-AD and VisA.

\item {\bf Limitations}
    \item[] Question: Does the paper discuss the limitations of the work performed by the authors?
    \item[] Answer: \answerYes{}
    \item[] Justification: Limitations on IQA metrics tailored to defects and on the computational cost of the two-stage training are discussed in the last section of the main paper and in the appendix.

\item {\bf Theory assumptions and proofs}
    \item[] Question: For each theoretical result, does the paper provide the full set of assumptions and a complete (and correct) proof?
    \item[] Answer: \answerNA{}
    \item[] Justification: The paper is empirical and does not state formal theorems that require proofs.

    \item {\bf Experimental result reproducibility}
    \item[] Question: Does the paper fully disclose all the information needed to reproduce the main experimental results of the paper to the extent that it affects the main claims and/or conclusions of the paper (regardless of whether the code and data are provided or not)?
    \item[] Answer: \answerYes{}
    \item[] Justification: We describe the dataset construction pipeline (Sec.~3.1), the MM-DiT-based model architecture (Sec.~3.2), and the two-stage Diversity-SFT/Consistency-RFT training recipe including optimizer, learning rate and iteration counts in the main paper and the appendix.

\item {\bf Open access to data and code}
    \item[] Question: Does the paper provide open access to the data and code, with sufficient instructions to faithfully reproduce the main experimental results, as described in supplemental material?
    \item[] Answer: \answerYes{}
    \item[] Justification: Code and released checkpoints are available at \url{https://github.com/RetoFan233/UniDG} under the Apache License 2.0. The appendix gives the implementation and training details. The UDG dataset and UDG-Reward-Bench will be released. The curation record, source citations, and the restriction against redistributing medical images are in Sec.~\ref{sec:app_asset_documentation}.

\item {\bf Experimental setting/details}
    \item[] Question: Does the paper specify all the training and test details (e.g., data splits, hyperparameters, how they were chosen, type of optimizer) necessary to understand the results?
    \item[] Answer: \answerYes{}
    \item[] Justification: The model training details (resolution, batch size, learning rate, optimizer, number of iterations, and sampling settings) are listed in the implementation-details subsection of the main paper and appendix. Downstream evaluation protocols are described in the experiments section, and complete reproduction scripts with all evaluator hyperparameters will accompany the public release.

\item {\bf Experiment statistical significance}
    \item[] Question: Does the paper report error bars suitably and correctly defined or other appropriate information about the statistical significance of the experiments?
    \item[] Answer: \answerNo{}
    \item[] Justification: Because foundation-model-scale training runs are prohibitively expensive, we report single-run numbers for the main tables rather than error bars; we do, however, report consistent gains across all 15 MVTec-AD and 12 VisA categories, which serves as a proxy for robustness.

\item {\bf Experiments compute resources}
    \item[] Question: For each experiment, does the paper provide sufficient information on the computer resources (type of compute workers, memory, time of execution) needed to reproduce the experiments?
    \item[] Answer: \answerYes{}
    \item[] Justification: The compute resources (number and type of GPUs, training time, and total compute for both the SFT and RFT stages, as well as for the multi-agent dataset curation pipeline) are reported in the appendix.

\item {\bf Code of ethics}
    \item[] Question: Does the research conducted in the paper conform, in every respect, with the NeurIPS Code of Ethics \url{https://neurips.cc/public/EthicsGuidelines}?
    \item[] Answer: \answerYes{}
    \item[] Justification: The research conforms to the NeurIPS Code of Ethics: the anomaly detection datasets used are publicly available research benchmarks, no personally identifiable information is involved, and the limited human-subject components are minimal-risk image-rating/annotation tasks disclosed in Items 14--15.

\item {\bf Broader impacts}
    \item[] Question: Does the paper discuss both potential positive societal impacts and negative societal impacts of the work performed?
    \item[] Answer: \answerYes{}
    \item[] Justification: We discuss positive impact on industrial inspection and medical/safety-critical anomaly detection, and potential negative impact (e.g., misuse of defect-synthesis models to fabricate visual evidence) together with mitigation via responsible release, in the broader-impact paragraph of the conclusion.

\item {\bf Safeguards}
    \item[] Question: Does the paper describe safeguards that have been put in place for responsible release of data or models that have a high risk for misuse (e.g., pre-trained language models, image generators, or scraped datasets)?
    \item[] Answer: \answerYes{}
    \item[] Justification: Although UniDG is intended for benign defect synthesis, it is an image generator and may be misused to fabricate defect evidence. Sec.~\ref{sec:app_broader_impacts} states intended use, out-of-scope use, known limitations, and the restriction against deceptive or evidentiary misuse. Medical images are not redistributed.

\item {\bf Licenses for existing assets}
    \item[] Question: Are the creators or original owners of assets (e.g., code, data, models), used in the paper, properly credited and are the license and terms of use explicitly mentioned and properly respected?
    \item[] Answer: \answerYes{}
    \item[] Justification: All existing assets (MVTec-AD, VisA, SAM, FLUX.1-Fill-dev, Gemini-3-Pro-Image, and the other 50+ anomaly detection datasets used for data curation) are properly cited with their licenses and terms of use acknowledged in the appendix.

\item {\bf New assets}
    \item[] Question: Are new assets introduced in the paper well documented and is the documentation provided alongside the assets?
    \item[] Answer: \answerYes{}
    \item[] Justification: The appendix documents the UDG dataset and UniDG model, including curation, source citations, intended use, out-of-scope use, and known limitations. The code at \url{https://github.com/RetoFan233/UniDG} is released under the Apache License 2.0. The UDG dataset and UDG-Reward-Bench will be released. Upstream dataset licenses, including the medical sources, are not replaced by that code license.

\item {\bf Crowdsourcing and research with human subjects}
    \item[] Question: For crowdsourcing experiments and research with human subjects, does the paper include the full text of instructions given to participants and screenshots, if applicable, as well as details about compensation (if any)?
    \item[] Answer: \answerNo{}
    \item[] Justification: The paper involves human subjects in two places: (i) a $30$-participant user study that produces the Good-Same-Bad evaluation comparing UniDG-SFT and UniDG-RFT (Fig.~\ref{fig:gsb}), and (ii) the human-annotated UDG-Reward-Bench and the captioner-calibration check (Sec.~\ref{sec:app_curation_thresholds}). Sec.~\ref{sec:app_human_subjects} reports the approximate pay (RMB 60 per hour), median time per task, and per-annotator workload. The full participant instructions and rating-interface screenshots are not included.

\item {\bf Institutional review board (IRB) approvals or equivalent for research with human subjects}
    \item[] Question: Does the paper describe potential risks incurred by study participants, whether such risks were disclosed to the subjects, and whether Institutional Review Board (IRB) approvals (or an equivalent approval/review based on the requirements of your country or institution) were obtained?
    \item[] Answer: \answerYes{}
    \item[] Justification: The user study only asked raters to compare already-synthesized defect images on aesthetic and consistency dimensions. It involves no sensitive content, no personally identifying data, and no foreseeable risk beyond routine image-quality rating. Raters were informed of the task purpose, the voluntary nature of participation, and the intended research use, and they gave informed consent before starting. The study was reviewed internally under the institution's procedure for minimal-risk image-rating studies (Sec.~\ref{sec:app_human_subjects}).

\item {\bf Declaration of LLM usage}
    \item[] Question: Does the paper describe the usage of LLMs if it is an important, original, or non-standard component of the core methods in this research? Note that if the LLM is used only for writing, editing, or formatting purposes and does \emph{not} impact the core methodology, scientific rigor, or originality of the research, declaration is not required.
    \item[] Answer: \answerYes{}
    \item[] Justification: LLMs/MLLMs are a core methodological component: they are used as the Captioner and Verifier agents in the UDG dataset curation pipeline (described in Sec.~3.1 and the appendix), as teachers for the Defect-Und-Reward model, and as independent evaluators for the MLLM-based synthesis metrics. LLMs are also used for light writing polish without affecting technical content.

\end{enumerate}

\end{document}